\documentclass[lettersize,journal]{IEEEtran}

\usepackage{booktabs} 
\usepackage{tabularx}
\usepackage{graphicx}
\usepackage{amsmath}
\usepackage{amsthm}
\usepackage{amssymb}

\usepackage{dsfont}
\usepackage{fontawesome}
\usepackage{amsfonts}
\usepackage{multirow}
\usepackage{algorithm}
\usepackage[table,xcdraw]{xcolor}
\usepackage{bm} 
\usepackage{algorithmicx}
\usepackage{algpseudocode}
\usepackage{enumitem}
\usepackage{xurl}
\usepackage{hyperref}
\usepackage{cleveref}
\usepackage{tikz}
\usepackage{bbm}
\usepackage{comment}
\usepackage{mathtools}
\usepackage{stmaryrd}
\usepackage{threeparttable}  
\usepackage{tabularx}
\usepackage{array}
\usepackage{pifont}
\usepackage{makecell}

\newcommand{\fullcircle}[1][1ex]{%
  \raisebox{-0.5ex}{ 
    \begin{tikzpicture}
      \fill[fill=black] (0,0) circle (#1); 
    \end{tikzpicture}%
  }%
}

\newcommand{\halfcircle}[1][1ex]{%
  \raisebox{-0.5ex}{ 
    \begin{tikzpicture}
      \draw (0,0) circle (#1); 
      \fill[fill=black] (-#1,0) arc (180:360:#1); 
    \end{tikzpicture}%
  }%
}

\newcommand{\emptycircle}[1][1ex]{%
  \raisebox{-0.5ex}{ 
    \begin{tikzpicture}
      \draw (0,0) circle (#1); 
    \end{tikzpicture}%
  }%
}

\renewcommand{\arraystretch}{1.2}

\usepackage{tabularx}
\usepackage{graphicx} 
\usepackage{booktabs} 
\usepackage{caption}  
\ifCLASSOPTIONcompsoc
  \usepackage[nocompress]{cite}
\else
  \usepackage{cite}
\fi
\ifCLASSINFOpdf
\else
\fi

\usepackage{amsmath}
\ifCLASSOPTIONcompsoc
  \usepackage[caption=false,font=footnotesize,labelfont=sf,textfont=sf]{subfig}
\else
  \usepackage[caption=false,font=footnotesize]{subfig}
\fi

\begin{document}
\title{A Comprehensive Survey on Self-Interpretable Neural Networks}

\author{Yang Ji, 
Ying Sun$^{\ast}$\IEEEmembership{, Member, IEEE}, 
Yuting Zhang, 
Zhigaoyuan Wang, 
Yuanxin Zhuang, 
Zheng Gong,
Dazhong Shen\IEEEmembership{, Member, IEEE}, 
Chuan Qin\IEEEmembership{, Member, IEEE}, 
Hengshu Zhu\IEEEmembership{, Senior Member, IEEE}, 
and Hui Xiong$^{\ast}$\IEEEmembership{, Fellow, IEEE}

\thanks{Ying Sun and Hui Xiong are corresponding authors.}
\thanks{Yang Ji, Ying Sun, Yuting Zhang, Zhigaoyuan Wang, Yuanxin Zhuang, and Zheng Gong are with the Artificial Intelligence Thrust, The Hong Kong University of Science and Technology (Guangzhou), Guangzhou, China (e-mail: \{yji655, yzhang755, zwang901, yzhuang436, zgong768\}@connect.hkust-gz.edu.cn, yings@hkust-gz.edu.cn).}
\thanks{Dazhong Shen is with the College of Computer Science and Technology Nanjing University of Aeronautics and Astronautics, Nanjing 211100, China (e-mail: dazh.shen@gmail.com).}
\thanks{Chuan Qin and Hengshu Zhu are with the Computer Network Information Center, Chinese Academy of Sciences, Beijing 100083, China (e-mail: \{chuanqin0426, zhuhengshu\}@gmail.com).}
\thanks{Hui Xiong is with the Thrust of Artificial Intelligence, The Hong Kong University of Science and Technology (Guangzhou), China and Department of Computer Science and Engineering, The Hong Kong University of Science and Technology, Hong Kong SAR, China (e-mail: xionghui@ust.hk).}
}


\IEEEtitleabstractindextext{
\begin{abstract}
Neural networks have achieved remarkable success across various fields. However, the lack of interpretability limits their practical use, particularly in critical decision-making scenarios. Post-hoc interpretability, which provides explanations for pre-trained models, is often at risk of robustness and fidelity. This has inspired a rising interest in self-interpretable neural networks, which inherently reveal the prediction rationale through model structures. Despite this progress, existing research remains fragmented, relying on intuitive designs tailored to specific tasks. To bridge these efforts and foster a unified framework, we first collect and review existing works on self-interpretable neural networks and provide a structured summary of their methodologies from five key perspectives: attribution-based, function-based, concept-based, prototype-based, and rule-based self-interpretation. 
We also present concrete, visualized examples of model explanations and discuss their applicability across diverse scenarios, including image, text, graph data, and deep reinforcement learning. Additionally, we summarize existing evaluation metrics for self-interpretation and identify open challenges in this field, offering insights for future research. To support ongoing developments, we present a publicly accessible resource to track advancements in this domain: \href{https://github.com/yangji721/Awesome-Self-Interpretable-Neural-Network}{https://github.com/yangji721/Awesome-Self-Interpretable-Neural-Network}.
\end{abstract}
}

\maketitle

\IEEEdisplaynontitleabstractindextext

\IEEEpeerreviewmaketitle
\section{Introduction}
\label{sec:intro}
\IEEEPARstart{O}{ver} the past decade, neural networks have achieved remarkable success in solving complex problems across various fields. This success is largely due to their vast hypothesis space, facilitated by deep signal propagation through hidden units. 
Despite their impressive predictive power, the lack of interpretability poses significant challenges. Without a clear understanding of how the model arrives at its decisions, it becomes difficult to build user trust in the model's predictions, particularly in critical decision-making scenarios. 
As shown in \Cref{tab:motivation}, interpretability serves essential roles across multiple domains, from healthcare to scientific discovery. For example, while deep learning models achieve pathologist-level accuracy in mammography analysis, clinicians need interpretable diagnostic features to trust these predictions~\cite{barnett2021case}. Similarly, in drug discovery, understanding how models predict molecular interactions is fundamental for ensuring development safety~\cite{bai2023interpretable}.

\begin{figure}[t]
    \centering
    \includegraphics[width=\linewidth]{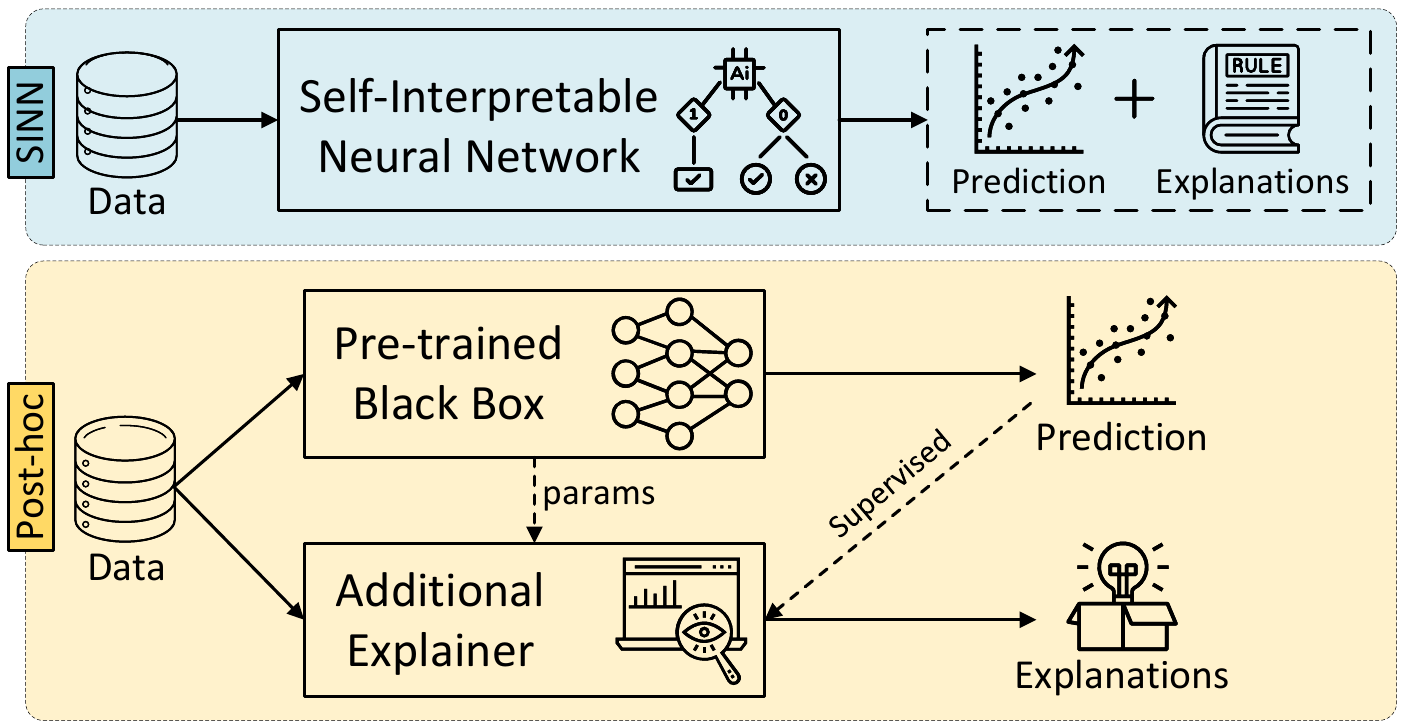} 
    \vspace{-5mm}
    \caption{Comparison between self-interpretation and post-hoc.}
    \label{fig:post-hoc}
    \vspace{-4mm}
\end{figure}

Explainable AI (XAI)~\cite{arrieta2020explainable, lipton2016mythos, doshi2017towards, dwivedi2023explainable} has emerged as a popular research area to address the interpretability challenges of neural networks. Post-hoc interpretability methods~\cite{ribeiro2016should, scott2017unified, shrikumar2017learning}, which provide explanations for pre-trained models, have been widely adopted to enhance model transparency. These methods aim to generate human-understandable explanations for pre-trained model predictions. Despite the flexibility, post-hoc methods often fall short of offering a transparent view of the model's internal workings~\cite{scott2017unified,shrikumar2017learning}. Moreover, they are often computationally expensive and struggle to faithfully capture how pre-trained models actually process and transform inputs to make predictions~\cite{ghorbani2019interpretation, slack2020fooling, slack2021reliable, li2024graph}. As highlighted in recent studies~\cite{laugel2019dangers,frye2020asymmetric}, the inherent opaqueness of these models hinders the understanding of the decision-making process. This growing awareness of the limitations of post-hoc explanations has driven the demand for neural network architectures that can intrinsically reveal the reasoning behind their predictions. As shown in \Cref{fig:post-hoc}, post-hoc methods explain pre-trained models after inference, whereas self-interpretable models inherently encode the reasoning process within the architecture.

\begin{table}[t]
    \centering
    \caption{Applications of Neural Network Interpretability}
    \resizebox{\linewidth}{!}{%
    \begin{tabularx}{\linewidth}{|>{\centering\arraybackslash}m{0.3\linewidth}|>{\centering\arraybackslash}m{0.6\linewidth}|}
    \hline
    \rule{0pt}{2.5ex}Role of Interpretability & Representative Applications\\ \hline
    \rule{0pt}{2.5ex}User Trust Building & Clinical Decision Support~\cite{barnett2021case, mullenbach2018explainable, jiang2024protogate}, Stakeholder Confidence~\cite{luo2018beyond}\\ \hline
    \rule{0pt}{2.5ex}Scientific Discovery & Mechanisms of Drug Action~\cite{bai2023interpretable, wong2024discovery}, \quad Market Trend Analysis~\cite{sun2021market, Yang2024TOIS}\\ \hline
    \rule{0pt}{2.5ex}Model Diagnosis & Error \& Bias Detection~\cite{wong2021using}, Debugging~\cite{bontempelli2022concept}\\ \hline
    \rule{0pt}{2.5ex}System Monitoring & Anomaly Detection~\cite{liu2024towards, wang2023data}\\ \hline
    \end{tabularx}
    }
    \label{tab:motivation}
    \vspace{-3mm}
\end{table}

Traditional machine learning models, such as decision trees and linear models~\cite{rudin2019stop}, are naturally interpretable. However, their optimization and learning paradigms differ fundamentally from those of neural networks, limiting their integration in the current AI landscape, where neural networks dominate. 
For instance, while decision trees are effective for classification tasks when the features are pre-defined, they cannot operate within the gradient-based framework of neural networks, which prevents end-to-end learning~\cite{yang2018deep, wan2020nbdt}. This incompatibility highlights the need to explore self-interpretable neural network approaches that retain the powerful learning capabilities of deep models while offering intrinsic interpretability~\cite{rudin2022interpretable, alvarez2018towards}. 

Recent years have witnessed substantial research efforts in enhancing the self-interpretation of neural networks. Despite the diversity of approaches, these methods share fundamental commonalities in their architectural designs.
However, existing reviews either focus on specific domains or post-hoc explanations~\cite{yang2022unbox, burkart2021survey, arrieta2020explainable, chander2024toward, dwivedi2023explainable}, lacking a unified cross-domain perspective that could effectively connect these self-interpretable designs. To the best of our knowledge, there still lacks a comprehensive review to systematically summarize the current advances of self-interpretable neural networks (SINNs) from a general perspective. This survey aims to distill and categorize the self-interpretable structures found in existing research, establishing a unified framework that facilitates cross-domain integration. We hope this work will help researchers understand the connections between existing methods and offer insights for advancing future research on interpretable neural networks.

In this paper, we provide a comprehensive review of state-of-the-art SINNs. We propose a taxonomy that categorizes existing work into five key categories. This taxonomy design is grounded in the various explanation forms embedded within the architectures of these networks.
\begin{itemize}[noitemsep, left=0pt]
    \item \textbf{Attribution-based} methods identify which input elements most strongly influence the model's prediction.
    \item \textbf{Function-based} methods incorporate explicit, transparent functional components into the network architecture to build interpretable relationships between internal variables.
    \item \textbf{Concept-based} methods design network architectures that explicitly learn and utilize human-understandable concepts as part of their internal representations.
    \item \textbf{Prototype-based} methods structure the network architecture to make predictions through explicit comparisons between inputs and representative cases learned during training.
    \item \textbf{Rule-based} methods incorporate logical rules into the neural network architecture to achieve symbolic reasoning.
\end{itemize}
To further contextualize these approaches, we discuss widely used interpretability techniques and highlight emerging research directions across various domains, including image, text, graph data, and Deep Reinforcement Learning (DRL). By providing a comprehensive review, we aim to analyze the current applications, technical advances, and future directions of SINNs across various domains.

To summarize, our contributions are as follows: 
\begin{itemize}[noitemsep, left=0pt]
    \item \textbf{New Taxonomy and Systematic Review.} We present a unified perspective and conduct an up-to-date, systematic review of SINNs. Existing methods are categorized into five categories, including attribution, function, concept, prototype, and rule-based methods. For each category, we abstract its foundational frameworks, and analyze its interpretable modeling techniques. To the best of our knowledge, this is the first systematic and comprehensive review of SINNs.
    \item \textbf{Application Perspectives.} To better understand the practical implications of SINNs, we explore their applications across four popular domains. We provide concrete examples of model explanations with customized interpretation techniques in these domains.
    \item \textbf{Quantitative Evaluation Metrics.} We offer an extensive overview of the state-of-the-art quantitative evaluation metrics for SINNs, providing researchers with concrete guidelines and references for comparing both existing and emerging methods.
    \item \textbf{Potential Research Directions.} We provide comprehensive analysis and discussions on the relationship between post-hoc interpretability and self-interpretation. We also identify potential research directions for model design advances and emerging application scenarios, such as applications to large language models (LLMs).
\end{itemize}

The rest of this survey is structured as follows: \Cref{sec:preliminary} provides background on the terminology and related research areas. \Cref{sec:methods} presents our taxonomy of SINNs and their common structures. \Cref{sec:applications} explores interpretability techniques from key application perspectives. \Cref{sec:evaluation} summarizes quantitative evaluation metrics. \Cref{sec:discussion} discusses the relationship between post-hoc and self-interpretation and outlines future research directions. \Cref{sec:conclusion} gives a conclusion.

\begin{table*}[t]
    \centering
    \begin{threeparttable}
    \caption{An overview of representative XAI Surveys on different interpretability aspects. \emptycircle[0.8ex] represents ``Not Covered'', \halfcircle[0.8ex] represents ``Partially Covered'', and \fullcircle[0.8ex] represents ``Fully Covered''.
    }
    \vspace*{-1mm}
    \begin{tabular}{c|c|ccccc|cccc}
        \hline
                                                                                                              &                            & \multicolumn{5}{c|}{Self-Interpretation}                                                          & \multicolumn{4}{c}{Application Domains}                                                                   \\ \cline{3-11} 
        \multirow{-2}{*}{Paper}                                                                               & \multirow{-2}{*}{Post-hoc} & Attribution       & Function          & Concept           & Prototype         & Rule              & Image                    & Text                     & Graph                    & DRL                      \\ \hline
        \rowcolor[HTML]{EFEFEF} 
        \cite{burkart2021survey}                                                                              & \fullcircle[1ex]           & \halfcircle[1ex]  & \emptycircle[1ex] & \emptycircle[1ex] & \emptycircle[1ex] & \halfcircle[1ex]  &                          &                          &                          &                          \\
        \cite{arrieta2020explainable}                                                                         & \fullcircle[1ex]           & \halfcircle[1ex]  & \emptycircle[1ex] & \halfcircle[1ex]  & \emptycircle[1ex] & \halfcircle[1ex]  & {\ding{52}} & {\ding{52}} &                          &                          \\
        \rowcolor[HTML]{EFEFEF} 
        \cite{chander2024toward}                                                                              & \halfcircle[1ex]           & \halfcircle[1ex]  & \emptycircle[1ex] & \emptycircle[1ex] & \emptycircle[1ex] & \emptycircle[1ex] &                          &                          &                          &                          \\
        \cite{dwivedi2023explainable}                                                                         & \fullcircle[1ex]           & \halfcircle[1ex]  & \emptycircle[1ex] & \emptycircle[1ex] & \emptycircle[1ex] & \halfcircle[1ex]  &                          &                          &                          &                          \\
        \rowcolor[HTML]{EFEFEF} 
        \cite{gao2024going}                                                                                   & \fullcircle[1ex]           & \halfcircle[1ex]  & \emptycircle[1ex] & \emptycircle[1ex] & \emptycircle[1ex] & \emptycircle[1ex] & {\ding{52}} & {\ding{52}} &                          &                          \\
        \cite{madsen2022post}                                                                                 & \fullcircle[1ex]           & \emptycircle[1ex] & \emptycircle[1ex] & \emptycircle[1ex] & \emptycircle[1ex] & \emptycircle[1ex] &                          & {\ding{52}} &                          &                          \\
        \rowcolor[HTML]{EFEFEF} 
        \cite{calderon2024behalf, lyu2024towards, luo2024local, danilevsky2020survey, zini2022explainability} & \halfcircle[1ex]           & \halfcircle[1ex]  & \emptycircle[1ex] & \halfcircle[1ex]  & \halfcircle[1ex]  & \halfcircle[1ex]  &                          & {\ding{52}} &                          &                          \\
        \rowcolor[HTML]{FFFFFF} 
        \cite{agarwal2021towards, yuan2022explainability, kakkad2023survey}                                   & \halfcircle[1ex]           & \halfcircle[1ex]  & \emptycircle[1ex] & \emptycircle[1ex] & \halfcircle[1ex]  & \emptycircle[1ex] &                          &                          & {\ding{52}} &                          \\
        \rowcolor[HTML]{EFEFEF} 
        \cite{qing2022survey, vouros2022explainable, hickling2023explainability, milani2024explainable}       & \halfcircle[1ex]           & \halfcircle[1ex]  & \emptycircle[1ex] & \halfcircle[1ex]  & \halfcircle[1ex]  & \halfcircle[1ex]  &                          &                          &                          & {\ding{52}} \\ \hline
        \rowcolor[HTML]{FFFFFF} 
        \textbf{Ours}                                                                                         & \emptycircle[1ex]          & \fullcircle[1ex]  & \fullcircle[1ex]  & \fullcircle[1ex]  & \fullcircle[1ex]  & \fullcircle[1ex]  & {\ding{52}} & {\ding{52}} & {\ding{52}} & {\ding{52}} \\ \hline
    \end{tabular}
    \label{tab:survey_comparison}
    \end{threeparttable}%
    \vspace{-1mm}
\end{table*}

\section{Preliminaries}\label{sec:preliminary}
In this section, we begin by describing SINNs and distinguishing them from post-hoc interpretability methods to clarify the scope of our survey. We then provide an overview of existing surveys in interpretable machine learning, emphasizing the unique focus of our survey.

\subsection{Survey Scope}
Building on the definition of interpretable machine learning by~\cite{rudin2019stop, al2020contextual}, SINNs are neural architectures where interpretability is built-in architecturally and enforced through domain-specific constraints. The neural components (neurons, layers, or modules) could represent human-understandable units, allowing the network to explain its decisions based directly on its internal representations, without relying on external parsers or explainers.~\footnote{In this paper, the terms ``interpretability'' and ``explainability'' are used interchangeably to refer to the ability to explain the networks' decisions.} These constraints can vary significantly depending on the application domain. SINNs have two key properties: self-interpretation~\footnote{Throughout the literature, the term ``self-interpretable'' is also known as ``ante-hoc'', ``inherently interpretable'', or ``explainable-by-design'', or ``built-in explanations''. They all refer to models with built-in interpretability rather than post-hoc explanations.}, where the model provides explanations simultaneously with its predictions, and end-to-end differentiability, allowing for efficient neural network training. As shown in~\Cref{fig:post-hoc}, SINNs differ from post-hoc methods in how model explanations are generated. Post-hoc methods work on pre-trained models, interpreting their predictions through a separate explainer. In contrast, SINNs produce both predictions and explanations simultaneously during the inference process. This paper specifically focuses on the interpretability arised directly from the neural structures, excluding non-neural network-based methods, such as algorithmic interpretability~\cite{moshkovitz2020explainable}, dataflow analysis~\cite{qin2023reliable}, and parts of neuro-symbolic methods which reply on external program parsers for structured module organization~\cite{norcliffe2018learning}.

We focus on gathering and reviewing SINN-related papers published in top-tier AI conferences and journals, along with other influential works from the broader research community.
Our goal is to provide a systematic and accessible overview of the state-of-the-art design principles, applications, and evaluation metrics of SINNs. Additionally, we aim to identify commonalities and differences across various SINN architectures and applications, thereby facilitating the development of novel SINN models and applications.

\subsection{Related Surveys}
Recent advancements in XAI have been extensively documented through various surveys. This subsection reviews existing XAI surveys and elucidates the distinct focus of our survey on self-interpretable neural networks.

Several important surveys~\cite{miller2019explanation, langer2021we, mueller2019explanation, saeed2023explainable, vilone2021notions} take an interdisciplinary approach, using ideas from philosophy, psychology, and cognitive science. For example, Miller~\cite{miller2019explanation} combines philosophical and social science views to build a theory of explanations in XAI. Langer et al.\cite{langer2021we} discuss what different users, like developers and regulators, expect from model explanations. Vilone and Longo\cite{vilone2021notions} review definitions and methods for interpretability, connecting theory with practice. These surveys have shaped both technical and interdisciplinary research in XAI.

As summarized in~\Cref{tab:survey_comparison}, numerous studies~\cite{yang2022unbox, burkart2021survey, arrieta2020explainable, chander2024toward, rudin2022interpretable, zhang2021survey, dwivedi2023explainable, gao2024going} has also involved into specific aspects of self-interpretation. However, their primary focus is still on post-hoc interpretability methods, only briefly addressing self-interpretable approaches without thoroughly exploring their unique designs and contributions. Some surveys~\cite{chander2024toward, rudin2022interpretable, zhang2021survey, dwivedi2023explainable} cover some self-interpretation related techniques within the broader XAI context, but they lack a systematic and in-depth exploration of SINNs as an independent topic. This gap underscores the need for a focused survey on SINNs, given their diverse implementations across various domains and different innovative approaches.

Furthermore, several surveys target specific data types or application domains, such as NLP~\cite{luo2024local, danilevsky2020survey, cambria2023survey, zhao2024explainability, zini2022explainability, madsen2022post}, graph data~\cite{agarwal2021towards, yuan2022explainability, kakkad2023survey}, and RL~\cite{qing2022survey, vouros2022explainable, hickling2023explainability, milani2024explainable}. 
While these surveys provide valuable domain-specific taxonomies, SINN architectures across these diverse domains actually share common foundational ideas in their design principles. However, existing surveys lack a unified framework to analyze and categorize these approaches in a systematic way. 
By integrating these shared concepts and examining domain-specific adaptations, our survey provides a cohesive framework for understanding self-interpretable network designs, thereby addressing a significant gap in previous surveys.

\begin{figure*}[h]
    \centering
    \includegraphics[width=0.95\linewidth]{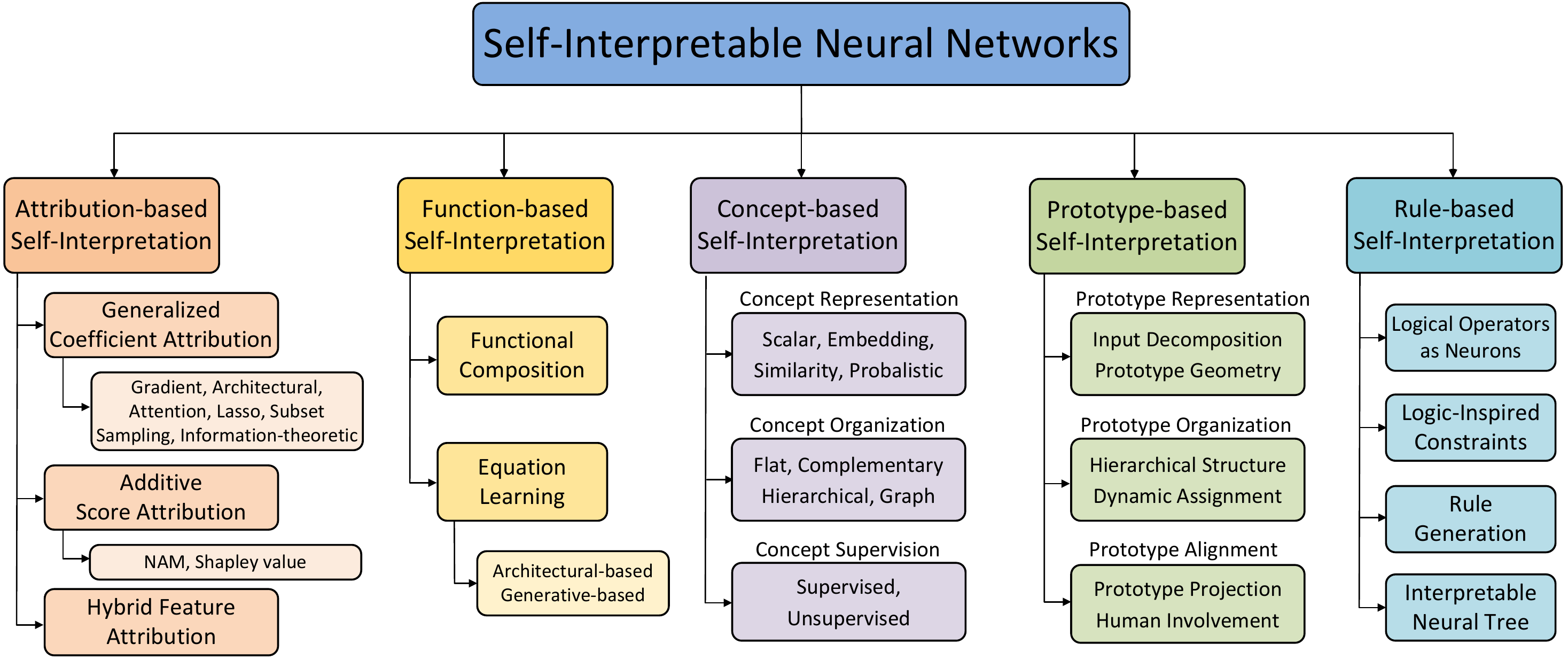}
    \caption{Taxonomy of self-interpretable neural networks.}
    \label{fig:taxonomy}
    \vspace{-2mm}
\end{figure*}

\section{Self-Interpretation Neural Networks}
\label{sec:methods}
In this section, we introduce the general principles of self-interpretation, focusing on ideas that are broadly applicable across various fields while omitting specific application details. For example, we distinguish general interpretable modules from image feature extraction backbones.
Notably, we highlight common structures used in SINNs and classify them based on their desired type of knowledge for interpretation.
As shown in Figure~\ref{fig:taxonomy}, we categorize SINNs into attribution, function, concept, prototype, and rule-based approaches. 

\subsection{Attribution-based Self-Interpretation}
\label{sec:attribution}
Feature attribution is a fundamental paradigm in model interpretation, quantifying how individual features contribute to predictions~\cite{afchar2021towards, schulzrestricting}. Building on this foundation, both post-hoc attribution and self-attribution methods aim to provide meaningful, decomposed explanations of feature contributions.

Notably, while tasks may involve varying input forms, feature attribution methods typically focus on low-dimensional representations. For example, even with high-dimensional image inputs, many self-interpretation models perform attribution on a small set of extracted feature maps, which often use approximation methods to extend attribution to the pixel level.
In this survey, we focus on the attribution component to clarify the core idea and relevance of different self-attribution network structures proposed across domains in a unified way. Domain-specific works incorporating special properties or constraints will be discussed further in \Cref{sec:applications}.

To formalize, let $r$ denote the original raw sample, and let $x = [x_1, x_2, ..., x_N]$ represent the features extracted by a function $h(\cdot)$, where $x_i \in \mathbb{R}^d$. Without loss of generality, we assume a consistent dimensionality $d$ across all features, as shape discrepancies can be managed via projection or reshaping. Common examples include: (1) \textit{Image data:} $r$ is the raw image, $N$ is the number of extracted feature maps, and $d$ is the spatial size of each map; (2) \textit{Tabular data:} $r = x$ is the input feature vector, with $N$ as the number of features and $d = 1$ if each feature is scalar; (3) \textit{Recommender systems:} $r = x$, where $x_i$ is a feature field of dimension $d$, typically one-hot vectors. The task of attribution is to measure the contribution value for each $x_i$ to the output.

The key distinction is that post-hoc methods measure contribution after training, while self-attributing models explicitly model feature contributions as intermediate components in the prediction process. However, training a neural network with a separate structure to model explicit contributions does not guarantee meaningful, domain-grounded attributions. Without explicit constraints, intermediate values in a neural network may encode spurious correlations or depend on higher-order feature interactions, leading to unstable and inconsistent interpretations. As a result, various self-attributing models have been proposed, incorporating regularizers and structural inductive biases to impose constraints on the attribution terms. The following sections categorize these models based on their attribution forms and constraints. We classify them into three categories according to how contributions are measured: \textit{generalized coefficient attribution}, \textit{additive score attribution}, and \textit{hybrid feature attribution}. \Cref{tab:attribution_comparison} provides a comparative summary of these methods and their constraints.

\begin{table*}[t]
    \centering
    \caption{Comparison of Attribution Methods, Constraints, and Representative Studies}
    \resizebox{\linewidth}{!}{%
    \begin{tabular}{cc|c|c}
        \hline
        \multicolumn{1}{c|}{}                                                                                                                                         & Attribution Methods                           & Constraints                                                                                                                                           & Representative Studies                                                                                                                                                                                                                                                  \\ \hline
        \multicolumn{1}{c|}{}                                                                                                                                         & \cellcolor[HTML]{EFEFEF}Gradient-based        & \cellcolor[HTML]{EFEFEF}$\mathcal{C}=\{\boldsymbol{\alpha}(x) \in \mathbb{R}^N :\boldsymbol{\alpha}(x)\approx\nabla_xf(x)\}$                                            & \cellcolor[HTML]{EFEFEF}SENN~\cite{alvarez2018towards}                                                                                                                                                                                                                  \\
        \multicolumn{1}{c|}{}                                                                                                                                         & Architectural                                 & $\mathcal{C} = \left\{ \bm{\alpha}(x) \in \mathbb{R}^N : \bm{\alpha}(x) = \prod_{i=1}^{L} g(W_i x + b_i) \right\}$                                      & Coda-Nets~\cite{bohle2021convolutional}, B-cos Networks~\cite{bohle2024b, bohle2022b}                                                                                                                                                                                   \\
        \multicolumn{1}{c|}{}                                                                                                                                         & \cellcolor[HTML]{EFEFEF}Attention-based       & \cellcolor[HTML]{EFEFEF} $\mathcal{C} = \left\{ \boldsymbol{\alpha}(x) \in \mathbb{R}^N : \boldsymbol{\alpha}_i(x) \geq 0, \ \sum_{i=1}^{N} \boldsymbol{\alpha}_i(x) = 1 \right\}$ & \cellcolor[HTML]{EFEFEF}Attention variants~\cite{zhang2022query, bai2023interpretable, zhu2024propagation, yeh2019interpretable, ren2022diversified, luo2018beyond, wu2021evidence, xu2022towards, zhu2020modeling, arik2021tabnet}                                     \\
        \multicolumn{1}{c|}{}                                                                                                                                         & Lasso-based                                   & $\mathcal{C}=\{\boldsymbol{\alpha}(x) \in \mathbb{R}^N : \|\boldsymbol{\alpha}(x)\|_1 \leq \lambda\}$                                                  & Lasso variants~\cite{lemhadri2021lassonet, he2019interpretable, zhang2025comprehensive, thompson2023contextual, dinh2020consistent}                                                                                                                                     \\
        \multicolumn{1}{c|}{}                                                                                                                                         & \cellcolor[HTML]{EFEFEF}Subset Sampling       & \cellcolor[HTML]{EFEFEF}$\mathcal{C} = \left\{ \boldsymbol{\alpha}(x) \in \{0,1\}^N : \sum_{i=1}^N \boldsymbol{\alpha}_i(x) = K \right\}$               & \cellcolor[HTML]{EFEFEF}Soft~\cite{yoon2018invase, jethani2021have, jiang2024protogate,yang2022locally, yamada2020feature, lee2022self, li2020mri}, Hard~\cite{leonhardt2023extractive, chen2018learning, balin2019concrete, si2024interpretabnet}                      \\
        \multicolumn{1}{c|}{\multirow{-6}{*}{\begin{tabular}[c]{@{}c@{}}Generalized Coefficient \\ Attribution~(Sec \ref{sec:generalized_coefficient})\end{tabular}}} & Information-theoretic                         & $\mathcal{C} = \left\{ \bm{\alpha}(x) : \text{maximize} \ I(\bm{\alpha}(x) x; y) - \beta I(\bm{\alpha}(x) x; x) \right\}$                         & Substructure extraction~\cite{yu2020graph, yu2021recognizing, seo2024interpretable, fang2024exgc, liu2024towards, chen2024tempme, miao2022interpretable, sekhon2023improving, miaointerpretable2023}                                                                    \\ \hline
        \multicolumn{1}{c|}{}                                                                                                                                         & \cellcolor[HTML]{EFEFEF}Neural Additive Model & \cellcolor[HTML]{EFEFEF}Uniqueness of Feature Dependence, Additive Structure                                                                          & \cellcolor[HTML]{EFEFEF}NAM variants~\cite{changnode, ibrahim2024grand, dubey2022scalable, radenovic2022neural, jiao2023naisr, liu2023n, brendel2019approximating, tsang2018neural, xu2022sparse, liu2020sparse, bouchiat2023improving, enouen2022sparse, yang2021gami} \\
        \multicolumn{1}{c|}{\multirow{-2}{*}{\begin{tabular}[c]{@{}c@{}}Additive Score \\ Attribution~(Sec \ref{sec:additive_score})\end{tabular}}}                   & Shapley-based                                 & Efficiency, Linearity, Nullity, Symmetry                                                                                                              & SASANet~\cite{sun2023towards}, SHAPNet~\cite{wang2021shapley}                                                                                                                                                                                                           \\ \hline
        \multicolumn{2}{c|}{Hybrid Feature Attribution~(Sec \ref{sec:attribution_hybrid})}                                                                                                                            & \cellcolor[HTML]{EFEFEF}Coefficient Constraints, Additive Score Constraints                                                                           & \cellcolor[HTML]{EFEFEF}SSCN~\cite{sun2021market}                                                                                                                                                                                                                       \\ \hline
    \end{tabular}
    }
    \label{tab:attribution_comparison}
\end{table*}

\subsubsection{Generalized Coefficient Attribution}
\label{sec:generalized_coefficient}
The idea of using coefficients to interpret models is long-standing in statistics, with its roots in linear regression (LR) and generalized linear model (GLM). LR, expressed as $f(x) = \sum^N_{i=1} a_i x_i + a_0$, is one of the most interpretable modeling techniques, where the coefficients $a_i$ quantify the sensitivity of the output to each input feature. GLM extends LR by introducing link functions and probabilistic frameworks, enabling broader applications while retaining the interpretable structure. Moreover, coefficient interpretation can be generalized to neural networks by applying generalized coefficients to the feature layer. These coefficients act as proxies for feature re-weighting or selection and are learned by the model to quantify the importance of each feature in the prediction, as expressed:
\begin{equation}
    \label{equ:glm}
    \begin{aligned}
    f({x};\theta, \phi_0) & = \sum^N_{i=1} \alpha_i({x};\theta) {x}_i + \phi_0. \\
    {x} & = h(r), \ \text{s.t.} \ \bm{\alpha} \in \mathcal{C}.
    \end{aligned}
\end{equation}
In this context, ${x}_i \in \mathbb{R}^d$ represents the input features, which are aggregated in the high-dimensional space, and the coefficients $\alpha_i(x; \theta)$ quantify the attribution of the $i$-th feature.
Various constraints $\mathcal{C}$ can be applied to ensure meaningful attributions. Notably, the coefficients can be either global or local, depending on the model context. Global coefficients are shared across samples~\cite{guo2021edge, afchar2021towards, jiang2024protogate}, capturing general feature importance, while local coefficients, specific to each sample, capture instance-level feature importance~\cite{li2020mri, quinn2020deepcoda}. Following common practice in attribution research, we emphasize the local perspective, since the global formulation typically collapses into an ordinary linear model.

\textbf{Gradient-based Constraints.} SENN~\cite{alvarez2018towards} is a pioneering study to formulate self-attribution neural networks with generalized coefficients. SENN proposes to ensure that the coefficients $\alpha(x)$ mirror the sensitivity of the output $f(x)$ to the input ${x}$ by introducing the constraint set: $\mathcal{C}_{\mathrm{grad}}=\{\boldsymbol{\alpha}(x) \in \mathbb{R}^N: \boldsymbol{\alpha}(x) \approx \nabla_x f(x)\}$. Using the chain rule, SENN generalizes this gradient alignment to the Jacobian matrix of the feature transformation function $h(\cdot)$, denoted as $\nabla_x f(x) \approx \boldsymbol{\alpha}(r)^\top J^h_r(r)$.
Essentially, SENN utilizes a composition of local linear models to represent a complex nonlinearship, with each coefficient indicating the importance of a feature by its sensitivity to local disturbances\footnote{While subsequent works~\cite{quinn2020deepcoda, wang2021self, sun2023towards} have followed the SENN framework, they notably diverge from its original gradient-based constraints.}.

\textbf{Architectural Constraints.} The difficulty of feature attribution lies in the highly nonlinearity of neural networks. To address this, several studies propose dynamic linear models~\cite{bohle2021convolutional} that make predictions through input-dependent linear transformations. Each layer computes the output as $l(x) = \bm{\alpha}_i(x)x$, where $\bm{\alpha}_i(x)$ is input-varying and bounded. In this way, the constraint can be expressed as: $\mathcal{C}_{\mathrm{arc}}=\{\bm{\alpha} (x) \in \mathbb{R}^N: \bm{\alpha}(x) = \prod_{i=1}^{L} \bm{\alpha}_i(x), \ \text{and} \ \bm{\alpha}_i(x)= g({W_i}x + {b_i}) \}$, where $g(\cdot)$ is a nonlinear function (e.g., unit norm or squashing function), and ${W_i}, {b_i}$ are learnable parameters. This constraint ensures that $\bm{\alpha}(x)$ encodes the most frequent input patterns while maintaining attribution consistency across layers. B-cos Networks~\cite{bohle2024b, bohle2022b} extend this idea by introducing an exponent $B$ for flexible weight adjustment: 
$\bm{\alpha}_i(x)= |\cos\left(\angle(x, w_i)\right)|^B \times \operatorname{sgn}(\cos\left(\angle(x, w_i)\right))$, where $w_i$ is a learnable vector with unit norm. All these studies aim to capture network nonlinearity into the layer-wise coefficients in the linear form, allowing linear contribution mapping to the output.

\textbf{Attention-based Constraints.} Attention mechanisms have been widely used in neural networks to dynamically highlight the most informative features for the prediction. The inherent interpretability lies in its attention weights, which typically constrained to be nonnegative and to sum to one, ensuring that they form a interpretable distribution: $\mathcal{C}_{\text{att}} = \left\{ \boldsymbol{\alpha}(x) \in \mathbb{R}^N : \alpha_i(x) \geq 0, \ \sum_{i=1}^{N} \alpha_i = 1 \right\}$. In essence, attention weights are generally derived by applying a softmax function to a set of relevance scores. To enhance interpretability, alternatives such as Sparsemax~\cite{martins2016softmax} and entmax~\cite{correia2019adaptively} have been proposed to produce sparse probability distributions by assigning zero to less relevant features.
There are two main types of attention scoring functions:
(1) \textit{Additive attention}~\cite{BahdanauCB14} computes attention weights between the decoder hidden state \( s_t \) and each encoder hidden state \( h_i \) using a feedforward neural network:
$\alpha_i(x) = \text{softmax}\left( \boldsymbol{v}_a^{\top} \tanh(W_1 h_i + W_2 s_t) \right),$
where \( W_1 \), \( W_2 \), and \( \boldsymbol{v}_a \) are learnable parameters.
(2) \textit{Scaled dot-product attention}, used in Transformers~\cite{vaswani2017attention}, computes
$\boldsymbol{\alpha}(x) = \text{softmax}\left( \frac{K^\top Q}{\sqrt{m}} \right),$
where $Q$ and $K$ are derived from input $x$ via linear projections.
In practice, diverse attention mechanisms adapt to specific contexts, including cross-attention~\cite{zhang2022query, bai2023interpretable} with separate queries and keys, multi-head attention~\cite{zhu2024propagation, yeh2019interpretable, ren2022diversified} for multi-faceted learning, and hierarchical attention~\cite{luo2018beyond, wu2021evidence, xu2022towards, zhu2020modeling, arik2021tabnet} for capturing layered information.

Studies raise concerns about attention weights as explanatory mechanisms, highlighting their misalignment with actual feature importance~\cite{wiegreffe2019attention, serrano2019attention, meister2021sparse, bibal2022attention}. This discrepancy often stems from biased and noisy training data~\cite{bai2021attentions}. Brunner et al.\cite{brunner2019identifiability} address these issues by decomposing attention weights into ``effective attention" that better captures feature importance. This approach has inspired further developments by other researchers\cite{sun2021effective, kobayashi2020attention}. Additional solutions include SEAT~\cite{hu2023seat}, which stabilizes attention weights against perturbations, and Mohankumar's~\cite{mohankumar2020towards} diversity-driven approach with orthogonalization penalties. Recent work also explores human oversight through annotated rationales~\cite{nguyen2023learning, pruthi2020learning, stacey2022supervising}. In this direction, Rigotti et al.~\cite{rigotti2021attention} propose aligning model attention with human concepts through cross-attention mechanisms. These studies collectively aim to improve the faithfulness and interpretability of attention mechanisms.

\textbf{Lasso-based Constraints.} 
The lasso regularization, also known as $\ell_1$ norm, plays a fundamental role in achieving feature sparsity through coefficient regularization. When applied to model parameters, it effectively drives some coefficients to zero, naturally performing feature selection. A common approach is to directly impose $\ell_1$ norm on the input features~\cite{quinn2020deepcoda, li2019exploiting, lei2016rationalizing}, which can be formulated as: $\mathcal{C}_{\mathrm{lasso}}=\{\boldsymbol{\alpha}(x) \in \mathbb{R}^N: \|\boldsymbol{\alpha}(x)\|_1 \leq \lambda\}$, where $\lambda$ is a hyperparameter. However, in the context of neural networks, this straightforward approach suffers from several interpretability issues, such as (1) \textit{hierarchy problem}: due to multi-layer nonlinearity, a feature with a minuscule weight might still impact the output significantly, and (2) \textit{stability}: the selected features may vary across similar samples. To address these issues, recent studies propose to preserve feature relationship and stability across the whole networks. For example, LassoNet~\cite{lemhadri2021lassonet} adopts a residual structure and impose hierarchical sparsity constraints to ensure direct and consistent feature selection across layers. Several lasso variants are also proposed to adjust the penalization degree of features, such as weighted lasso~\cite{he2019interpretable}, truncated lasso~\cite{zhang2025comprehensive}, contextual lasso~\cite{thompson2023contextual}, and group lasso~\cite{dinh2020consistent}.

\textbf{Subset Sampling Constraints.} These constraints, also known as $\ell_0$ norm, enhance interpretability by restricting the model to a discrete subset of features. A common formulation is to enforce $\boldsymbol{\alpha}_i(x) \in\{0,1\}$ and $\sum_{i=1}^N \boldsymbol{\alpha}_i(x) \leq K$, allowing at most $K$ features to be selected~\cite{yoon2018invase}. Alternatively, one can impose $\sum_{i=1}^N \boldsymbol{\alpha}_i(x)=K$ to sample exactly $K$ features~\cite{chen2018learning}. Comparing to $\ell_1$ norm, the subset sampling constraints naturally lead to discrete inference process and avoid the hierarchy problem. However, it is computationally intractable and not differentiable for neural networks, which makes the optimization problem particularly challenging. A common approach is to activate each feature with a probability, expressed as: $\bm{\alpha}(x) \sim P(x)$. In practice, $P(x)$ in SINNs could be modelled an independent Bernoulli distribution~\cite{yoon2018invase, jethani2021have, jiang2024protogate} or a parameterized distribution for stochastic gates ~\cite{yang2022locally, yamada2020feature, lee2022self, li2020mri}. 
Another method is to use a selection limit $K$ that sets the number of samples taken from a concrete distribution, assigning a hard bound on the number of features selected. Reparameterization gradients, such as Gumbel-softmax~\cite{jang2017categorical, xie2019reparameterizable}, are often used to optimize the model, allowing end-to-end training. This method is widely used in text ranking~\cite{leonhardt2023extractive}, salient mapping~\cite{chen2018learning, balin2019concrete}, and tabular data analysis~\cite{si2024interpretabnet}.

\textbf{Information-theoretic Constraints.} The Information Bottleneck (IB) principle is a commonly used constraint to guide the learned attributions $\bm{\alpha}(x)$ toward a compressed yet predictive representation. Formally, one enforces $\mathcal{C} = \left\{ \bm{\alpha}(x) : \text{maximize} \ I(\bm{\alpha}(x) x; y) - \beta I(\bm{\alpha}(x) x; x) \right\},$ where $I(\cdot;\cdot)$ is the mutual information, and $\beta$ balances prediction fidelity versus input compression. $\bm{\alpha}(x)$ could be continuous or discrete, depending on the application. To optimize this objective, it typically introduces a variational lower bound of the IB objective~\cite{alemi2022deep} for approximation. Comparing to subset sampling constraints that explicitly choose $K$ features, information-theoretic methods primarily regulate how ``much'' information is retained. It ensures a concise yet informative representation that captures the essential decision-making information. This approach has been widely used in various domains, such as graph bottleneck for efficient subgraph recognition~\cite{yu2020graph, yu2021recognizing, seo2024interpretable, fang2024exgc, liu2024towards, chen2024tempme, miao2022interpretable}, word extraction~\cite{sekhon2023improving}, and geometric learning~\cite{miaointerpretable2023}.

\subsubsection{Additive Score Attribution}
\label{sec:additive_score}
Comparing to learning interpretable coefficients that modulate feature importance, another approach directly learns transformation functions that map features to their contributions. 
This distinction is fundamental: instead of separating feature importance (coefficients) from the predictor, additive score attribution unifies them into a single step through feature-specific scoring functions. Formally, the additive score attribution model can be expressed as:
\begin{equation}
    f({x}; \theta, \phi_0) = \sum_{i=1}^N g_i(x; \theta_i) + \phi_0, \quad \text{s.t.} \ \bm{g} \in \mathcal{C}.
    \label{eq:ASA-first-order}
\end{equation}
Here $g_i(x; \theta_i): \mathbb{R}^{N \times d} \to \mathbb{R}$ represents the contribution of the $i$-th feature, and $\phi_0$ is a global bias. Based on the constraints $\mathcal{C}$, we categorize additive score attribution into two groups: \textit{neural additive model} and \textit{Shapley-based attribution}.

\textbf{Neural Additive Model.}
NAM~\cite{agarwal2021neural} is a fundamental study that implements the additive score attribution model in neural networks. NAM follows Generalized Additive Models~\cite{hastie2017generalized}, which decomposes the prediction into a sum of univariate functions. Specifically, NAM has two key properties: 

\begin{enumerate}
    \item \textit{Uniqueness of feature dependence}:
    \[
        \forall {x}, {x}' \in \mathbb{R}^N,\quad
        x_i = x'_i \;\;\implies\;\;
        g_i({x}; \theta_i) = g_i({x}'; \theta_i).
    \]
    This ensures that the contribution of each feature is uniquely determined by its value.
    
    \item \textit{Additional interpretability properties}: 
    one might require monotonicity, or smoothness constraints. For instance,
    \[
        \int g_i(x_i)h_v(x_v)w(x)dx = 0, \quad \forall v \subset i, \ \forall h \in L^2(\mathbb{R}^v)
    \]
    This ensures the pureness of the decomposition~\cite{sun2022puregam}, where each component function $g_i$ is orthogonal to other sub-functions. $w(x)$ represents the probability density.
\end{enumerate}

Rather than directly using traditional MLPs, NAM~\cite{agarwal2021neural} designs ExU (exp-centered) units to learn jumpy, non-smooth functions in logarithmic space: $h(x) = f(e^w * (x - b))$. This helps capture sharp changes in feature effects that standard neural nets struggle with. Following this line, various studies have proposed extensions to NAM to improve the expressiveness and efficiency of function designs. For example, NODE-GA$^2$M~\cite{changnode} employs the expressiveness of neural oblivious decision ensembles, while GRAND-SLAMIN~\cite{ibrahim2024grand} introduces soft trees based on hyperplane splits. Some approaches focus on tensor-based decompositions to decompose higher-order interactions into a set of low-rank tensors for learning efficiency, such as SPAM~\cite{dubey2022scalable}, CAT~\cite{duong2024cat}, and NBM~\cite{radenovic2022neural}. It also inspired interpretable applications in a range of domains: NAISR~\cite{jiao2023naisr} for 3D shape representation, N$\text{A}^\text{2}$Q~\cite{liu2023n} for coordinating collaborative behaviors, and~\cite{brendel2019approximating} for approximating CNNs.

While \Cref{eq:ASA-first-order} focuses on unary (first-order) contributions, many real-world tasks require higher-order feature interactions, which can be interpreted as new, interaction-based features.
In this research line, NAM can be expanded to the $N$-th order, which is expressed as:
\begin{equation}
    f(\boldsymbol{x})=\sum_{i=1}^N \underbrace{g_i\left(x_i\right)}_{\text {order } 1 \text { (unary) }}+\sum_{j>i}^N \underbrace{g_{i j}\left(x_i, x_j\right)}_{\text {order } 2 \text { (pairwise) }}+\cdots+\underbrace{g_{1 \ldots N}(x)}_{\text {order } N},
\end{equation}
where $g_i(\cdot)$ captures the contribution of the $i$-th feature, $g_{ij}(\cdot)$ captures the pairwise interaction between features $i$ and $j$, and so forth, culminating in a potential $N$-way interaction term $g_{1...N}(\cdot)$. Models restricted to the up-to-$k$-th order interactions are often called NA$^k$M, which balances the trade-off between model complexity and interpretability. 
While higher-order interactions can improve predictive accuracy, they also lead to exponential growth in model complexity and reduced interpretability. A common strategy to address this challenge is to impose sparsity, which is aligned with feature selection methods. One line of work enforces sparsity through regularization constraints, for instance, using disentangled group regularizers~\cite{tsang2018neural}, group LASSO~\cite{xu2022sparse}, data-dependent kernel expansions~\cite{liu2020sparse} or adaptive log-marginal likelihood regularization~\cite{bouchiat2023improving}. 
Another research line focuses on structural constraints to achieve sparsity. NODE-GA$^2$M~\cite{changnode} utilizes sparse Entmax transformation to produce a one-hot vector, forcing each tree to select only one specific feature. GRAND-SLAMIN~\cite{ibrahim2024grand} proposes a sparse back-propagation approach by progressively suppressing uninformative components through the Smooth-Step function. 
Finally, some approaches perform post-training pruning: SIAN~\cite{enouen2022sparse} employs a specialized selection algorithm, while GAMI-Net~\cite{yang2021gami} prunes according to the weak heredity principle. 
In particular, such post-training pruning methods can be computationally more efficient for higher-order expansions and can readily adapt to existing models without retraining.

\textbf{Shapley-Based Attribution.} 
Shapley values, rooted in coalition game theory~\cite{shapley1953value}, have emerged as a popular way to measure the contribution of individual features (or players) in a predictive model (or coalition). They satisfy four key axioms: (1) \textit{Efficiency}: the sum of individual attributions equals the total model output; (2) \textit{Linearity}: Shapley values respect linear combinations of models; (3) \textit{Nullity}: a feature with no effect on the output receives zero attribution; and (4) \textit{Symmetry}: two features with identical effects on the model receive identical attributions. While most Shapley-based explanations are post-hoc interpretation~\cite{scott2017unified, bento2021timeshap, jiang2024seqshap}, recent research has sought to embed Shapley values directly into model architectures to encourage unified self-attributing behavior~\cite{wang2021shapley, sun2023towards}.

SHAPNet~\cite{wang2021shapley} incorporates Shapley values as latent representations in deep neural networks. It starts from a simplified notion of ``presence'' and ``absence" of features. Let $z \in \{0,1\}^N$ be a binary indicator for the features used in a model input. A baseline or reference vector $r$ represents the ``absence" of a feature, giving the following mapping: $\Psi_{x,r}(z) = z \odot x + (1-z) \odot r.$ It approximates the Shapley value $g_i(x,f)$ for feature $i$ by summing over all $z$ subsets that do not include $i$:
\begin{equation}
    g_i(x,f) \triangleq \frac{1}{N} \sum_{z \in Z_i} w(z)[f(\Psi_{x,r}(z \cup \{i\})) - f(\Psi_{x,r}(z))],
\end{equation}
where $w(z) = \binom{N-1}{\|z\|_1}^{-1}$. Unlike post-hoc methods, SHAPNet encodes this Shapley mechanism within each network layer as a \emph{Shapley transform} module, providing \emph{layer-wise} attributions. 

SASANet~\cite{sun2023towards} achieves model-wise Shapley attribution by introducing a \emph{set-based} modeling strategy.
Instead of toggling features on or off against a baseline, SASANet formulates its model $f(x) = \sum_{i=1}^N g_i(x;\theta) + g_0$, so that each $g_i(x;\theta)$ corresponds to the exact Shapley marginal contribution of feature $i$ across \emph{all} possible subsets and orders:
\begin{equation}
    g_i(x;\theta) = \frac{1}{N!} \ \ \sum_{\mathclap{O \in \pi(N)}} \ \ \sum_{k=1}^N \mathbb{I}\{O_k = i\}\triangle(x_i, {x}_{O_{1: k-1}};\theta),
\end{equation}
where $\pi(N)$ is the set of all permutations of $N$, and $\triangle(x_i, x_{O_{1: k-1}})= f({x}_{O_{1: k-1} \cup\{i\}} ; \theta)-f({x}_{O_{1: k-1}} ; \theta)$. A crucial aspect is that SASANet avoids approximation by training an internal \emph{positional distillation} mechanism that converges to the true Shapley value of its own output. This eliminates the need for a handcrafted baseline $r$ and provides a \emph{model-wide} Shapley attribution rather than layer-specific ones.

\subsubsection{Hybrid Feature Attribution}
\label{sec:attribution_hybrid}
To capture richer feature information, a unified framework has been proposed that jointly models coefficient and additive score attribution~\cite{swamy2025intrinsic, sun2021market}. This hybrid approach combines the feature importance weighting of coefficient attribution with non-linear, feature-specific transformations. The general form is expressed as:
\begin{equation}
    \label{equ:jfa}
    f({x}; \theta, g_0) = \sum_{i=1}^N \alpha_i({x};\theta_i) g_i(x; \theta_i) + g_0, \quad \text{s.t.} \ \bm{\alpha}, \bm{g} \in \mathcal{C},
\end{equation}
where $\alpha_i(x;\theta_i)$ denotes the context-dependent coefficient for the $i$-th feature, and $g_i(x; \theta_i)$ is the feature-specific scoring function. This formulation enables a more expressive attribution model, capturing both the global importance of features via coefficients and their individual non-linear contributions through the scoring functions.

SSCN~\cite{sun2021market} is a representative model that adopts this hybrid feature attribution framework. SSCN is a domain-specific model aiming at capturing the dynamic influence of skills on salary under various contexts, which models job salary as the weighted average of required skills' values, where $\alpha_i(x; \theta_i)$ denotes the skill's relative importance in a specific job context and $g_i(x;\theta_i)$ denotes the skill's context-aware value. Regarding each skill as a feature for salary prediction, SSCN intrinsically aligns with Eq.~\ref{equ:jfa}, subject to constraints: (1) uniqueness of feature dependence: $\forall \, x, x', x_i = x'_i \rightarrow g_i(x; \theta_i) = g_i(x'; \theta_i)$, and (2) normalized coefficients: $\sum_i \alpha_i(x; \theta_i) = 1$. 
To realize such constraints, SSCN models $g_i(x; \theta_i)$ with a single-skill network that captures contextual interactions, which is similar to the idea of NAM, and models $\alpha_i(x; \theta_i)$ with a graph-based attention mechanism that accounts for skill co-occurrence patterns. This enables SSCN to analyze both global skill values and their local weights in specific job contexts, offering insights into how skills influence salary across different contexts.

\subsection{Function-based Self-Interpretation}
\label{sec:function}
Function-based SINNs aim to render neural computations interpretable through structured mathematical forms. 
In this subsection, we discuss two key paradigms: (1) structured compositions of univariate functions (functional decomposition) and (2) multivariate symbolic equations (equation learning), which aim to discover explicit mathematical formulas. \Cref{tab:function} summarizes the key principles and representative works.

\subsubsection{Functional Decomposition}
\label{sec:decomposition}
Functional decomposition provides a principled way to break down complex neural networks into interpretable sub-functions, each focusing on specific relationships among input features and hidden concepts. Formally, a neural network can be expressed as \begin{equation} f(\bm{x}) := \bigl(\Phi_{L} \circ \Phi_{L-1} \circ \cdots \circ \Phi_{2} \circ \Phi_{1}\bigr) \bm{x}, \end{equation} where $\Phi_{i}$ denotes the transformation (e.g., layer) at stage $i$. Unlike standard MLPs, functional decomposition replaces linear transformations with expressive yet interpretable nonlinear functions This creates a powerful network using shallow and narrow function transformations, making it transparent.

VOTEN~\cite{sun2021discerning} implements functional decomposition via a hierarchical ``voting'' strategy. Each layer constructs higher-level concepts by applying univariate nonlinear transformations to the current variables, then aggregating them with weights that sum to one. Formally, the $j$-th concept in layer $l+1$ is computed as:
\begin{equation}
x_j^{(l+1)} = \sum_{k=1}^{n_l} a_{k,j}^{(l)}f_{k,j}^{(l)}(x_k^{(l)}) \quad \text{s.t.} \quad \sum_{k=1}^{n_l} a_{k,j}^{(l)} = 1,
\end{equation}
where $x_j^{(l+1)}$ is the $j$-th concept in layer $l+1$, $a_{k,j}^{(l)}$ are the voting weights, and $n_l$ denotes the number of concepts in the $l$-th layer. By enforcing normalized weights and single-valued transformations, VOTEN's design reveals how each feature evolves into more abstract concepts at successive layers.

Similar to VOTEN, KAN~\cite{liu2024kan} also replaces standard neural connections with learnable univariate functions. Formally, the activation value of the $j$-th neuron in layer $l+1$ is simply summed over all neurons in layer $l$:
\begin{equation}
x_j^{(l+1)} = \sum_{k=1}^{n_l} \phi_{k,j}^{(l)}(x_k^{(l)}),
\end{equation}
where $\phi_{k,j}^{(l)}$ represents the univariate function connecting the $k$-th neuron in layer $l$ to the $j$-th neuron in layer $l+1$. The key difference between VOTEN and KAN in modeling the univariate function is that VOTEN uses a univariate neural network, whereas KAN employs a spline interpolation method. In KAN, each $\phi_{k,j}^{(l)}$ is parameterized as:
\begin{equation}
    \phi_{k,j}^{(l)}(x) = w_b b(x) + w_s \sum_{i} c_i B_i(x),
\end{equation}
where $b(x)$ is a base function (e.g., silu activation, $\text{silu}(x) = x/(1+e^{-x})$), $B_i(x)$ are B-spline basis functions, and $w_b$, $w_s$, $c_i$ are learnable parameters. The spline parameterization allows for flexible function learning while preserving the ability to visualize and understand each transformation's behavior.

Despite different implementations, both VOTEN and KAN share the core objective of disentangling a high-dimensional function into interpretable sub-functions. The Kolmogorov-Arnold representation theorem provides theoretical guarantee for the expressiveness of such network structures. 
To further optimize the transformation structure, both methods begin with dense transformations and use pruning strategies to remove redundant components, highlighting the most salient sub-functions. Along this direction, automating network design and progressive growing techniques can be explored to simplify interpretable functional decomposition models.

\begin{table}[t]
    \centering
    \caption{Function-based Self-Interpretation Methods}
    \vspace{-1mm}
    \resizebox{\linewidth}{!}{%
    \begin{tabular}{c|c|c}
        \toprule
        Methods                            & Principles                                       & Representative Works                                                                                                                     \\ \midrule
        Functional                         & Break down NNs into                              & \multirow{2}{*}{VOTEN~\cite{sun2021discerning}, KAN~\cite{liu2024kan}}                                                                   \\
        \multicolumn{1}{l|}{Decomposition} & \multicolumn{1}{l|}{interpretable sub-functions} &                                                                                                                                          \\ \midrule
        Equation                           & Directly                                         & Architectural: \cite{kim2020integration, ranasinghe2024ginn, kim2020integration}                 \\
        Learning                           & Learning Equations                               & Generative\cite{kamienny2022end, bendinelli2023controllable, biggio2021neural, valipour2021symbolicgpt, landajuela2022unified, holtdeep} \\ \bottomrule
        \end{tabular}
        }
    \label{tab:function}
    \vspace{-2mm}
\end{table}

\subsubsection{Equation Learning}
\label{sec:equation}
Compared to the functional decomposition paradigm, equation learning focuses on directly learning concise, generalizable, and interpretable mathematical equations from data. It provides a more straightforward and global view on model interpretability, as the learned equations can be directly inspected and analyzed by domain experts. Equation learning has shown particular success in discovering physical laws and scientific relationships. Typically, equation-learning optimization target is formulated as:
\begin{equation}
f^*(\bm{x}) = \arg\min_{f \in \mathcal{F}} \frac{1}{|\mathcal{D}|} \sum_{i=1}^{|\mathcal{D}|} \text{loss}(f(\bm{x}_i), y_i),
\end{equation}
where $\mathcal{F}$ denotes the space of interpretable functions (such as polynomials, trigonometric functions, and their combinations), $\mathcal{D}$ is the training dataset consisting of input-output pairs, and $\text{loss}(\cdot)$ is typically similar regression losses. In essence, this is a constrained optimization task aimed at identifying the most interpretable function $f^*(\cdot)$ that fits the data well while maintaining mathematical simplicity and physical plausibility. In the context of SINNs, we introduce two categories of methods that learn the desired function through neural networks.

\textbf{Architecture-based methods} typically rely on multi-layer feed-forward networks that incorporate specialized units reflecting basic algebraic operators~\cite{kim2020integration, sahoo2018learning, ranasinghe2024ginn}. These units often include unary operations (e.g., identity mapping $f(x)=x$, trigonometric functions sin(x) and cos(x), exponential functions $e^x$), binary operations (e.g., multiplication $x_1 \cdot x_2$, division ${x_1}/{x_2}$)~\cite{sahoo2018learning}, and linear transformations implemented through weight matrices.
For example, given input multivariates and corresponding scalar output, GINN-LP\cite{ranasinghe2024ginn} uses a parallel structure of Power-Term Approximator (PTA) blocks to learn multivariate Laurent polynomials $P = \sum_{i=1}^n c_i \prod_{j=1}^k V_j^{p_i(j)}$ with real coefficients $c_i$ and integer powers $p_i(j)$. 
Each PTA block applies logarithmic transformation, linear combination, and exponential activation to capture a polynomial term. 
These terms are then combined through a final linear layer, enabling the discovery of physics equations like Coulomb's law $F = \frac{1}{4\pi\epsilon_0}\frac{q_1q_2}{r^2}$ and kinetic energy $E = \frac{1}{2}mv^2$.
A key limitation of these models is their reliance on pre-defined structures, which can result in unnecessarily complex equations and reduced flexibility in capturing data relationships.
Therefore, these approaches employ sparsity-inducing constraints~\cite{kim2020integration} (like $\ell_0$, $\ell_1$  regularization term), systematic architecture selection~\cite{sahoo2018learning} through validation performance, or adaptive network growth strategies~\cite{ranasinghe2024ginn} to balance conciseness with predictive expressiveness. For instance, GINN-LP employs an iterative growth strategy where the optimization objective combines prediction error and model complexity: $L = \frac{1}{n}\sum_{i=1}^n (f(x_i) - y_i)^2 + \lambda_1\sum_{i=1}^n |W_i| + \lambda_2\sum_{i=1}^n W_i^2$, where the last two terms promote sparsity and prevent overfitting by penalizing activated weights $W_i$.

\textbf{Generation-based methods} model equation learning as a \emph{set-to-sequence} task, wherein a set of numerical data points is transformed into a sequence of symbolic tokens~\cite{kamienny2022end, bendinelli2023controllable, biggio2021neural}. 
These methods aim to generate symbolic expressions based on the learned data distribution, enabling discovering interpretable equations without pre-defined structures.
Architecturally, they typically employ an encoder-decoder architecture where the encoder captures data dependencies and the decoder autoregressively generates symbolic expressions.
Two predominant workflows are employed in this domain. 
The first is to predict a symbolic skeleton (e.g., the equation structure) and then fit the constants~\cite{biggio2021neural, valipour2021symbolicgpt, landajuela2022unified}. For example, if $f = 2.21\sin(2x_1) + 7.21\exp(x_2)$, the model first predicts the skeleton $f = a\sin(bx_1) + c\exp(dx_2)$ and then fits the unknown constants. The second approach is to simultaneously predict the model structure and constants~\cite{kamienny2022end, bendinelli2023controllable, holtdeep}, generating tokens that represent both operators and numerical values in a unified process. Recent studies reveal that these models benefit from pre-training on large corpora of symbolic equations, thereby improving generalization~\cite{valipour2021symbolicgpt, landajuela2022unified, biggio2021neural}.

\begin{table*}[t]
    \centering
    \caption{Summary of representative concept-based self-interpretation methods. ``S/U'' indicates supervised/unsupervised concept labeling, while ``PTM'' refers to pre-trained models. ``MLM'' stands for BERT-based masked language models, and ``LLM'' and ``VLM'' refer to large language and vision language models, respectively. Papers are organized by publication year.}
    \resizebox{\linewidth}{!}{%
    \begin{tabular}{c|ccc|cc}
        \hline
        \multirow{2}{*}{Methods}                                                     & \multicolumn{3}{c|}{Concept}                                & \multirow{2}{*}{Human Intervention}                                                           & \multirow{2}{*}{\begin{tabular}[c]{@{}c@{}}Required\\ Architectures\end{tabular}} \\ \cline{2-4}
                                                                                     & Organization                 & Representation & Supervision &                                                                                               &                                                                                   \\ \hline
        \rowcolor[HTML]{EFEFEF} CBM~\cite{koh2020concept}                            & Flat                         & Scalar         & S           & Test-time                                                                                     & -                                                                                 \\
        \rowcolor[HTML]{FFFFFF} Concept Whitening~\cite{chen2020concept}             & Flat                         & Embedding      & S           & -                                                                                             & -                                                                                 \\
        \rowcolor[HTML]{EFEFEF} ENN~\cite{blazek2021explainable}                     & Hierarchical                 & Scalar      & S           & -                                                                                             & -                                                                                 \\
        \rowcolor[HTML]{FFFFFF} ConceptTransformer~\cite{rigotti2021attention}       & Flat                         & Embedding      & S           & Test-time                                                                                     & Transformer                                                                       \\
        \rowcolor[HTML]{EFEFEF} CEM~\cite{espinosa2022concept}                       & Flat                         & Embedding      & S           & Test-time                                                                                     & -                                                                                 \\
        \rowcolor[HTML]{FFFFFF} PCBM~\cite{yuksekgonul2022post}                      & Flat                         & Similarity     & S           & Test-time, Global-aware                                                                       & PTM                                                                               \\
        \rowcolor[HTML]{EFEFEF} Glancenets~\cite{marconato2022glancenets}            & Flat                         & Embedding      & S, U        & Test-time                                                                                     & VAE                                                                               \\
        \rowcolor[HTML]{FFFFFF} ProbCBM~\cite{kim2023probabilistic}                  & Flat                         & Probalistic    & S           & Test-time                                                                                     & -                                                                                 \\
        \rowcolor[HTML]{EFEFEF} Side-channel CBM~\cite{havasi2022addressing}         & Flat + Side-channel          & Probalistic    & S, U        & Test-time                                                                                     & -                                                                                 \\
        \rowcolor[HTML]{FFFFFF} CB2M~\cite{steinmann2023learning}                    & Flat + 2-fold Memory         & Scalar         & S           & Test-time, Memory-guided                                                                      & -                                                                                 \\
        \rowcolor[HTML]{EFEFEF} Interactive CBM~\cite{chauhan2023interactive}        & Flat                         & Scalar         & S           & Test-time, Interactive                                                                        & -                                                                                 \\
        \rowcolor[HTML]{FFFFFF} Label-free CBM~\cite{oikarinen2023label}             & Flat                         & Similarity     & U           & Test-time                                                                                     & LLM, VLM                                                                          \\
        \rowcolor[HTML]{EFEFEF} Labo~\cite{yang2023language}                         & Flat                         & Similarity     & U           & Test-time                                                                                     & LLM, VLM                                                                          \\
        \rowcolor[HTML]{FFFFFF} Coarse-to-Fine CBM~\cite{panousis2024coarse}         & Coarse-to-Fine               & Similarity     & S, U        & Test-time                                                                                     & VLM                                                                               \\
        \rowcolor[HTML]{EFEFEF} Res-CBM~\cite{shang2024incremental}                  & Flat + Residual Bank         & Similarity     & S, U        & Test-time                                                                                     & VLM                                                                               \\
        \rowcolor[HTML]{FFFFFF} IntCEM~\cite{espinosa2024learning}                   & Flat                         & Embedding      & S           & Test-time, Policy-guided                                                                      & -                                                                                 \\
        \rowcolor[HTML]{EFEFEF} Energy-based CBM~\cite{xu2024energy}                 & Flat                         & Probalistic    & S           & Test-time, Dependency-aware                                                                   & PTM                                                                               \\
        \rowcolor[HTML]{FFFFFF} Stochastic CBM~\cite{vandenhirtz2024stochastic}      & Flat                         & Probalistic    & S           & \begin{tabular}[c]{@{}c@{}}Test-time, Dependency-aware \end{tabular} & -                                                                                 \\
        \rowcolor[HTML]{EFEFEF} Relational CBM~\cite{barbierorelational}             & Hypergraph                   & Embedding      & S           & Test-time, Rule-guided                                                                         & -                                                                                 \\
        \rowcolor[HTML]{FFFFFF} SparseCBM~\cite{tan2024sparsity}                     & Flat                         & Scalar         & S           & Test-time                                                                                     & LLM                                                                               \\
        \rowcolor[HTML]{EFEFEF} MICA~\cite{bie2024mica}                              & Hierarchical                 & Embedding      & S           & Test-time                                                                                     & -                                                                                 \\
        \rowcolor[HTML]{FFFFFF} Counterfactual CBM~\cite{dominici2024counterfactual} & Flat                         & Scalar         & S           & Test-time, Counterfactual                                                                     & VAE                                                                               \\
        \rowcolor[HTML]{EFEFEF} Causal CGM~\cite{dominici2024causal}                 & Directed Acyclic Graph       & Embedding      & S           & Test-time, Counterfactual                                                                     & -                                                                                 \\
        \rowcolor[HTML]{FFFFFF} CB-LLM~\cite{sun2024concept}                         & \cellcolor[HTML]{FFFFFF}Flat & Similarity     & U           & Test-time                                                                                     & LLM                                                                               \\
        \rowcolor[HTML]{EFEFEF} CB-pLM~\cite{ismail2024concept}                      & Flat + Unknown Parts         & Scalar         & S, U        & Test-time                                                                                     & MLM                                                                               \\
        \rowcolor[HTML]{FFFFFF} VisCoIN~\cite{parekh2024restyling}                   & Flat                         & Scalar         & U           & Test-time                                                                                     & PTM                                                                               \\ \hline
    \end{tabular}
    }
    \label{tab:concept}
    \vspace*{-1mm}
\end{table*}

\subsection{Concept-based Self-Interpretation}
\label{sec:concept}
Concept-based methods aim to enhance model interpretability by structuring learned representations around human-understandable concepts. 
In this way, these models enable understanding on both the presence of specific concepts in raw inputs and their influence on the model's predictions.

Despite different architectural designs, concept-based models generally has a two-stage structure: 
(1) mapping raw inputs to interpretable concepts, and (2) leveraging these concepts for final predictions. The core idea is to design a specialized ``bottleneck'' layer that represents human-interpretable concepts, which can be pre-defined, learned from data, or a combination of both. 
Formally, concept models predict a target $y \in \mathbb{R}$ from input $x \in \mathbb{R}^d$ through intermediate concepts $c \in \mathbb{R}^k$ using a composite function $f(g(x))$:
\begin{enumerate}[noitemsep, left=0pt]
    \item $g(\cdot): \mathbb{R}^d \rightarrow \mathbb{R}^k$, mapping raw inputs $x$ into the representation restricted to a concept space $c$.
    \item $f(\cdot): \mathbb{R}^k \rightarrow \mathbb{R}$, mapping the concept-based representation $c$ to the target $y$, typically through a linear or logic layer.
\end{enumerate}
The model minimizes loss functions $L_{C}$ for concepts and $L_Y$ for the target, supporting three training schemes~\cite{koh2020concept, espinosa2022concept}: \textit{independent}, where $f(\cdot)$ and $g(\cdot)$ are learned separately; \textit{sequential}, where $g(\cdot)$ is trained first followed by $f(\cdot)$; and \textit{joint} training, optimizing both functions simultaneously with a balancing hyperparameter $\lambda$. 
Most concept models also allows users to directly manipulate the predicted concepts at test time and observe how these changes impact the final prediction. This facilitates counterfactual reasoning (``What if the model didn't think this feature was present?''), and allows users to correct potential model errors~\cite{dominici2024counterfactual, dominici2024causal, ismail2024concept}.

Recent theoretical analysis by Luyten et al.~\cite{luyten2024theoretical} reveals that concept model performance depends on two critical factors: \textit{concept expressiveness} (the capacity to capture salient information) and \textit{model-aware inductive bias} (coherence between concept set and model architecture). This is also evidenced by the model's reliance on concept annotations, which can be costly and incomplete. Therefore, recent works have focused on enhancing the performance and scalability by designing a more expressive and model-aware concept layer.
We categorize these into three key directions: \textit{concept representation}, \textit{concept organization} and \textit{concept supervision}. \Cref{tab:concept} summarizes representative concept-based self-interpretation methods, highlighting their concept layer designs and human intervention strategies.

\subsubsection{Concept Representation}
In concept-based models, concept representation is crucial for regularizing the model's latent space, thereby shaping both interpretability and performance. The Concept Bottleneck Model (CBM)\cite{koh2020concept} employs direct scalar activation, mapping inputs to scalar concept scores via an activation function $s(\cdot): \mathbb{R} \rightarrow [0,1]$. This mapping can be hard, applying thresholding $(s(x) = \mathbbm{1}_{x>0.5})$, or soft, using a sigmoid activation function $(s(x) = 1/(1+e^{-x}))$. ENN~\cite{blazek2021explainable} refines concept mapping with \textit{differentia} neurons that distinguish sub-concept pairs and \textit{sub-concept neurons} that aggregate these distinctions. Despite their intuitive interpretability, they often struggle to capture the nuanced semantics of complex concepts.

To enhance expressiveness, the Concept Embedding Model (CEM)\cite{espinosa2022concept} and IntCEM\cite{espinosa2024learning} extend representations into high-dimensional embeddings, where each concept $c_i$ comprises active and inactive states $(\hat{c}^+_i, \hat{c}^-_i \in \mathbb{R}^m)$, later mapped by a sigmoid to determine concept presence. Concept Whitening~\cite{chen2020concept} aligns pre-defined concepts with latent axes via whitening and orthogonal transformations. Meanwhile, approaches such as PCBM~\cite{yuksekgonul2022post} and Concept Transformer~\cite{rigotti2021attention} adopt a geometric perspective, using similarity scores to project input embeddings onto concept subspaces. Besides, recognizing the inherent ambiguity of real-world concepts, recent studies~\cite{kim2023probabilistic, havasi2022addressing, xu2024energy, vandenhirtz2024stochastic} propose to model concept distributions, for example, Gaussian distributions $p(z_c \mid x) \sim \mathcal{N}(\mu_c, \text{diag}(\Sigma_c))$. This probabilistic framework naturally captures uncertainty and partial presence.

\subsubsection{Concept Organization}
Vanilla CBMs employ a flat structure with independent concepts, which can be difficult to capture the complex interdependencies inherent in human conceptual understanding. As shown by Raman et al.~\cite{raman2024understanding}, such assumptions of independence can lead to instability and robustness issues in concept models. Therefore, various approaches propose to organize concepts and model relationships between concepts, each offering distinct perspectives on capturing concept dependencies and tailoring relationship modeling to different contexts.

One perspective on concept organization focuses on complementary information channels~\cite{shang2024incremental, havasi2022addressing, steinmann2023learning}, which provide additional insights beyond directly encoded concepts. For instance, Side-channel CBM~\cite{havasi2022addressing} uses an auto-regressive architecture to incorporate both input features and previously predicted concepts in its predictions. CB2M~\cite{steinmann2023learning} extends this by incorporating memory systems to generalize corrections made from past mistakes, enabling the model to reapply user corrections in future predictions. Res-CBM~\cite{shang2024incremental} uses optimizable vectors to address missing concepts, drawing from a candidate concept bank and a base concept bank to identify and integrate the missing information.

Hierarchical organization is an effective way to structure concepts, particularly in complex domains like medical diagnosis and scene understanding. It organizes concepts across different levels, where higher-level concepts guide the interpretation of lower-level ones. 
For example, in Coarse-to-Fine CBM~\cite{panousis2024coarse}, the high-level concept (e.g., ``Arctic Fox") provides an overarching label, while low-level concepts describe specific details like body parts or background. This hierarchy ensures that only relevant low-level concepts are activated based on the high-level context. In MICA~\cite{bie2024mica}, multi-level alignment links image regions to high-level clinical concepts. ENN~\cite{blazek2021explainable} introduces differentia, sub-concept, and concept neurons to progressively refine input distinctions, forming a hierarchical structure for concept interpretation.

Additionally, graph-based structures offer a flexible way to model complex relationships between concepts.
Relational CBM (R-CBM)~\cite{barbierorelational} uses directed hypergraphs to represent concept dependencies, with hyperedges defining relational concept bottlenecks. This enables R-CBM to capture interactions between multiple entities, refining concept predictions through message passing and aggregation. In contrast, Causal CGM~\cite{dominici2024causal} focuses on causal relationships and uses directed acyclic graphs (DAGs) to model cause-effect chains between concepts. By explicitly representing causal structures, Causal CGM supports interventions and counterfactual reasoning. 
These diverse perspectives on concept organization offer a rich landscape for structuring concept relationships.

\subsubsection{Concept Supervision}
Various methods has been proposed to define and supervise the learning of concepts, striking a balance between model interpretability and scalable concept learning. 
Vanilla CBMs employ human-annotated concept supervision, offering direct interpretability but requiring substantial annotation effort and potentially missing task-relevant information. To address these limitations, several unsupervised approaches have been developed for automatic concept discovery. Post-hoc CBM~\cite{yuksekgonul2022post} discovers concepts by applying linear SVMs to pre-trained feature spaces, while other methods~\cite{yan2023towards} identify concepts through data-driven analysis like GradCAM visualization and spectral clustering. Foundation models provide another avenue for concept discovery, with large language models generating concept candidates~\cite{yuksekgonul2022post} and CLIP-based models enabling concept-image alignment~\cite{oikarinen2023label, yang2023language, srivastava2024vlg}.

Hybrid approaches combine supervised and unsupervised learning to leverage their complementary strengths. Side-channel CBM~\cite{havasi2022addressing} employs a generative framework that jointly models pre-defined concepts $c$ and latent concepts $z$ through $p_\theta(c, z|x)$, capturing information beyond known concepts. Res-CBM~\cite{shang2024incremental} enhances this approach by developing methods to translate learned residual concepts into human-understandable terms, effectively bridging automated discovery with interpretability. 
Human intervention also plays a crucial role in concept supervision, which enables users to correct model errors and provide additional context.
Beyond basic error correction on concept mapping, more nuanced and efficient interactive frameworks have been developed to optimize human involvement.
For example, memory-based methods~\cite{steinmann2023learning} store users' corrections in memory and reuse these annotations in future predictions, which enhances the model's generalizability.
Strategic intervention policies~\cite{raman2024understanding, xu2024energy, vandenhirtz2024stochastic} propose to selectively request human input based on uncertainty, which allows users to prioritize interventions on concepts with high uncertainty. Causal CGM~\cite{dominici2024causal} allows users to manipulate concepts and observe the model's response, supporting interventional and counterfactual reasoning. Together, these strategies enhance the efficiency and effectiveness of human-in-the-loop systems in concept models.

\subsection{Prototype-based Self-Interpretation}
\label{sec:case-reasoning}
Prototype-based models are inspired by case-based reasoning, a human-like problem-solving paradigm that leverages past experiences to address new challenges. This approach treats previously encountered situations as ``representative examples"~\cite{van2010example}, guiding decision-making for new cases. As end-to-end trainable models, prototype-based neural networks aim to learn representative global examples from data by automatically identifying prototypes within the input feature space and using them to make predictions.
The prediction process involves computing the similarity between the input data and the learned prototypes and aggregating these similarities through a transparent function. This methodology not only facilitates accurate and robust predictions but also clearly indicates the decision-making pathway. Formally, this process can be abstracted as:
\begin{equation} 
    \begin{aligned} 
        z &= f_{enc}(G(x)), \\
        \hat{y} &= \sum_{j=1}^{N_p} W_{j} \cdot \text{sim}(z, p_{j}), 
    \end{aligned} 
\end{equation}
Here, $G(\cdot)$ represents an input processing function (e.g., substructure extraction), and $f_{\text{enc}}(\cdot)$ is an encoder network that maps the processed input to a latent representation $z$. Each prototype embedding $p_j$ acts as an anchor reference point in the latent space shared with $z$. $N_p$ denotes the total number of prototypes, and the similarity function $\text{sim}(z, p_j)$ is commonly defined as a log-activation: $$\text{sim}(z, p_{j}) = \log \left( \frac{(z - p_{j})^2 + 1}{(z - p_{j})^2 + \epsilon} \right),$$ which monotonically decreases with the squared $L_2$ distance between $z$ and $p_j$. A small constant $\epsilon$ is included to avoid numerical instability. It could be replaced by other similarity metrics, such as cosine similarity~\cite{Yang2024TOIS, tucker2022prototype}. The prediction $\hat{y}$ is obtained by aggregating similarities through a linear weight matrix $W$, optionally followed by activation functions or additional operations to adapt to different tasks.
For classification tasks, the output layer often applies a Softmax function to produce a probability distribution over the classes.\footnote{To illustrate the core design principles of prototype models, we use standard regression and classification tasks as examples. Notably, the prototype-based framework is adaptable to more complicated task settings~\cite{rymarczyk2023icicle, jiang2023fedskill}.}

Prototype-based models typically use an iterative optimization method~\cite{chen2019looks}, as outlined in~\Cref{alg:prototype}. This method consists of three stages: (1) the initial training phase, which aims to learn diverse and representative prototype embeddings in the hidden state, (2) the prototype refinement phase, which maps learned embeddings into the nearest training data samples, and then prunes or merges duplicated prototypes for better human interpretability, and (3) the model fine-tuning phase, which fixes all the parameters of the backbone encoder and prototype embeddings, and learns a sparse weight matrix to focus on the most informative prototype-to-output connections.
All these stages aim to enhance the model's interpretability by learning meaningful prototype embeddings or aligning them with human-understandable case examples. Following this fundamental framework, recent research has focused on enhancing prototype representations, organizations, and alignment strategies.

\begin{algorithm}[t]
    \caption{Generalized Prototype Learning Algorithm}
    \label{alg:prototype}
    \begin{algorithmic}[1]
    \Require Training dataset $\mathcal{D}$, number of prototypes $m$
    \Ensure Encoder $f_{enc}$, prototypes $\{p_j\}$, classifier weights $W$
    
    \Procedure{TrainModel}{}
        \For{epoch $= 1$ to $T$}
            \For{each batch $\{x_i, y_i\}$}
                \State $z_i \gets f_{enc}(G(x_i))$ \Comment{Encode input}
                \State $\hat{y}_i \gets \sum_{j} W_j \cdot \text{sim}(z_i, p_j)$ \Comment{Comparison}
                \State $L_{total} \gets L_{ce}(\hat{y}_i, y_i) + L_{interp}(z_i, \{p_j\})$
                \State Update parameters using $\nabla L_{total}$
            \EndFor
    
            \If{epoch $\bmod~\tau = 0$}
                \State $p_j \gets \text{Search\_Examples}(p_j, \mathcal{D})$ for each $p_j$
            \EndIf
        \EndFor
    \EndProcedure
    
    \Procedure{Inference}{$x$}
        \State \Return $\sum_{j} W_j \cdot \text{sim}(f_{enc}(x), p_j)$
    \EndProcedure
    \end{algorithmic}
\end{algorithm}

\subsubsection{Prototype Representations} 
Recent studies have mainly explored two distinct but complementary approaches to enhance prototype representations.
The first direction focuses on optimizing prototype representation by preserving concise yet informative input features. 
This is based on the assumption that prototypes should capture the most important features of the input data. 
One strategy is to decompose inputs into parts, separating complex patterns and capturing fine-grained local features. For example, image-based models~\cite{kim2021xprotonet, keswani2022proto2proto, nauta2023pip, sacha2023protoseg, rymarczyk2023icicle, carmichael2024pixel} partition images into patches to capture local features. Similarly, ProtoryNet~\cite{hong2023protorynet} uses syntactic decomposition for text data, while PGIB~\cite{seo2024interpretable} applies information-theoretic subgraph extraction for motif discovery on graph data. Given the noise and redundancy in real-world data, some works aim to filter out irrelevant information from inputs. For instance, ProtoPFormer~\cite{xue2022protopformer} uses a foreground-preserving mask to reduce background noise, and LucidPPN~\cite{pach2025lucidppn} isolates color information to avoid ambiguity in learned prototypical images.

The second direction focuses on optimizing prototype geometry to improve expressiveness. A notable example is TesNet~\cite{wang2021interpretable}, which constructs a transparent embedding space based on the Grassmann manifold to connect the high-level input space with learned prototypes. Several VAE-based methods~\cite{kjaersgaard2024pantypes, gautam2022protovae, haselhoff2025gaussian} treat prototypes as the center of a Gaussian distribution in the latent space, providing a probabilistic anchor for representations. ProtoConcepts~\cite{ma2024looks} extends this idea by introducing a ball-based prototype geometry, representing each prototype as a set rather than a single point.

\subsubsection{Prototype Organization}
Traditionally, prototype-based models use a fixed number of prototypes in a single layer, which may not be optimal for all tasks. Recent research has explored various strategies for organizing prototypes to better adapt to different data complexities. One approach is hierarchical prototype structures~\cite{nauta2021neural, hase2019interpretable}, which organize prototypes across multiple layers to capture different levels of abstraction. For example, ProtoTree~\cite{hase2019interpretable} uses a tree-like routing structure, enabling more complex decision boundaries and decision paths.
Similarly, ST-ProtoPNet~\cite{wang2023learning} employs a dual prototype structure, learning two distinct sets of prototypes to separate boundary-distant and boundary-proximate representations. 
Another strategy is dynamic prototype assignment, designed to reduce redundancy and enhance model efficiency. For example, ProtoPool~\cite{rymarczyk2022interpretable} maintains a global set of prototypes and uses a soft assignment strategy based on prototype distributions for making predictions. This substantially reduces the number of prototypes required for accurate predictions.

\subsubsection{Prototype Alignment} 
Since most prototype models learn latent embeddings to represent prototypes, prototype alignment with human-interpretable concepts has become a key focus in the development of these models. While traditional approaches rely on iterative search to project prototypes onto the nearest training samples~\cite{chen2019looks, ragodos2022protox}, newer methods have introduced more efficient alternatives, particularly for data with inherent structural complexities, such as graphs and sequences.
ProtGNN~\cite{zhang2022protgnn} and PGIB~\cite{seo2024interpretable} utilize Monte Carlo Tree Search (MCTS) to efficiently navigate complex search spaces. ProSeNet~\cite{ming2019interpretable} applies a greedy beam search algorithm, while SeSRDQN~\cite{Yang2024TOIS} combines neural decoders with progressive MCTS to balance computational efficiency and alignment accuracy. Human integration has also become increasingly important, with models like PIP-Net~\cite{nauta2023pip} and ProtoPDebug~\cite{bontempelli2022concept} incorporating direct human feedback to refine prototypes. PW-Net~\cite{kenny2023towards} takes a different approach by integrating human-designed prototype layers into pre-trained models.
Additionally, optimization strategies have evolved to include cross-class prototype sharing, as demonstrated by ProtoPShare~\cite{rymarczyk2021protopshare}, which employs data-dependent similarity metrics to merge redundant prototypes while maintaining prediction performance.

\subsection{Rule-based Self-Interpretation}
\label{sec:rule}
Rule-based methods incorporate explicit rules into their neural networks. 
These rules can be either pre-defined or learned during training, providing clear, human-understandable inference logics.
Based on design principles, we categorize rule-based self-interpretation neural networks into four groups: (1) \textit{logical operators as neurons}, (2) \textit{logic-inspired constraints}, (3) \textit{rule generation networks}, and (4) \textit{interpretable neural trees}.

\subsubsection{Logical Operators as Neurons}
\label{sec:logic_rule1}
A natural way to enable neural networks to learn rules is by replacing standard neurons with differentiable logical operators. This is similar to architectural-based equation learning (\Cref{sec:equation}) but focuses specifically on logical operations.
However, the difficulty lies in enabling gradient-based optimization, which requires the logical activation function $f_{\text{logic}}(\cdot)$ to be differentiable. 
Traditional Boolean logic operators ($\text{AND}$, $\text{OR}$, $\text{NOT}$) produce discrete outputs \{0, 1\}, making direct optimization infeasible.
To address this, logical neurons use continuous fuzzy approximations like t-norms or t-conorms~\cite{van2022analyzing, petersen2022deep}. \Cref{tab:tnorms} summarizes the operations of three fundamental t-norms: Product, Gödel, and Łukasiewicz~\cite{marra2020relational, van2022analyzing}. These functions could smoothly approximate Boolean logic while remaining differentiable for inputs in [0, 1]. 

A key design challenge lies in specifying or discovering optimal network architectures that integrate logical operators, given the vast space of possible configurations~\cite{donadello2017logic}.
Inspired by Disjunctive Normal Form (DNF)'s ability to represent any Boolean formula, MLLP~\cite{wang2020transparent} and DR-Net~\cite{qiao2021learning} build a hierarchical structure through alternating 
conjunction and disjunction layers, as illustrated in~\Cref{fig:rule-1}.
In conjunction layers, neurons compute $r_i^{(l)} = \bigwedge_{W_{i,j}^{(l)}=1} s_j^{(l-1)}$, while in disjunction layers, they compute $s_i^{(l+1)} = \bigvee_{W_{i,j}^{(l+1)}=1} r_j^{(l)}$, where $W_{i,j} \in \{0,1\}$ are binary weight matrices indicating connections. To enable differentiable optimization, MLLP applies Random Binarization, temporarily binarizing weights during training, while DR-Net uses straight-through estimators with gradient clipping.
However, this framework faces vanishing gradients in high dimensions since its logical activation functions multiply many terms between 0 and 1, causing gradients to approach zero exponentially. To address this, RRL~\cite{wang2021scalable, wangrule2023learning} proposes improved activation functions that convert multiplications into additions via logarithm, with a projection mechanism $\text{Proj}(v) = -1/(-1+\log(v))$ to maintain the properties of logical operations while preventing gradient vanishing. 
HyperLogic~\cite{yanghyperlogic} further improves its scalability and flexibility on large-scale datasets by using an extra hyper-network to generate weights for the base rule network instead of directly optimizing them.
Beyond fixed logical operators, several studies~\cite{petersen2022deep, petersenconvolutional, zhang2023learning_neuralrule} also propose to learn a probability distribution over logic gates using softmax during training: $f_{\text{logic}}({x}) = \sum_{g \in \mathcal{G}} p(g) g(\bm{x})$, where $\mathcal{G}$ represents the 16 possible logic gates, and $p(g)$ denotes the learned probability for each gate. This enables more flexible, efficient, and adaptive rule learning.

\begin{table}[t]
    \centering
    \caption{The algebraic operations for t-norm logics~\cite{marra2020relational}.}
    \begin{tabular}{@{}lcccc@{}}
    \toprule
    \multirow{2}{*}{\textbf{Operators}} & \multicolumn{3}{c}{\textbf{t-norm}} \\ \cmidrule(lr){2-4}
                                 & \textbf{Product}       & \textbf{Gödel}         & \textbf{Łukasiewicz}        \\ \midrule
    $x \wedge y$        & $x \cdot y$            & $\min(x, y)$           & $\max(0, x + y - 1)$        \\
    $x \vee y$          & $x + y - x \cdot y$    & $\max(x, y)$           & $\min(1, x + y)$            \\
    $\neg x$            & $1 - x$                & $1 - x$                & $1 - x$                    \\
    $x \Rightarrow y$   & $x \leq y ? 1 : \frac{y}{x}$ & $x \leq y ? 1 : y$ & $\min(1, 1 - x + y)$       \\ \bottomrule
    \end{tabular}
    \vspace{-3mm}
    \label{tab:tnorms}
\end{table}

\subsubsection{Logic-Inspired Constraints}
\label{sec:logic_rule2}
Rather than directly using logical operators as neurons, several studies~\cite{ciravegna2020human, ciravegna2023logic, jain2022extending} focus on imposing logic-inspired constraints to guide neural network training, enabling logic extraction. The core idea involves interpreting the neural network's behavior in boolean terms, where each neuron's input-output is approximated to its closest integer value (0 or 1). The learned boolean function of each neuron can be represented as an empirical truth table, which can then be converted into a logical formula in DNF. As illustrated in~\Cref{fig:rule-2}, hidden and output neurons are paired with truth tables and their corresponding logical formulas. These truth tables include real-valued neuron outputs (third column) and their boolean approximations (last column). The derivation of logical formulas from truth tables is a well-studied problem in logic synthesis, where the goal is to find the simplest logical expression that represents the truth table. This typically involves constructing a disjunction (OR) of minterms, where each minterm is a conjunction (AND) of literals (a variable or its negation). The inference path from input features to the targeted activation neuron (e.g., class label) establishes logical relationships between input features and the final prediction. In the example of~\Cref{fig:rule-2}, the neuron $h_1^{(2)}$ is represented by the logical formula $\neg h_1^{(1)} \land \neg h_2^{(1)}$, which explicitly captures the logical dependencies among the input features $x_1, x_2, x_3$.

To promote sparsity and enhance logic extraction, these networks are typically regularized using sparsity or activation constraints. For example, some studies~\cite{ciravegna2020constraint, barbiero2022entropy} introduce an entropy layer that identifies sparse features through weight-based relevance scores and competition between input features. Compared to relying solely on concept truth degrees, DCR~\cite{barbiero2023interpretable} incorporates concept embeddings as inputs to neural modules, determining each concept's role (positive or negative) and relevance within the rule. These concept embeddings, alongside truth degrees, provide a richer representation, capturing more complex concept relationships.

\subsubsection{Rule Generation Networks}
\label{sec:logic_rule3}
\begin{figure}[t]
    \centering
    \subfloat[Logical Operators as Neurons~(\Cref{sec:logic_rule1})]{
        \includegraphics[width=0.95\linewidth]{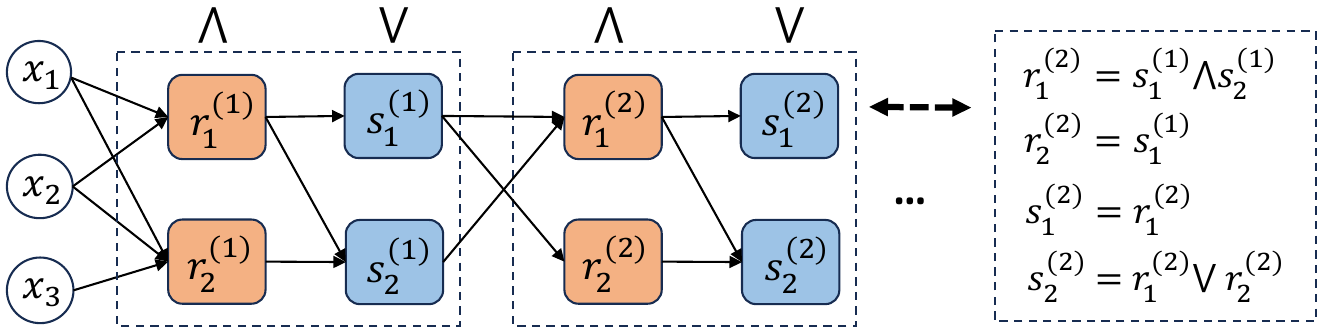}
        \label{fig:rule-1}
    }

    \subfloat[Logic-Inspired Constraints~(\Cref{sec:logic_rule2})]{
        \includegraphics[width=0.95\linewidth]{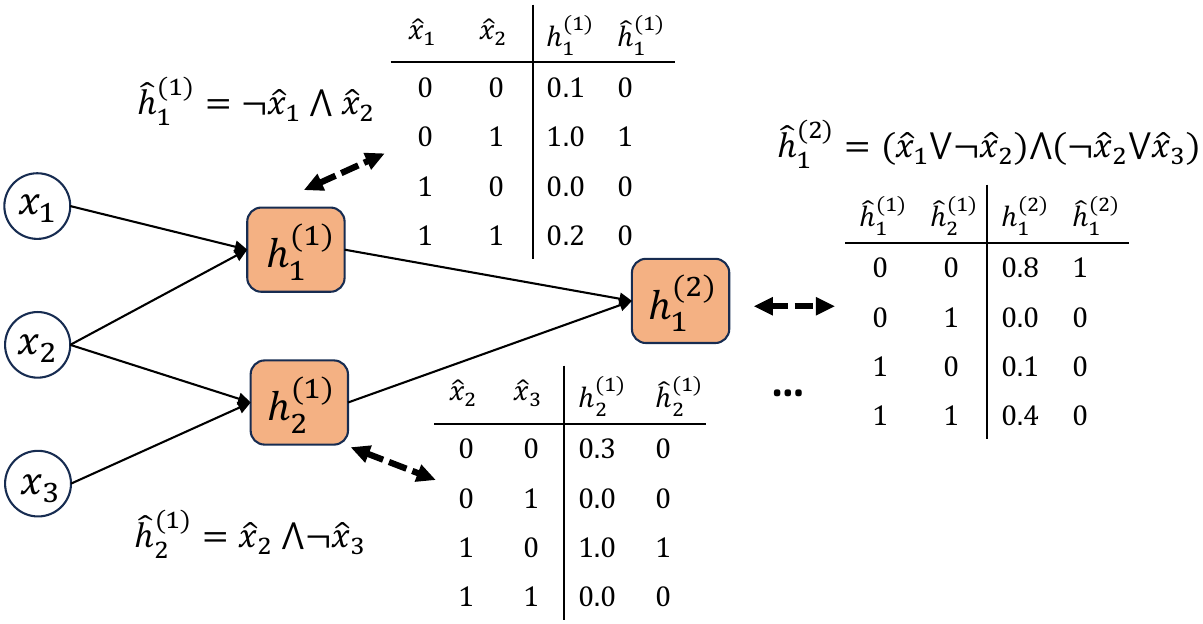}
        \label{fig:rule-2}
    }
    
    \subfloat[Rule Generation Network~(\Cref{sec:logic_rule3})]{
        \includegraphics[width=0.95\linewidth]{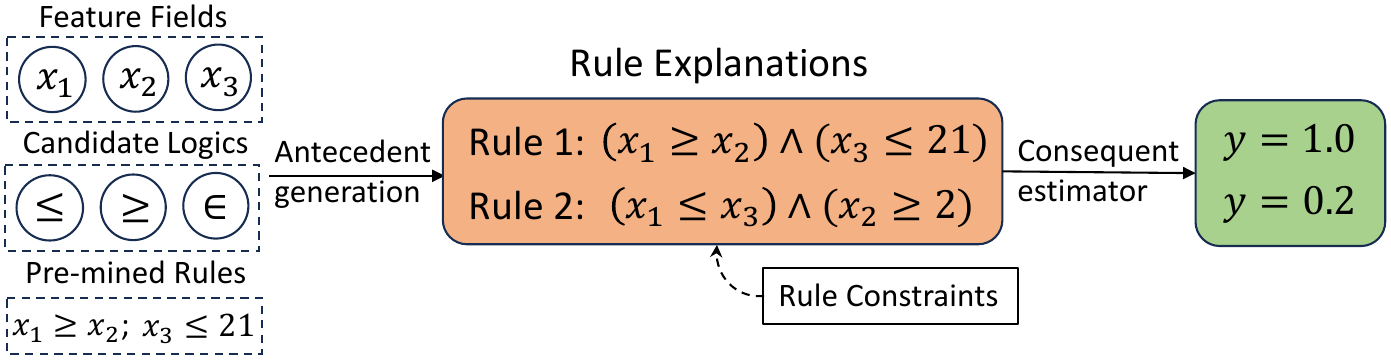}
        \label{fig:rule-3}
    }
    
    \subfloat[Interpretable Neural Tree~(\Cref{sec:logic_rule4})]{
        \includegraphics[width=0.95\linewidth]{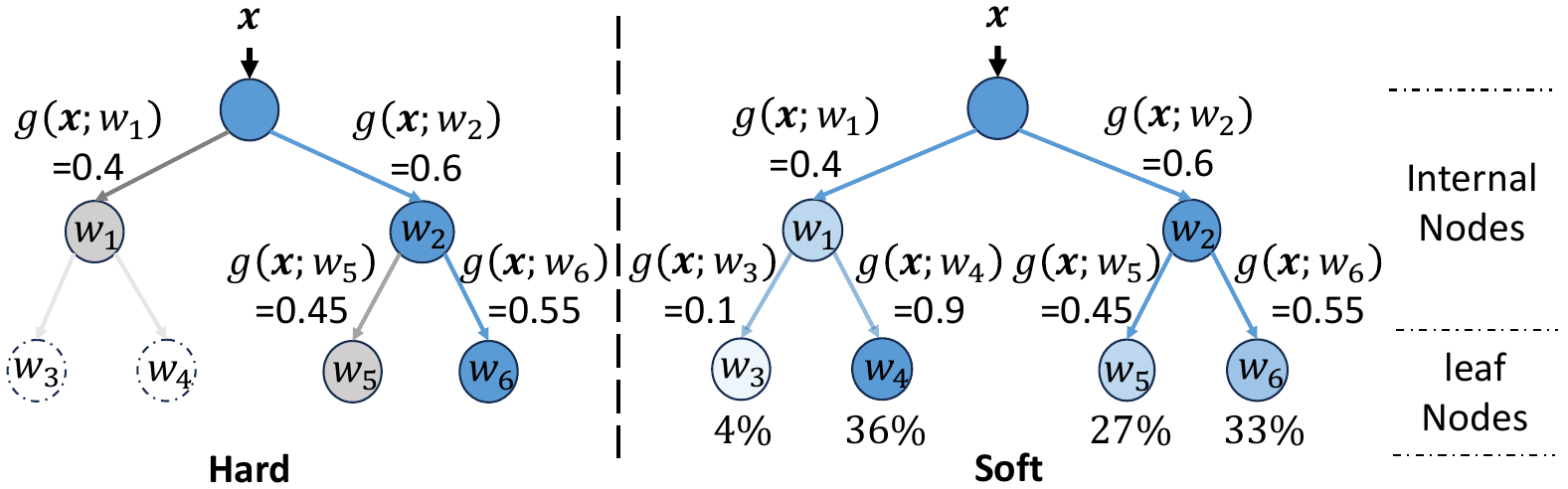}
        \label{fig:rule-4}
    }
    
    \caption{Graphical illustrations of rule-based self-interpretation.}
    \label{fig:rule}
    \vspace{-3mm}
\end{figure}

Rule generation networks encode logical rules as hidden representations within neural architectures, capturing dependencies among input features without explicit logical neurons or binary constraints. 
Similar in spirit to generative equation learning approaches (\Cref{sec:equation}), these methods are often guided by inductive biases or symbolic constraints that encourage the model to learn rules that are both accurate and interpretable.
As illustrated in~\Cref{fig:rule-3}, rule generation typically follows a two-step process: (1) \textit{Antecedent Generation}: constructing a structured rule space either by selecting pre-defined rules or dynamically composing rules from basic atoms, and (2) \textit{Consequent Evaluation}: assessing rules to derive predictions.

Early works like Rule-Constrained Network (RCN)~\cite{okajima2019deep} construct rule spaces using conventional data mining techniques, including frequent itemset mining (e.g., Eclat, FP-Growth) or decision trees. RCN embeds these pre-mined rules into latent neural representations, then uses attention mechanisms to measure similarity between rule embeddings and input features. This encourages the model to select rules whose antecedents are satisfied by input features and whose consequents align with target classes. However, pre-mined rule spaces may lack sufficient coverage, potentially overlooking novel or significant logical patterns.

Dynamic rule composition methods address this limitation by adaptively learning expressive and semantically meaningful rules during training. 
For example, SELOR~\cite{lee2022self} incrementally generates logical antecedents by recursively sampling atomic conditions in a differentiable manner, and systematically evaluates their global consistency and confidence using a neural consequent estimator. 
ENRL~\cite{shi2022explainable} and FINRule~\cite{yu2023finrule} frame rule learning as Neural Architecture Search (NAS), and propose Explainable Condition Modules (ECMs) to assess the satisfaction of basic rule propositions via learnable feature fields, operator modules, and context embeddings. ENRL organizes ECMs into multi-tree structures and aggregates activated rules through voting mechanisms for prediction, whereas FINRule uses a combination of context embeddings of ECMs and GNN-based attentive methods to capture higher-order logical interactions. Additionally, EPR~\cite{wuweakly}, specialized for natural language inference, utilizes transformer encoders to align semantically similar phrases between sentences and predicts phrase-level logical relationships (Entailment, Contradiction, Neutral) using a neural classifier. It subsequently aggregates these phrase-level predictions into sentence-level outcomes via geometric means.

\subsubsection{Interpretable Neural Tree}
\label{sec:logic_rule4}
The neural tree is a differentiable neural architecture that mimics the hierarchical structure of traditional decision trees. 
As illustrated in~\Cref{fig:rule-4}, the neural tree simulates a decision tree by representing each internal node with a routing function and each leaf node with a decision function. Unlike traditional trees that make discrete decisions, the neural tree generally employs soft decisions, allowing inputs to be routed through multiple paths in a differentiable manner. This enables end-to-end training using back-propagation, integrating tree-like interpretability with neural network learning capabilities.

Formally, the neural tree is defined as:
\[
T = (\mathcal{N}, E),
\]
where \(\mathcal{N} = \mathcal{U} \cup \mathcal{W}\)  is the set of nodes and \(E\) is the set of directed edges. Nodes are categorized as:
\begin{itemize}[noitemsep, left=0pt]
    \item \textit{Internal node} \(u \in \mathcal{U}\) with a soft routing function:
    \[
    g_u(\bm{x}; \theta_u) : \mathbb{R}^d \to [0,1],
    \]
    determining the probability of routing input \(\bm{x}\) to each child.
    
    \item \textit{Leaf node} \(w \in \mathcal{W}\) with a decision function:
    \[
    f_w(\bm{x}; \theta_w) : \mathbb{R}^d \to \Delta^{|Y|-1},
    \]
    providing the final prediction results as probabilistic class distribution or single class label.
\end{itemize}

To unify these nodes, let \(h_n(\bm{x})\) denote the output of any node \(n \in \mathcal{N}\). The neural tree output is recursively defined as:
\begin{equation}
h_n(\bm{x})
=
\begin{cases}
\displaystyle
\sum_{c \in \text{child}(n)}
g_c(\bm{x}; \theta_c)\,h_c(\bm{x}), 
& \text{if } n \in \mathcal{U}, \\[6pt]
f_n(\bm{x}; \theta_n),
& \text{if } n \in \mathcal{W}.
\end{cases}
\label{eq:neural_tree_output}
\end{equation}
Finally, the overall model output is \(h_\text{root}(\bm{x})\), where \(\text{root}\) is the root node that aggregates all child nodes' outputs in the tree.

The core challenge in designing interpretable neural trees lies in defining the routing functions. Various studies have proposed different routing mechanisms to enhance expressiveness while maintaining interpretability.  
Early works, such as DNDT~\cite{yang2018deep} and DNDF~\cite{kontschieder2015deep} introduce a soft-binning function for path routing:  
$
g_u(\bm{x}; \theta_u) = \mathrm{softmax} \Bigl(\left(\bm{w}_u^\top \bm{x} + b_u\right)/{\tau}\Bigr),
$
where $\theta_u = \{\bm{w}_u, b_u\}$ defines learned cut points, enabling smooth transitions between paths while preserving explicit decision boundaries. During inference, DNDT applies a hard decision strategy at each internal node by selecting the child with the highest probability, proceeding step-by-step to a leaf node for final prediction, as illustrated in~\Cref{fig:rule-4} (Left).
In contrast, NBDT~\cite{wan2020nbdt} proposes similarity-based routing, where decisions are guided by learned node embeddings:
$
g_u(\bm{x}; \theta_u) = \mathrm{softmax} \left( \langle \bm{n}_u, \bm{x} \rangle \right),
$
Unlike DNDT's hard binary decisions, NBDT performs soft inference by aggregating predictions across all paths weighted by their probabilities, which allows it to better handle uncertainty at intermediate nodes, as shown in~\Cref{fig:rule-4} (Right).
Further advancements~\cite{nauta2021neural, kim2022vit, lao2024vitree} explore prototype-based routing, where decisions rely on the similarity between the input embeddings and learned prototypical patterns:  
$
g_u(\bm{x}; \theta_u) = \min \|\bm{x} - \bm{p}_u\|^2.
$
Here, $\theta_u = \{\bm{p}_u\}$ represents learned prototypes, making reasoning inherently interpretable by aligning the input with stored data patterns.  
Several works~\cite{tanno2019adaptive, ji2020attention} further enhance learned representations of the input as the model deepens. Rather than restricting routing to partitioning the shared data space, these methods incorporate transformer or attention mechanisms at decision edges, applying additional nonlinear transformations to the input for better expressiveness.

\begin{figure}[t]
    \centering
    \includegraphics[width=0.9\linewidth]{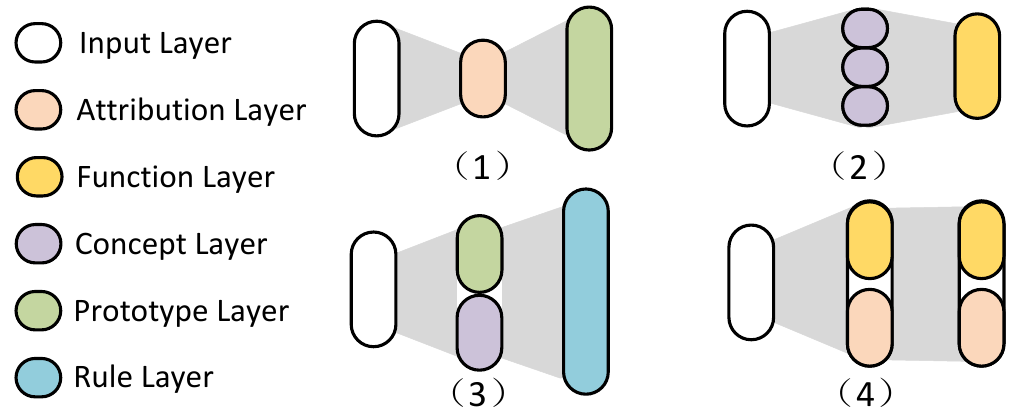}
    \caption{A graphical illustration of hybrid self-interpretation methods that combine different self-interpretation components.}
    \label{fig:design}
\end{figure}
  
\subsection{Hybrid Self-Interpretation with Combinatorial Methods}
\label{sec:hybrid}
While previous sections focused on the design of individual self-interpretation methods, these modules can be integrated together instead of existing in isolation. 
Many works discussed earlier exhibit this characteristic. For example, ProtoTree~\cite{nauta2021neural} combines prototype-based routing with rule-based decision trees, and DCR~\cite{barbiero2023interpretable} incorporates concept layers with logical inference rules. These advanced approaches highlight the feasibility and advantages of hybrid self-interpretation, providing richer and more comprehensive explanations. Following the order of sub-figures in \Cref{fig:design}, we sequentially illustrate representative hybrid methods with concrete exemplar studies, highlighting how different self-interpretation components are effectively integrated into a cohesive design.

\begin{enumerate}[noitemsep, left=0pt]
    \item \textit{Attribution-guided prototype learning.} Extracting substructures from inputs before prototype learning has dual benefits. It not only distills local discriminative features from the input but also guides prototypes toward global informative features. For example, PGIB~\cite{seo2024interpretable} applies an information bottleneck to compress graphs, subsequently feeding them into a prototype-based module to identify core subgraphs.
    \item \textit{Concept-based transparent functions.} For high-dimensional data, directly constructing transparent functions on raw inputs is challenging. Introducing concept learning as an intermediate step can simplify the input space and improves interpretability. For example, CAT~\cite{duong2024cat} maps inputs into concepts and designs a white-box Taylor Neural Network that directly learns the relationships between concepts and output with polynomials.
    \item \textit{Concept/Prototype-based rule construction.} Effective rule learning requires interpretable data representations. Concepts or prototypes can serve as building blocks for conditional decisions, particularly with high-dimensional data. For example, LEN and variants~\cite{ciravegna2020constraint, barbiero2022entropy, barbiero2023interpretable} first project inputs into concept spaces, then apply logical constraints to extract logical rules. ProtoTree~\cite{nauta2021neural} and ViT-Tree~\cite{kim2022vit} utilize interpretable prototype comparisons as routing functions within neural trees.
    \item \textit{Stacked function-based attribution.} A single interpretable network module can be repeatedly stacked to provide multiple levels of self-interpretation. Some studies~\cite{bohle2021convolutional, bohle2024b, bohle2022b} explicitly design transparent functions, such as $l(x) = \bm{\alpha}(x)x$, between layers, clearly attributing feature contributions and modeling variable relationships across network layers.
\end{enumerate}

These examples illustrate the potential of hybrid self-interpretation methods to provide richer and more comprehensive explanations. By combining different self-interpretation components, hybrid SINNs can leverage the strengths of individual techniques while mitigating their limitations, resulting in more robust and multifaceted interpretability.

\section{Domain-Specific Self-Interpretation}
\label{sec:applications}
Due to the heterogeneous nature of data types and tasks, SINNs have been tailored to meet specific requirements across different fields. In this section, we review their applications for image, text, graph, and DRL. For each domain, we briefly summarize their popular explanation forms, customized methodologies, and discuss emerging domain-specific topics.

\begin{table}[t]
  \centering
  \caption{Explanation Visualizations of Images}
  \label{tab:cv}
  \renewcommand{\arraystretch}{0.85} %
  \setlength{\tabcolsep}{2pt} %
  \small %
  \resizebox{0.9\columnwidth}{!}{ %
  \begin{tabular}{>{\centering\arraybackslash}m{2.5cm} >{\centering\arraybackslash}m{3.0cm} >{\centering\arraybackslash}m{2.5cm}}
    \toprule
    \textbf{Understandable Terms} & 
    \textbf{Visualizations} & 
    \textbf{Explanation Types} \\
    \midrule
    \multirow{4}{*}{Pixels} & \raisebox{-0.5\height}{\includegraphics[width=2.5cm,height=2.0cm,keepaspectratio]{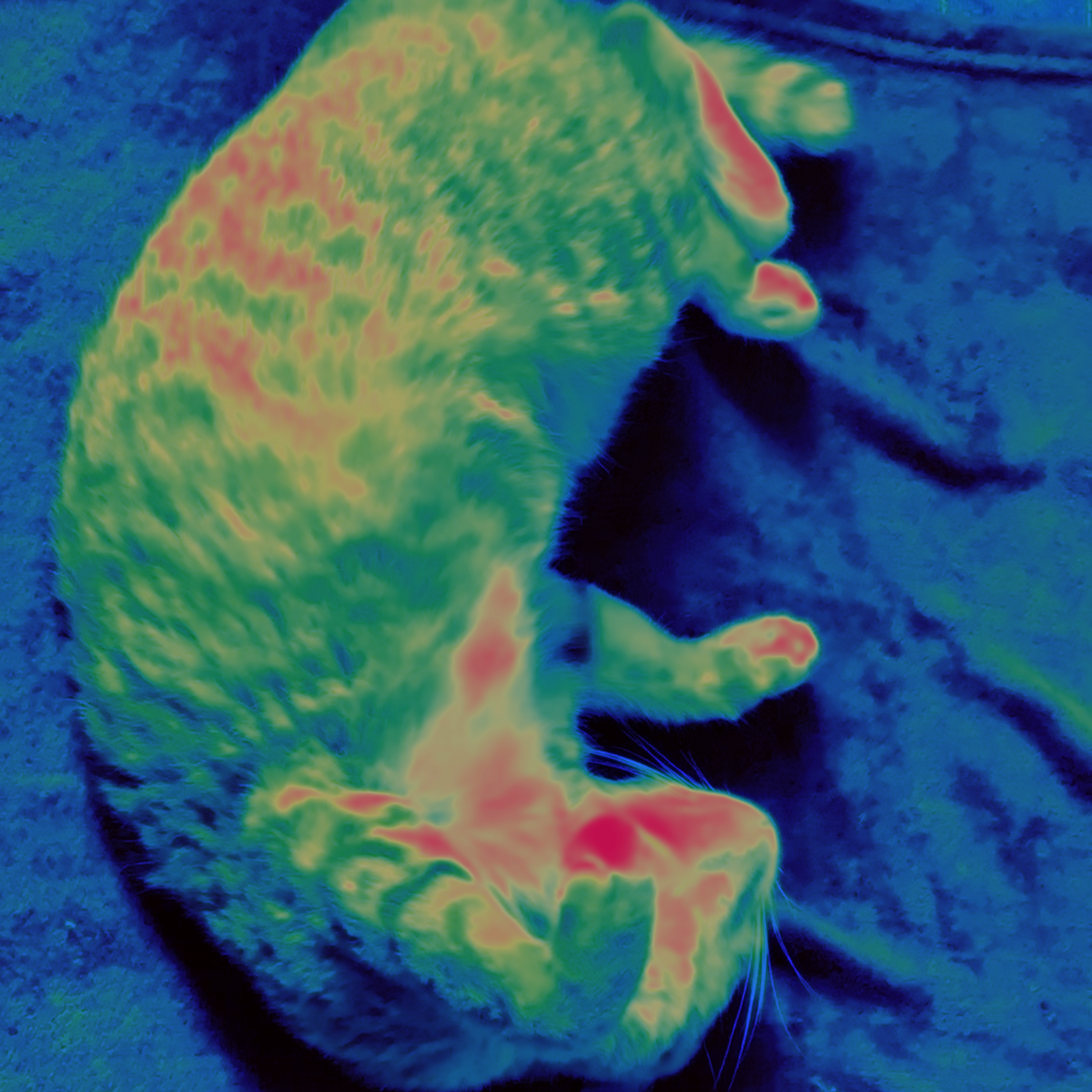}} & \multirow{4}{2.5cm}{\centering Spatial Regions, \\ Patches}  \\
    \midrule
    \multirow{4}{*}{Patterns} & \raisebox{-0.5\height}{\includegraphics[width=2.5cm,height=2.0cm,keepaspectratio]{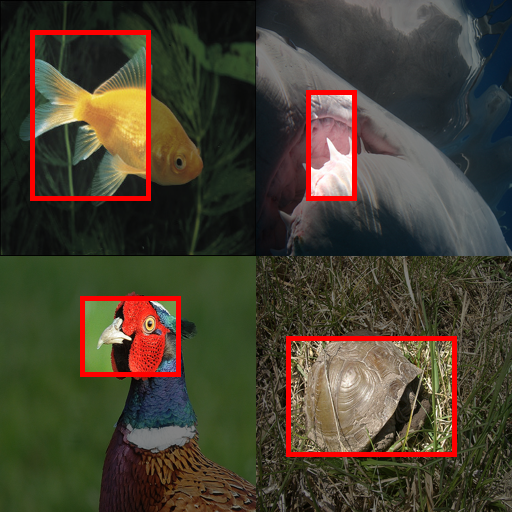}} & \multirow{4}{2.5cm}{\centering Colors, Materials, \\ Textures, Scenes} \\
    \midrule
    \multirow{4}{*}{Objects} & \raisebox{-0.5\height}{\includegraphics[width=2.5cm,height=2.0cm,keepaspectratio]{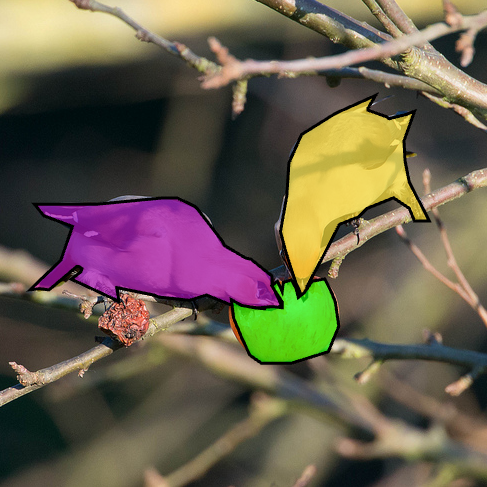}} & \multirow{4}{2.5cm}{\centering High-Level\\Object Concepts}
    \\
    \bottomrule
  \end{tabular}}
\end{table}

\subsection{Image Data}
\label{sec:cv}
SINNs for image data often follow general paradigms such as prototype learning~\cite{chen2019looks}, concept modeling~\cite{koh2020concept}, and attribution approaches~\cite{chen2018learning}. Many of these techniques originated in image-related applications like image classification~\cite{rymarczyk2022interpretable, wang2023learning}, and segmentation~\cite{sacha2023protoseg}. A distinct characteristic of image data is that individual pixels typically lack standalone semantics. Thus, effective explanations must aggregate pixels into meaningful visual units, resulting in multiple granularities of explanation. We focus on two complementary perspectives uniquely suited for image data: \textit{visual explanation granularity} and \textit{interpretable CNN filters}. The former categorizes the abstraction level of visual explanations, while the latter leverages CNN structures to capture interpretable features.

\textbf{Visual explanation granularity} varies in abstraction, including pixel, pattern, and object-level explanations.
As shown in~\Cref{tab:cv}, each offers unique visual insights and serves distinct applications. 
\textit{Pixel explanations} group pixels into cohesive regions that form informative visual units while maintaining spatial coherence. These regions can be pre-defined using fixed grids~\cite{nauta2023pip, wangvisual, chen2024neural, wang2023learning} or learned dynamically through masking techniques like attention~\cite{zhang2019pathologist} and information bottleneck methods~\cite{schulzrestricting}. Dynamic regions offer greater adaptability to complex visual patterns and often incorporate sparsity and semantic regularization for improved interpretability. \textit{Pattern explanations} identify recurring visual motifs such as textures or colors rather than focusing on individual pixels. They are often associated with higher-level concept modeling~\cite{koh2020concept, espinosa2022concept, kim2023probabilistic}, where images are represented using pattern-level attributes derived from human annotations or learned through vision-language models (VLMs)\cite{oikarinen2023label, yang2023language}. \textit{Object-level explanations} emphasize relationships between distinct objects, making them particularly useful for object detection and visual question answering. Methods\cite{xu2020explainable, yi2018neural} leverage pre-trained models to extract object boundaries and learn symbolic relationships, offering insights into structural composition and visual interactions.

\textbf{Interpretable CNN filters} enhance CNN transparency by structuring feature representations in convolutional layers. Traditional CNN filters extract hierarchical features, with lower layers capturing edges and textures and deeper layers detecting object parts and high-level semantics. However, these filters often learn entangled patterns, making it unclear how they contribute to specific predictions. Interpretable filters address this by enforcing constraints that disentangle feature maps and align them with human-understandable concepts.
For example, ICNN~\cite{zhang2020interpretable} enforces spatially localized filters through information-theoretic constraints, while ICCNN~\cite{shen2021interpretable} extends this to both structured object parts and unstructured image regions. Recognizing that filters often respond to multiple classes, complicating interpretability, Liang et al.~\cite{liang2020training} propose assigning each filter to one or a few classes, ensuring activations occur only for relevant samples. PICNN~\cite{guo2024picnn} further refines this approach by modeling probabilistic filter-class relationships for greater flexibility.

\subsection{Text Data}
\label{sec:nlp}
\begin{figure}[t]
  \centering
  \includegraphics[width=0.9\linewidth]{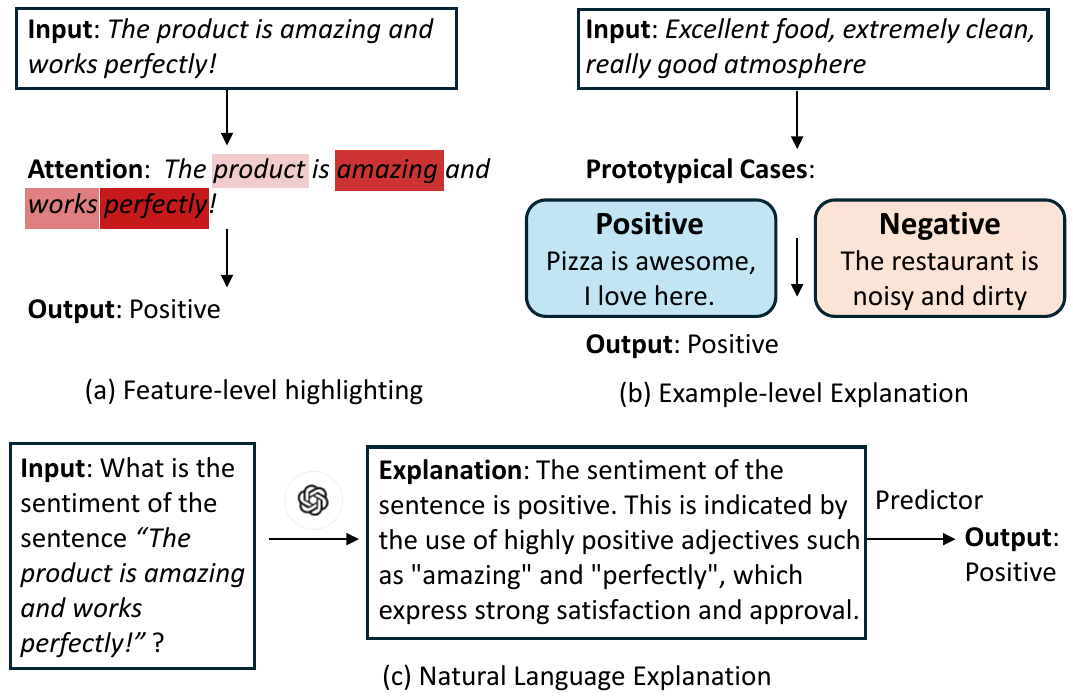}
  \caption{Examples for text explanations.}
  \label{fig:nlp}
\end{figure}

SINNs for text data often provide justifications for natural language processing (NLP) tasks such as sentiment analysis~\cite{luo2018beyond}, text classification~\cite{hong2023protorynet, rajagopal2021selfexplain}, and machine translation~\cite{BahdanauCB14}. Unlike images, which have a spatial structure, text is sequential and symbolic, making interpretation more challenging. In this subsection, we review SINNs for text data based on their explanation forms. \Cref{fig:nlp} and \Cref{tab:nlp} illustrate examples and types of text explanations, including feature, example, and natural language explanations, each operating at different abstraction levels.

\begin{table}[t]
\centering
\caption{Explanation forms of text data}
\label{tab:nlp}
\renewcommand{\arraystretch}{0.85} %
\setlength{\tabcolsep}{2pt} %
\small %
\resizebox{0.98\columnwidth}{!}{ %
\begin{tabular}{>{\centering\arraybackslash}m{2.5cm} >{\centering\arraybackslash}m{3.0cm} >{\centering\arraybackslash}m{2.5cm}}
  \toprule
  \textbf{Understandable Terms} & 
  \textbf{Explanation Type} & 
  \textbf{Representative Works} \\
  \midrule
  \textbf{Feature} &  words, n-grams, phrases, sub-sentence,   &  \cite{lei2016rationalizing, greff2019interpretable,luo2018beyond, yang2016hierarchical, ghaeini2018interpreting, subramanian2018spine, chen2018learning, du2019learning}   \\
  \midrule
  \textbf{Example}  &  Prototypical cases  & \cite{ming2019interpretable, arik2020protoattend, hong2023protorynet}  \\
  \midrule
  \textbf{Natural Language} & explain-then-predict, predict-and-explain  & \cite{chan2022models, zhang2023summarize, jacovi2021aligning, kumar2022nile, rajani2019explain, sun2023lirex, camburu2018towards, wiegreffe2021measuring, marasovic2022few, narang2020wt5, liu2019towards} \\
  \bottomrule
\end{tabular}}
\end{table}

\textbf{Feature-level explanations} identify key textual components (e.g., words, phrases, or subsequences) that influence model decisions. Techniques such as sampling-based approaches~\cite{lei2016rationalizing, greff2019interpretable}, attention mechanisms~\cite{luo2018beyond, yang2016hierarchical, ghaeini2018interpreting}, and sparsity-inducing constraints~\cite{subramanian2018spine, chen2018learning} highlight important input elements. While effective for local interpretability, these methods can be further enhanced by integrating external knowledge bases~\cite{du2019learning} to improve semantic coherence.

\textbf{Example-level explanations} provide a more abstract perspective by grounding model decisions in representative examples. ProSeNet~\cite{ming2019interpretable} determines sentiment by comparing a given review with learned prototypes of positive and negative sentiments, aggregating their influence into the final prediction. Beyond document-level prototypes~\cite{ming2019interpretable, arik2020protoattend}, ProtoryNet~\cite{hong2023protorynet} refines predictions by incorporating sentence-level prototypes that capture finer-grained sentiment trajectories.

\textbf{Natural language explanations} generate textual justifications to clarify model predictions, making interpretability more accessible to end users~\cite{calderon2024behalf, zhang2023triple}. These methods typically follow one of two strategies: \textit{explain-then-predict}\cite{camburu2018snli, kumar2022nile}, where an intermediate explanation is first generated to guide predictions\cite{chan2022models, zhang2023summarize, jacovi2021aligning, rajani2019explain, sun2023lirex, camburu2018towards}, or \textit{predict-and-explain}\cite{wiegreffe2021measuring, marasovic2022few, narang2020wt5, liu2019towards}, where explanations accompany predictions in an integrated manner. With the rise of large language models (LLMs), approaches like Chain-of-Thought\cite{wei2022chain} and Tree-of-Thought~\cite{yao2024tree} have further popularized natural language explanations. However, challenges remain in ensuring their reliability and faithfulness~\cite{madsen2024self, calderon2024behalf, yeo2024interpretable}, particularly in aligning generated explanations with human reasoning and leveraging them transparently for inference.

\begin{figure}[t]
    \centering
    \includegraphics[width=0.95\columnwidth]{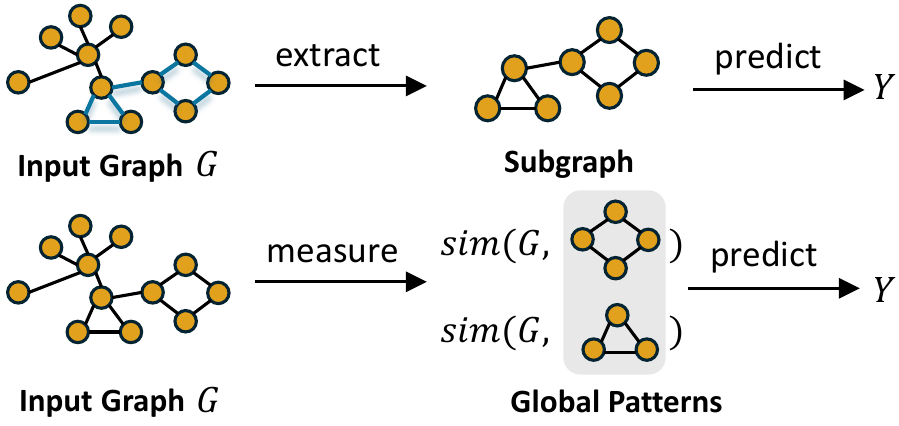}
    \caption{Framework of self-interpretable GNNs.}
    \label{fig:graph}
\end{figure}

\subsection{Graph Data}
\label{sec:graph}
Graphs are widely used to model complex systems, such as molecular structures~\cite{jin2020multi, MMGNN, liu2024towards}, traffic flow~\cite{chen2024tempme}, and knowledge networks~\cite{lanciano2020explainable, huang2025sehg}.
Graphs have irregular structures and relational dependencies, making its interpretation designs distinct from images and texts. 
Unlike function- or concept-based approaches that rely on pre-defined semantic categories, SINNs for graphs focus on structural explanations, which directly relate to the relational nature of graph data. 
As illustrated in~\Cref{fig:graph}, we review the existing self-interpretable GNNs from two perspectives: (1) \textit{local subgraph extraction}, which attributes predictions to influential instance-specific substructures, and (2) \textit{global graph patterns}, which identify recurring structures across instances.

\textbf{Local subgraph extraction} aims to identify the most critical substructure within an input graph and use it for prediction. Since graphs lack a fixed spatial structure, extracting meaningful subgraphs requires selecting key nodes and edges while maintaining structural coherence. A common approach is the Graph Information Bottleneck (GIB), which optimizes a trade-off between informativeness and compression by extracting subgraphs that retain the most relevant predictive information while filtering out unnecessary details~\cite{yu2022improving, yugraph2021}.
This typically involves estimating mutual information between the extracted subgraph, the original graph, and the target, often achieved by sampling nodes or edges based on learned importance scores~\cite{chen2024interpretable, miaointerpretable2023, miao2022interpretable}. Another approach imposes structural constraints to ensure the extracted subgraphs align with interpretable patterns. These include selecting subgraphs that resemble labeled nodes~\cite{dai2021towards}, match pre-defined motifs~\cite{feng2022kergnns, giannini2024interpretable}, or exhibit stable causal relationships with the target~\cite{wudiscovering}.

\textbf{Global graph patterns} provide a holistic view of graph modeling by identifying key patterns across instances. 
Prototype learning is a prevalent approach, where models learn a set of prototypical graphs or subgraphs for each class to guide predictions~\cite{zhang2022protgnn, wang2024unveiling, seo2024interpretable}.
To capture fine-grained patterns, many methods incorporate multi-level structures, such as subgraph-level prototypes\cite{seo2024interpretable} or multi-granularity representations\cite{wang2024unveiling, yang2025from}. For instance, GIP~\cite{wang2024unveiling} clusters nodes to create a coarsened graph and then matches it with learnable global patterns, moving beyond purely local information to capture more abstract structural insights.

\subsection{Deep Reinforcement Learning}
\label{sec:rl}
Self-interpretation is crucial in DRL to ensure agents' transparent and accountable decisions. Unlike supervised learning, DRL involves sequential decision-making under uncertainty, where interpretability can be approached from different perspectives depending on the component being explained. 
We categorize SINNs for DRL into three key areas: (1) \textit{influencing state perception}, which attributes importance to input state features; (2) \textit{value function decomposition}, which breaks down learned value functions into interpretable components; (3) \textit{interpretable policy representation}, which learns structured, human-readable policies.

\textbf{Influencing state perception} explains an agent's decisions by attributing importance to different aspects of the input state. 
It answers the question, ``What part of the state influenced the agent's decision?''. 
These methods typically adapt feature attribution techniques discussed in~\Cref{sec:attribution} to DRL settings. For instance, in image-based tasks (e.g., Atari games), attention mechanisms highlight relevant regions or objects on the screen~\cite{annasamy2019towards, mott2019towards, shi2020self, liu2021learning}. In financial applications, models such as AlphaStock~\cite{wang2019alphastock} emphasize key financial indicators that drive buy-sell decisions. Conceptually, these methods align with SINNs tailored for different input modalities, using structural insights discussed earlier to enhance interpretability.

\textbf{Value function decomposition} clarifies how an agent evaluates its decisions over time by breaking down value functions into interpretable components. Since DRL relies on long-term reward optimization, understanding how rewards propagate through actions helps explain decision-making. Some methods redistribute rewards across trajectories to reveal causal links between actions and their consequences~\cite{liu2023n, yu2023explainable, zhang2024interpretable}. Scenario-based approaches use prototypes or memory mechanisms to recognize examples with high impact on value functions, providing intuitive insights into agent behavior~\cite{annasamy2019towards, ragodos2022protox, kenny2023towards}.
Other techniques focus on identifying critical decision points or generating contrastive explanations through belief maps and time-step analysis, often by decomposing value estimates across trajectories or states, thereby shedding light on how long-term strategies emerge from cumulative value assessments~\cite{guo2021edge, yau2020did}.

\textbf{Interpretable policy representation} aims to structure policies in a human-readable form, enabling users to understand decision-making through explicit rules.
This aligns with broader SINN approaches that enforce structured representations, such as symbolic rules.
One strategy leverages trained neural agents to guide the search for candidate-weighted logic rules, refining them through differentiable logic training~\cite{verma2023interpretable}. Given an input state, this approach extracts relevant entities and relations, converting raw observations into a structured logical representation for human understanding. Other methods, such as ESPL~\cite{guo2023efficient} and DILP~\cite{li2023differentiable}, represent policies as differentiable symbolic expressions. These approaches learn policies from scratch in an end-to-end fashion, designing differentiable symbolic operators and selectors to directly model decision rules without relying on pre-trained agents.

\section{Quantitative Evaluation Metrics}
\label{sec:evaluation}
While evaluation metrics for SINNs are diverse and often tailored to specific applications, they generally reflect shared principles of interpretability. In this section, we organize these metrics into three key aspects: \textit{model performance}, \textit{explanation evaluation}, and \textit{human-centric feedback}.

\textbf{Model Performance.} 
SINNs aim to maintain high predictive performance while offering inherent explanations. In contrast to post-hoc methods, where evaluation often focuses solely on explanation quality, SINNs are also assessed using standard metrics such as accuracy, efficiency, and generalizability, ensuring that interpretability is achieved alongside strong task performance.

\textbf{Explanation Evaluation.} Assessing explanation quality requires multiple perspectives, including stability, faithfulness, and inter-relationships among explanation units:
    (1) \textit{Stability} refers to whether explanations remain consistent under small variations in input. Ideally, similar inputs should yield similar explanations, ensuring robustness and avoiding inconsistencies. Common quantitative metrics for comparing explanations from slightly perturbed counterparts include rank order correlation~\cite{alvarez2018towards, verma2019lirme}, top-k intersection~\cite{ghorbani2019interpretation}, cosine similarity~\cite{chu2018exact, wang2021self}, and structural similarity~\cite{huang2023evaluation, carmichael2024pixel, sacha2024interpretability}.
    (2) \textit{Faithfulness} assesses how accurately the model's explanations reflect the actual importance. One approach involves comparing the model-generated explanations with ground truth explanations, when available. For example, concept-based SINNs measure faithfulness by matching learned concept representations with ground truth concept labels~\cite{espinosa2022concept, panousis2024coarse}. When ground truth is unavailable~\cite{zhou2023solvability}, an alternative strategy evaluates the impact on model performance when using or excluding highlighted explanations~\cite{yuan2022explainability, pope2019explainability, yeh2020completeness, qian2024towards}. 
    (3) \textit{Inter-relationship} evaluates interactions between different explanation units, ensuring explanations form coherent and meaningful structures rather than fragmented insights. This is particularly crucial for concept-based and prototype-based SINNs that identify high-level cognitive units. Explanations should emphasize distinct yet relevant patterns, minimizing redundancy. Metrics for assessing inter-dependency and redundancy among learned explanations include the Silhouette Score~\cite{wang2024unveiling} and Oracle/Niche Impurity Score~\cite{zarlenga2023towards}.

\textbf{Human-centric Feedback.} Although quantitative metrics provide structured evaluations, human feedback remains essential for assessing interpretability. User studies, questionnaires, and interviews~\cite{rong2023towards, Yang2024TOIS} measure user understanding and trust in the model's explanations. Metrics such as plausibility, usefulness, sufficiency, and trust are commonly evaluated through direct user ratings~\cite{yang2023language}.

\section{Discussions}
\label{sec:discussion}
This section explores the relationship between post-hoc and self-interpretable methods and suggests potential directions.

\subsection{Relationship between post-hoc and self-interpretation}
As two essential directions in XAI, both post-hoc interpretation and SINNs serve important but distinct purposes. Post-hoc methods excel at explaining existing models, while SINNs are designed for scenarios where interpretability is required from the ground up or when discovering patterns in data is the primary goal. Despite these differences, these two paradigms have intrinsic connections and can complement each other.

\textbf{Similar Interpretation Principles.}
Self-interpretable and post-hoc methods could share fundamental principles for model explanation. For example, attribution-based SINNs~\cite{alvarez2018towards} and post-hoc attribution methods~\cite{sundararajan2017axiomatic} both measure the contribution of individual features to the model output. Similarly, rule-based SINNs~\cite{barbiero2022entropy} and post-hoc rule extraction methods~\cite{setiono1995understanding} aim to generate human-readable rules that explain model decisions.
The detailed theoretical basis and core principles can also be shared across paradigms. For instance, both SASANet~\cite{sun2023towards} and SHAP~\cite{scott2017unified} leverage Shapley values for interpretation. Despite differences in how explanations are generated, these approaches can inspire each other and exchange their inherent interpretability insights.

\textbf{Self-interpretation for Post-hoc Analysis.}
Recent research has explored model distillation to transform black-box models into interpretable surrogates, such as decision trees~\cite{ghosh2023dividing, zhang2019interpreting, song2021tree}. This distillation process maps the original model's internal representations to more transparent decision structures. A key challenge with traditional interpretable surrogates is their limited expressiveness, which may fail to accurately capture the logic of the original network. The emergence of more flexible and expressive SINNs~\cite{swamy2025intrinsic, choi2024adaptive} offers promising solutions, enabling complex post-hoc explanations from multiple perspectives that better reflect the original model's decision-making process.

\textbf{SINNs as Diagnostic Tools for Post-hoc Methods.}
Although SINNs may have limited applicability compared to post-hoc methods, they have the potential to serve as valuable diagnostic tools. By training models with provably correct self-interpretation (such as Neural Additive Models~\cite{agarwal2021neural}), researchers can establish reliable ground truth interpretations. These potentially can be used to evaluate post-hoc interpretation methods by comparing their outputs against known correct interpretations~\cite{sun2023towards, enoueninstashap}, provided both approaches produce comparable forms of interpretation.

\textbf{Post-hoc Methods Enhancing Self-interpretable Models.}
While self-interpretable models strive for transparency, many methods incorporate approximations and less transparent components to maintain expressiveness. This makes post-hoc methods valuable diagnostic tools for identifying limitations in SINNs~\cite{krishna2024post, liu2025utilizing}. For example, in concept-based methods, the concept encoding process may be complicated for understanding. In such cases, post-hoc methods can complement SINNs by providing additional analysis to enhance robustness and interpretability.

\subsection{Potential Future Directions}
Despite the significant progress in SINNs, several critical challenges and exciting opportunities still remain. In this section, we highlight several promising research directions that could bridge existing gaps and advance this research field.

\textbf{Systematic Evaluation Metrics and Benchmarks.} 
A critical challenge in advancing SINNs is the development of rigorous evaluation frameworks.
While considerable progress has been made in application-specific model designs~\cite{huang2023evaluation, sacha2024interpretability}, standardized evaluation metrics still remain underdeveloped. 
Current approaches primarily access a single dimension (either faithfulness or stability), but fail to provide a comprehensive evaluation of interpretability quality.
We highlight the need for benchmarks that not only facilitate model comparison but also help identify flaws in interpretability mechanisms, promoting research reproducibility. 
As discussed in~\Cref{sec:evaluation}, model explanations should demonstrate robustness to adversarial examples, consistency across similar inputs, and alignment with the faithful factors, among other considerations. 
A comprehensive interpretability evaluation framework will be key to ensuring the reliability and real-world applicability of SINNs.

\textbf{Hybrid Model Interpretability.} Combining multiple interpretability approaches offers promising opportunities for enhancing model explanations. 
As demonstrated in~\Cref{sec:hybrid} and \Cref{sec:applications}, hierarchical explanations can offer deeper insights by providing multi-level interpretations. For example, in image classification, a hierarchical approach can first identify relevant objects and then explain their interactions through rules or global concepts to arrive at the final prediction~\cite{nauta2021neural, kim2022vit}. 
However, existing SINNs primarily focus on single-level explanations, leaving hybrid model interpretability largely unexplored. Future research should develop more sophisticated hybrid architectures that flexibly integrate different SINN paradigms. Our survey aims to help readers better understand the relationships between different SINN modules and inspire the design of more effective hybrid models.

\textbf{Multi-Modal Self-Interpretation.}  While current SINNs primarily focus on single modality, there is significant potential for progress in multi-modal contexts. Future research should investigate how well-established single-modality techniques, such as concept vectors and prototype learning, can be adapted to multi-modal settings. This includes developing architectures that preserve interpretability across modality interactions and designing tailored evaluation metrics for multi-modal interpretability. This could lead to more comprehensive explanations in multi-modal applications, such as autonomous driving and medical imaging. 

\textbf{Enhancing SINNs with LLMs.} The expressiveness of LLMs presents transformative opportunities for enhancing SINNs in four key directions.
First, LLMs can improve the interpretability of SINNs by generating more elaborate and nuanced natural language explanations~\cite{quan2024verification}. While SINNs are inherently self-interpretable, LLMs can augment this by providing richer explanations that complement the traditional self-interpretable components.
Second, LLMs can enhance data processing in SINNs by providing additional context for understanding patterns~\cite{srivastava2024vlg} and reducing manual data annotation needs~\cite{yang2023language}. This is particularly valuable for dataset explanation, where LLMs can help identify patterns and generate interactive explanations that bridge complex domains.
Third, integrating LLMs with SINNs, as demonstrated in recent works~\cite{singh2023augmenting, bostrom2024protolm, gandhi2024large, tan2024sparsity}, can enhance model generalization by leveraging the ability of LLMs to capture complex patterns. Specifically, LLMs can function as pre-trained backbones for encoding input data or structuring decision branches within SINNs, thereby improving their learning and generalization capabilities while preserving interpretability. 
Fourth, developing self-interpretable LLMs, which combine the expressiveness of LLMs with transparent reasoning mechanisms, is an important frontier for future research. While LLMs can generate textual explanations of their outputs, their internal mechanisms remain largely opaque~\cite{madsen2024self}, posing risks related to hallucination, bias, and unreliability. Existing methods mainly focus on attention visualization or post-hoc rationalization~\cite{huben2023sparse, dunefskytranscoders}, which do not necessarily reflect the actual reasoning process. It is promising to explore how to incorporate self-interpretable mechanisms (e.g., explicit rule learning, concept-based representations) into LLMs~\cite{ismail2024concept, sun2024concept} to enhance their transparency and reliability.

\section{Conclusion}
\label{sec:conclusion}
In this paper, we presented a comprehensive survey of self-interpretable neural networks, focusing on attribution-based, function-based, concept-based, prototype-based, and rule-based approaches. We provided concrete, visualized examples of model explanations and discussed their applicability across various domains, including image, text, graph, and deep reinforcement learning. Additionally, we summarized the quantitative evaluation metrics for SINNs. 
We also explored the relationship between post-hoc and self-interpretable methods, emphasizing the potential for mutual benefit. We hope this survey will inspire future research in the field of self-interpretable neural networks and contribute to the development of wide interpretable machine learning models.

\ifCLASSOPTIONcompsoc
  \section*{Acknowledgments}
\else
  \section*{Acknowledgment}
\fi

This work is partly supported by the National Key Research and Development Program of China (No. 2023YFF0725001), the National Natural Science Foundation of China (No. 62306255, 92370204, 62176014), the Natural Science Foundation of Guangdong Province (No. 2024A1515011839), the Fundamental Research Project of Guangzhou (No. 2024A04J4233), the Guangzhou-HKUST(GZ) Joint Funding Program (No.2023A03J0008), and the Education Bureau of Guangzhou Municipality. 
\ifCLASSOPTIONcaptionsoff
  \newpage
\fi

\bibliographystyle{IEEEtran}
\bibliography{ref}

\begin{thebibliography}{100}
\providecommand{\url}[1]{#1}
\csname url@samestyle\endcsname
\providecommand{\newblock}{\relax}
\providecommand{\bibinfo}[2]{#2}
\providecommand{\BIBentrySTDinterwordspacing}{\spaceskip=0pt\relax}
\providecommand{\BIBentryALTinterwordstretchfactor}{4}
\providecommand{\BIBentryALTinterwordspacing}{\spaceskip=\fontdimen2\font plus
\BIBentryALTinterwordstretchfactor\fontdimen3\font minus
  \fontdimen4\font\relax}
\providecommand{\BIBforeignlanguage}[2]{{%
\expandafter\ifx\csname l@#1\endcsname\relax
\typeout{** WARNING: IEEEtran.bst: No hyphenation pattern has been}%
\typeout{** loaded for the language `#1'. Using the pattern for}%
\typeout{** the default language instead.}%
\else
\language=\csname l@#1\endcsname
\fi
#2}}
\providecommand{\BIBdecl}{\relax}
\BIBdecl

\bibitem{barnett2021case}
A.~J. Barnett, F.~R. Schwartz, C.~Tao, C.~Chen, Y.~Ren, J.~Y. Lo, and C.~Rudin,
  ``A case-based interpretable deep learning model for classification of mass
  lesions in digital mammography,'' \emph{Nat. Mach. Intell.}, pp. 1061--1070,
  2021.

\bibitem{bai2023interpretable}
P.~Bai, F.~Miljkovi{\'c}, B.~John, and H.~Lu, ``Interpretable bilinear
  attention network with domain adaptation improves drug--target prediction,''
  \emph{Nat. Mach. Intell.}, vol.~5, no.~2, pp. 126--136, 2023.

\bibitem{arrieta2020explainable}
A.~B. Arrieta, N.~Díaz-Rodríguez, J.~Del~Ser, A.~Bennetot, S.~Tabik,
  A.~Barbado, S.~García, S.~Gil-López, D.~Molina, R.~Benjamins \emph{et~al.},
  ``Explainable artificial intelligence (xai): Concepts, taxonomies,
  opportunities and challenges toward responsible ai,'' \emph{Inf. Fusion}, pp.
  82--115, 2020.

\bibitem{lipton2016mythos}
Z.~C. Lipton, ``The mythos of model interpretability,'' in \emph{ICML Workshop
  on Human Interpretability in Machine Learning (WHI)}, 2016.

\bibitem{doshi2017towards}
F.~Doshi-Velez and B.~Kim, ``Towards a rigorous science of interpretable
  machine learning,'' \emph{arXiv preprint arXiv:1702.08608}, 2017.

\bibitem{dwivedi2023explainable}
R.~Dwivedi, D.~Dave, H.~Naik, S.~Singhal, R.~Omer, P.~Patel, B.~Qian, Z.~Wen,
  T.~Shah, G.~Morgan \emph{et~al.}, ``Explainable ai (xai): Core ideas,
  techniques, and solutions,'' \emph{ACM Comput. Surv.}, pp. 1--33, 2023.

\bibitem{ribeiro2016should}
M.~T. Ribeiro, S.~Singh, and C.~Guestrin, ``" why should i trust you?"
  explaining the predictions of any classifier,'' in \emph{SIGKDD}, 2016, pp.
  1135--1144.

\bibitem{scott2017unified}
M.~Scott, L.~Su-In \emph{et~al.}, ``A unified approach to interpreting model
  predictions,'' \emph{NeurIPS}, vol.~30, pp. 4765--4774, 2017.

\bibitem{shrikumar2017learning}
A.~Shrikumar, P.~Greenside, and A.~Kundaje, ``Learning important features
  through propagating activation differences,'' in \emph{ICML}, 2017, pp.
  3145--3153.

\bibitem{ghorbani2019interpretation}
A.~Ghorbani, A.~Abid, and J.~Zou, ``Interpretation of neural networks is
  fragile,'' in \emph{AAAI}, 2019, pp. 3681--3688.

\bibitem{slack2020fooling}
D.~Slack, S.~Hilgard, E.~Jia, S.~Singh, and H.~Lakkaraju, ``Fooling lime and
  shap: Adversarial attacks on post hoc explanation methods,'' in \emph{AIES},
  2020, pp. 180--186.

\bibitem{slack2021reliable}
D.~Slack, A.~Hilgard, S.~Singh, and H.~Lakkaraju, ``Reliable post hoc
  explanations: Modeling uncertainty in explainability,'' \emph{NeurIPS}, pp.
  9391--9404, 2021.

\bibitem{li2024graph}
J.~Li, M.~Pang, Y.~Dong, J.~Jia, and B.~Wang, ``Graph neural network
  explanations are fragile,'' in \emph{ICML}, 2024.

\bibitem{laugel2019dangers}
T.~Laugel, M.-J. Lesot, C.~Marsala, X.~Renard, and M.~Detyniecki, ``The dangers
  of post-hoc interpretability: Unjustified counterfactual explanations,'' in
  \emph{IJCAI}, 2019, pp. 2801--2807.

\bibitem{frye2020asymmetric}
C.~Frye, C.~Rowat, and I.~Feige, ``Asymmetric shapley values: incorporating
  causal knowledge into model-agnostic explainability,'' \emph{NeurIPS},
  vol.~33, pp. 1229--1239, 2020.

\bibitem{mullenbach2018explainable}
J.~Mullenbach, S.~Wiegreffe, J.~Duke, J.~Sun, and J.~Eisenstein, ``Explainable
  prediction of medical codes from clinical text,'' in \emph{NAACL}, 2018, pp.
  1101--1111.

\bibitem{jiang2024protogate}
X.~Jiang, A.~Margeloiu, N.~Simidjievski, and M.~Jamnik, ``Protogate:
  Prototype-based neural networks with global-to-local feature selection for
  tabular biomedical data,'' in \emph{ICML}, 2024.

\bibitem{luo2018beyond}
L.~Luo, X.~Ao, F.~Pan, J.~Wang, T.~Zhao, N.~Yu, and Q.~He, ``Beyond polarity:
  Interpretable financial sentiment analysis with hierarchical query-driven
  attention,'' in \emph{IJCAI}, 2018, pp. 4244--4250.

\bibitem{wong2024discovery}
F.~Wong, E.~J. Zheng, J.~A. Valeri, N.~M. Donghia, M.~N. Anahtar, S.~Omori,
  A.~Li, A.~Cubillos-Ruiz, A.~Krishnan, W.~Jin \emph{et~al.}, ``Discovery of a
  structural class of antibiotics with explainable deep learning,''
  \emph{Nature}, vol. 626, no. 7997, pp. 177--185, 2024.

\bibitem{sun2021market}
Y.~Sun, F.~Zhuang, H.~Zhu, Q.~Zhang, Q.~He, and H.~Xiong, ``Market-oriented job
  skill valuation with cooperative composition neural network,'' \emph{Nat.
  Commun.}, vol.~12, no.~1, pp. 1--12, 2021.

\bibitem{Yang2024TOIS}
Y.~Sun, Y.~Ji, H.~Zhu, F.~Zhuang, Q.~He, and H.~Xiong, ``Market-aware long-term
  job skill recommendation with explainable deep reinforcement learning,''
  \emph{TOIS}, 2024.

\bibitem{wong2021using}
C.~Duckworth, F.~P. Chmiel, D.~K. Burns, Z.~D. Zlatev, N.~M. White, T.~W.
  Daniels, M.~Kiuber, and M.~J. Boniface, ``Using explainable machine learning
  to characterise data drift and detect emergent health risks for emergency
  department admissions during covid-19,'' \emph{Scientific reports}, vol.~11,
  no.~1, p. 23017, 2021.

\bibitem{bontempelli2022concept}
A.~Bontempelli, S.~Teso, K.~Tentori, F.~Giunchiglia, A.~Passerini
  \emph{et~al.}, ``Concept-level debugging of part-prototype networks,'' in
  \emph{ICLR}, 2023.

\bibitem{liu2024towards}
Y.~Liu, K.~Ding, Q.~Lu, F.~Li, L.~Y. Zhang, and S.~Pan, ``Towards
  self-interpretable graph-level anomaly detection,'' \emph{NeurIPS}, 2024.

\bibitem{wang2023data}
Y.~Wang, X.~Jin, and P.~Li, ``Data-efficient and interpretable tabular anomaly
  detection,'' in \emph{SIGKDD}, 2023, pp. 2338--2348.

\bibitem{rudin2019stop}
C.~Rudin, ``Stop explaining black box machine learning models for high stakes
  decisions and use interpretable models instead,'' \emph{Nat. Mach. Intell.},
  pp. 206--215, 2019.

\bibitem{yang2018deep}
Y.~Yang, I.~G. Morillo, and T.~M. Hospedales, ``Deep neural decision trees,''
  \emph{arXiv preprint arXiv:1806.06988}, 2018.

\bibitem{wan2020nbdt}
A.~Wan, L.~Dunlap, D.~Ho, J.~Yin, S.~Lee, S.~Petryk, S.~A. Bargal, and J.~E.
  Gonzalez, ``Nbdt: Neural-backed decision tree,'' in \emph{ICLR}, 2020.

\bibitem{rudin2022interpretable}
C.~Rudin, C.~Chen, Z.~Chen, H.~Huang, L.~Semenova, and C.~Zhong,
  ``Interpretable machine learning: Fundamental principles and 10 grand
  challenges,'' \emph{Stat. Surv.}, pp. 1--85, 2022.

\bibitem{alvarez2018towards}
D.~Alvarez~Melis and T.~Jaakkola, ``Towards robust interpretability with
  self-explaining neural networks,'' \emph{NeurIPS}, vol.~31, 2018.

\bibitem{yang2022unbox}
G.~Yang, Q.~Ye, and J.~Xia, ``Unbox the black-box for the medical explainable
  ai via multi-modal and multi-centre data fusion: A mini-review, two showcases
  and beyond,'' \emph{Inf. Fusion}, pp. 29--52, 2022.

\bibitem{burkart2021survey}
N.~Burkart and M.~F. Huber, ``A survey on the explainability of supervised
  machine learning,'' \emph{J. Artif. Intell. Res.}, pp. 245--317, 2021.

\bibitem{chander2024toward}
B.~Chander, C.~John, L.~Warrier, and K.~Gopalakrishnan, ``Toward trustworthy
  artificial intelligence (tai) in the context of explainability and
  robustness,'' \emph{ACM Comput. Surv.}, 2024.

\bibitem{gao2024going}
Y.~Gao, S.~Gu, J.~Jiang, S.~R. Hong, D.~Yu, and L.~Zhao, ``Going beyond xai: A
  systematic survey for explanation-guided learning,'' \emph{ACM Comput.
  Surv.}, pp. 1--39, 2024.

\bibitem{madsen2022post}
A.~Madsen, S.~Reddy, and S.~Chandar, ``Post-hoc interpretability for neural
  nlp: A survey,'' \emph{ACM Comput. Surv.}, pp. 1--42, 2022.

\bibitem{calderon2024behalf}
N.~Calderon and R.~Reichart, ``On behalf of the stakeholders: Trends in nlp
  model interpretability in the era of llms,'' \emph{arXiv preprint
  arXiv:2407.19200}, 2024.

\bibitem{lyu2024towards}
Q.~Lyu, M.~Apidianaki, and C.~Callison-Burch, ``Towards faithful model
  explanation in nlp: A survey,'' \emph{Comput. Linguist.}, pp. 1--67, 2024.

\bibitem{luo2024local}
S.~Luo, H.~Ivison, S.~C. Han, and J.~Poon, ``Local interpretations for
  explainable natural language processing: A survey,'' \emph{ACM Comput.
  Surv.}, pp. 1--36, 2024.

\bibitem{danilevsky2020survey}
M.~Danilevsky, K.~Qian, R.~Aharonov, Y.~Katsis, B.~Kawas, and P.~Sen, ``A
  survey of the state of explainable ai for natural language processing,'' in
  \emph{AACL}, 2020.

\bibitem{zini2022explainability}
J.~E. Zini and M.~Awad, ``On the explainability of natural language processing
  deep models,'' \emph{ACM Comput. Surv.}, pp. 1--31, 2022.

\bibitem{agarwal2021towards}
C.~Agarwal, M.~Zitnik, and H.~Lakkaraju, ``Towards a rigorous theoretical
  analysis and evaluation of gnn explanations,'' \emph{arXiv preprint
  arXiv:2106.09078}, 2021.

\bibitem{yuan2022explainability}
H.~Yuan, H.~Yu, S.~Gui, and S.~Ji, ``Explainability in graph neural networks: A
  taxonomic survey,'' \emph{TPAMI}, pp. 5782--5799, 2022.

\bibitem{kakkad2023survey}
J.~Kakkad, J.~Jannu, K.~Sharma, C.~Aggarwal, and S.~Medya, ``A survey on
  explainability of graph neural networks,'' \emph{Bull. IEEE Comp. Soc. Tech.
  Comm. Data Eng.}, pp. 35--63, 2023.

\bibitem{qing2022survey}
Y.~Qing, S.~Liu, J.~Song, H.~Wang, and M.~Song, ``A survey on explainable
  reinforcement learning: Concepts, algorithms, challenges,'' \emph{arXiv
  preprint arXiv:2211.06665}, 2022.

\bibitem{vouros2022explainable}
G.~A. Vouros, ``Explainable deep reinforcement learning: state of the art and
  challenges,'' \emph{ACM Comput. Surv.}, pp. 1--39, 2022.

\bibitem{hickling2023explainability}
T.~Hickling, A.~Zenati, N.~Aouf, and P.~Spencer, ``Explainability in deep
  reinforcement learning: A review into current methods and applications,''
  \emph{ACM Comput. Surv.}, pp. 1--35, 2023.

\bibitem{milani2024explainable}
S.~Milani, N.~Topin, M.~Veloso, and F.~Fang, ``Explainable reinforcement
  learning: A survey and comparative review,'' \emph{ACM Comput. Surv.}, pp.
  1--36, 2024.

\bibitem{al2020contextual}
M.~Al-Shedivat, A.~Dubey, and E.~Xing, ``Contextual explanation networks,''
  \emph{JMLR}, pp. 1--44, 2020.

\bibitem{moshkovitz2020explainable}
M.~Moshkovitz, S.~Dasgupta, C.~Rashtchian, and N.~Frost, ``Explainable k-means
  and k-medians clustering,'' in \emph{ICML}, 2020, pp. 7055--7065.

\bibitem{qin2023reliable}
Z.~Qin, L.~Yang, Q.~Wang, Y.~Han, and Q.~Hu, ``Reliable and interpretable
  personalized federated learning,'' in \emph{CVPR}, 2023, pp.
  20\,422--20\,431.

\bibitem{norcliffe2018learning}
W.~Norcliffe-Brown, S.~Vafeias, and S.~Parisot, ``Learning conditioned graph
  structures for interpretable visual question answering,'' \emph{NeurIPS},
  vol.~31, 2018.

\bibitem{miller2019explanation}
T.~Miller, ``Explanation in artificial intelligence: Insights from the social
  sciences,'' \emph{Artif. Intell.}, pp. 1--38, 2019.

\bibitem{langer2021we}
M.~Langer, D.~Oster, T.~Speith, H.~Hermanns, L.~Kästner, E.~Schmidt,
  A.~Sesing, and K.~Baum, ``What do we want from explainable artificial
  intelligence (xai)?--a stakeholder perspective on xai and a conceptual model
  guiding interdisciplinary xai research,'' \emph{Artif. Intell.}, 2021.

\bibitem{mueller2019explanation}
S.~T. Mueller, R.~R. Hoffman, W.~Clancey, A.~Emrey, and G.~Klein, ``Explanation
  in human-ai systems: A literature meta-review, synopsis of key ideas and
  publications, and bibliography for explainable ai,'' \emph{arXiv preprint
  arXiv:1902.01876}, 2019.

\bibitem{saeed2023explainable}
W.~Saeed and C.~Omlin, ``Explainable ai (xai): A systematic meta-survey of
  current challenges and future opportunities,'' \emph{Knowl. Based Syst.},
  2023.

\bibitem{vilone2021notions}
G.~Vilone and L.~Longo, ``Notions of explainability and evaluation approaches
  for explainable artificial intelligence,'' \emph{Inf. Fusion}, pp. 89--106,
  2021.

\bibitem{zhang2021survey}
Y.~Zhang, P.~Tiño, A.~Leonardis, and K.~Tang, ``A survey on neural network
  interpretability,'' \emph{IEEE Trans. Emerg. Topics Comput. Intell.}, pp.
  726--742, 2021.

\bibitem{cambria2023survey}
E.~Cambria, L.~Malandri, F.~Mercorio, M.~Mezzanzanica, and N.~Nobani, ``A
  survey on xai and natural language explanations,'' \emph{Inf. Process.
  Manag.}, 2023.

\bibitem{zhao2024explainability}
H.~Zhao, H.~Chen, F.~Yang, N.~Liu, H.~Deng, H.~Cai, S.~Wang, D.~Yin, and M.~Du,
  ``Explainability for large language models: A survey,'' \emph{ACM Trans.
  Intell. Syst. Technol.}, pp. 1--38, 2024.

\bibitem{afchar2021towards}
D.~Afchar, V.~Guigue, and R.~Hennequin, ``Towards rigorous interpretations: a
  formalisation of feature attribution,'' in \emph{ICML}, 2021, pp. 76--86.

\bibitem{schulzrestricting}
K.~Schulz, L.~Sixt, F.~Tombari, and T.~Landgraf, ``Restricting the flow:
  Information bottlenecks for attribution,'' in \emph{ICLR}, 2020.

\bibitem{bohle2021convolutional}
M.~Bohle, M.~Fritz, and B.~Schiele, ``Convolutional dynamic alignment networks
  for interpretable classifications,'' in \emph{CVPR}, 2021, pp.
  10\,029--10\,038.

\bibitem{bohle2024b}
M.~Böhle, N.~Singh, M.~Fritz, and B.~Schiele, ``B-cos alignment for inherently
  interpretable cnns and vision transformers,'' \emph{TPAMI}, 2024.

\bibitem{bohle2022b}
M.~Böhle, M.~Fritz, and B.~Schiele, ``B-cos networks: Alignment is all we need
  for interpretability,'' in \emph{CVPR}, 2022, pp. 10\,329--10\,338.

\bibitem{zhang2022query}
Y.~Zhang, M.~Jiang, and Q.~Zhao, ``Query and attention augmentation for
  knowledge-based explainable reasoning,'' in \emph{CVPR}, 2022, pp.
  15\,576--15\,585.

\bibitem{zhu2024propagation}
J.~Zhu, C.~Gao, Z.~Yin, X.~Li, and J.~Kurths, ``Propagation structure-aware
  graph transformer for robust and interpretable fake news detection,'' in
  \emph{KDD}, 2024, pp. 4652--4663.

\bibitem{yeh2019interpretable}
C.-H. Yeh, Y.-C. Fan, and W.-C. Peng, ``Interpretable multi-task learning for
  product quality prediction with attention mechanism,'' in \emph{ICDE}, 2019,
  pp. 1910--1921.

\bibitem{ren2022diversified}
L.~Ren, G.~Yu, J.~Wang, L.~Liu, C.~Domeniconi, and X.~Zhang, ``A diversified
  attention model for interpretable multiple clusterings,'' \emph{TKDE},
  vol.~35, no.~9, pp. 8852--8864, 2022.

\bibitem{wu2021evidence}
L.~Wu, Y.~Rao, X.~Yang, W.~Wang, and A.~Nazir, ``Evidence-aware hierarchical
  interactive attention networks for explainable claim verification,'' in
  \emph{IJCAI}, 2021, pp. 1388--1394.

\bibitem{xu2022towards}
X.~Xu, Z.~Wang, C.~Deng, H.~Yuan, and S.~Ji, ``Towards improved and
  interpretable deep metric learning via attentive grouping,'' \emph{TPAMI},
  vol.~45, no.~1, pp. 1189--1200, 2022.

\bibitem{zhu2020modeling}
Y.~Zhu, D.~Xi, B.~Song, F.~Zhuang, S.~Chen, X.~Gu, and Q.~He, ``Modeling users'
  behavior sequences with hierarchical explainable network for cross-domain
  fraud detection,'' in \emph{WWW}, 2020, pp. 928--938.

\bibitem{arik2021tabnet}
S.~O. Arik and T.~Pfister, ``Tabnet: Attentive interpretable tabular
  learning,'' in \emph{AAAI}, vol.~35, no.~8, 2021, pp. 6679--6687.

\bibitem{lemhadri2021lassonet}
I.~Lemhadri, F.~Ruan, L.~Abraham, and R.~Tibshirani, ``Lassonet: A neural
  network with feature sparsity,'' \emph{JMLR}, vol.~22, no. 127, pp. 1--29,
  2021.

\bibitem{he2019interpretable}
S.~He, X.~Li, V.~Sivakumar, and A.~Banerjee, ``Interpretable predictive
  modeling for climate variables with weighted lasso,'' in \emph{AAAI}, 2019,
  pp. 1385--1392.

\bibitem{zhang2025comprehensive}
X.~Zhang, D.~Lee, and S.~Wang, ``Comprehensive attribution: Inherently
  explainable vision model with feature detector,'' in \emph{ECCV}, 2025, pp.
  196--213.

\bibitem{thompson2023contextual}
R.~Thompson, A.~Dezfouli, and R.~Kohn, ``The contextual lasso: Sparse linear
  models via deep neural networks,'' \emph{NeurIPS}, pp. 19\,940--19\,961,
  2023.

\bibitem{dinh2020consistent}
V.~C. Dinh and L.~S. Ho, ``Consistent feature selection for analytic deep
  neural networks,'' \emph{NeurIPS}, pp. 2420--2431, 2020.

\bibitem{yoon2018invase}
J.~Yoon, J.~Jordon, and M.~Van~der Schaar, ``Invase: Instance-wise variable
  selection using neural networks,'' in \emph{ICLR}, 2018.

\bibitem{jethani2021have}
N.~Jethani, M.~Sudarshan, Y.~Aphinyanaphongs, and R.~Ranganath, ``Have we
  learned to explain?: How interpretability methods can learn to encode
  predictions in their interpretations,'' in \emph{AISTATS}, 2021, pp.
  1459--1467.

\bibitem{yang2022locally}
J.~Yang, O.~Lindenbaum, and Y.~Kluger, ``Locally sparse neural networks for
  tabular biomedical data,'' in \emph{ICML}, 2022, pp. 25\,123--25\,153.

\bibitem{yamada2020feature}
Y.~Yamada, O.~Lindenbaum, S.~Negahban, and Y.~Kluger, ``Feature selection using
  stochastic gates,'' in \emph{ICML}, 2020, pp. 10\,648--10\,659.

\bibitem{lee2022self}
S.~Lee, X.~Wang, S.~Han, X.~Yi, X.~Xie, and M.~Cha, ``Self-explaining deep
  models with logic rule reasoning,'' \emph{NeurIPS}, pp. 3203--3216, 2022.

\bibitem{li2020mri}
W.~Li, X.~Feng, H.~An, X.~Y. Ng, and Y.-J. Zhang, ``Mri reconstruction with
  interpretable pixel-wise operations using reinforcement learning,'' in
  \emph{AAAI}, 2020, pp. 792--799.

\bibitem{leonhardt2023extractive}
J.~Leonhardt, K.~Rudra, and A.~Anand, ``Extractive explanations for
  interpretable text ranking,'' \emph{TOIS}, pp. 1--31, 2023.

\bibitem{chen2018learning}
J.~Chen, L.~Song, M.~Wainwright, and M.~Jordan, ``Learning to explain: An
  information-theoretic perspective on model interpretation,'' in \emph{ICML},
  2018, pp. 883--892.

\bibitem{balin2019concrete}
M.~F. Balın, A.~Abid, and J.~Zou, ``Concrete autoencoders: Differentiable
  feature selection and reconstruction,'' in \emph{ICML}, 2019, pp. 444--453.

\bibitem{si2024interpretabnet}
J.~Si, W.~Y. Cheng, M.~Cooper, and R.~G. Krishnan, ``Interpretabnet: Distilling
  predictive signals from tabular data by salient feature interpretation,'' in
  \emph{ICML}, 2024.

\bibitem{yu2020graph}
J.~Yu, T.~Xu, Y.~Rong, Y.~Bian, J.~Huang, and R.~He, ``Graph information
  bottleneck for subgraph recognition,'' in \emph{ICLR}, 2020.

\bibitem{yu2021recognizing}
------, ``Recognizing predictive substructures with subgraph information
  bottleneck,'' \emph{TPAMI}, vol.~46, no.~3, pp. 1650--1663, 2021.

\bibitem{seo2024interpretable}
S.~Seo, S.~Kim, and C.~Park, ``Interpretable prototype-based graph information
  bottleneck,'' \emph{NeurIPS}, vol.~36, 2024.

\bibitem{fang2024exgc}
J.~Fang, X.~Li, Y.~Sui, Y.~Gao, G.~Zhang, K.~Wang, X.~Wang, and X.~He, ``Exgc:
  Bridging efficiency and explainability in graph condensation,'' in
  \emph{WWW}, 2024, pp. 721--732.

\bibitem{chen2024tempme}
J.~Chen and R.~Ying, ``Tempme: Towards the explainability of temporal graph
  neural networks via motif discovery,'' \emph{NeurIPS}, 2024.

\bibitem{miao2022interpretable}
S.~Miao, M.~Liu, and P.~Li, ``Interpretable and generalizable graph learning
  via stochastic attention mechanism,'' in \emph{ICML}, 2022, pp.
  15\,524--15\,543.

\bibitem{sekhon2023improving}
A.~Sekhon, H.~Chen, A.~Shrivastava, Z.~Wang, Y.~Ji, and Y.~Qi, ``Improving
  interpretability via explicit word interaction graph layer,'' in \emph{AAAI},
  2023, pp. 13\,528--13\,537.

\bibitem{miaointerpretable2023}
S.~Miao, Y.~Luo, M.~Liu, and P.~Li, ``Interpretable geometric deep learning via
  learnable randomness injection,'' in \emph{ICLR}, 2023.

\bibitem{changnode}
C.-H. Chang, R.~Caruana, and A.~Goldenberg, ``Node-gam: Neural generalized
  additive model for interpretable deep learning,'' in \emph{ICLR}, 2024.

\bibitem{ibrahim2024grand}
S.~Ibrahim, G.~Afriat, K.~Behdin, and R.~Mazumder, ``Grand-slamin'interpretable
  additive modeling with structural constraints,'' \emph{NeurIPS}, vol.~36,
  2024.

\bibitem{dubey2022scalable}
A.~Dubey, F.~Radenovic, and D.~Mahajan, ``Scalable interpretability via
  polynomials,'' \emph{NeurIPS}, vol.~35, pp. 36\,748--36\,761, 2022.

\bibitem{radenovic2022neural}
F.~Radenovic, A.~Dubey, and D.~Mahajan, ``Neural basis models for
  interpretability,'' \emph{NeurIPS}, vol.~35, pp. 8414--8426, 2022.

\bibitem{jiao2023naisr}
Y.~Jiao, C.~Zdanski, J.~Kimbell, A.~Prince, C.~Worden, S.~Kirse, C.~Rutter,
  B.~Shields, W.~Dunn, J.~Mahmud \emph{et~al.}, ``Naisr: A 3d neural additive
  model for interpretable shape representation,'' in \emph{ICLR}, 2024.

\bibitem{liu2023n}
Z.~Liu, Y.~Zhu, and C.~Chen, ``N$\text{A}^\text{2}$q: Neural attention additive
  model for interpretable multi-agent q-learning,'' in \emph{ICML}, 2023, pp.
  22\,539--22\,558.

\bibitem{brendel2019approximating}
W.~Brendel and M.~Bethge, ``Approximating cnns with bag-of-local-features
  models works surprisingly well on imagenet,'' in \emph{ICLR}, 2018.

\bibitem{tsang2018neural}
M.~Tsang, H.~Liu, S.~Purushotham, P.~Murali, and Y.~Liu, ``Neural interaction
  transparency (nit): Disentangling learned interactions for improved
  interpretability,'' \emph{NeurIPS}, 2018.

\bibitem{xu2022sparse}
S.~Xu, Z.~Bu, P.~Chaudhari, and I.~J. Barnett, ``Sparse neural additive model:
  Interpretable deep learning with feature selection via group sparsity,'' in
  \emph{ICLR}, 2023.

\bibitem{liu2020sparse}
G.~Liu, H.~Chen, and H.~Huang, ``Sparse shrunk additive models,'' in
  \emph{ICML}, 2020, pp. 6194--6204.

\bibitem{bouchiat2023improving}
K.~Bouchiat, A.~Immer, H.~Yèche, G.~Ratsch, and V.~Fortuin, ``Improving neural
  additive models with bayesian principles,'' in \emph{ICML}, 2023.

\bibitem{enouen2022sparse}
J.~Enouen and Y.~Liu, ``Sparse interaction additive networks via feature
  interaction detection and sparse selection,'' \emph{NeurIPS}, vol.~35, pp.
  13\,908--13\,920, 2022.

\bibitem{yang2021gami}
Z.~Yang, A.~Zhang, and A.~Sudjianto, ``Gami-net: An explainable neural network
  based on generalized additive models with structured interactions,''
  \emph{Pattern Recognit.}, vol. 120, p. 108192, 2021.

\bibitem{sun2023towards}
Y.~Sun, H.~Zhu, and H.~Xiong, ``Towards faithful neural network intrinsic
  interpretation with shapley additive self-attribution,'' \emph{arXiv preprint
  arXiv:2309.15559}, 2023.

\bibitem{wang2021shapley}
R.~Wang, X.~Wang, and D.~Inouye, ``Shapley explanation networks,'' in
  \emph{ICLR}, 2021.

\bibitem{guo2021edge}
W.~Guo, X.~Wu, U.~Khan, and X.~Xing, ``Edge: Explaining deep reinforcement
  learning policies,'' \emph{NeurIPS}, vol.~34, pp. 12\,222--12\,236, 2021.

\bibitem{quinn2020deepcoda}
T.~Quinn, D.~Nguyen, S.~Rana, S.~Gupta, and S.~Venkatesh, ``Deepcoda:
  personalized interpretability for compositional health data,'' in
  \emph{ICML}, 2020, pp. 7877--7886.

\bibitem{wang2021self}
Y.~Wang and X.~Wang, ``Self-interpretable model with transformation equivariant
  interpretation,'' \emph{NeurIPS}, vol.~34, pp. 2359--2372, 2021.

\bibitem{martins2016softmax}
A.~Martins and R.~Astudillo, ``From softmax to sparsemax: A sparse model of
  attention and multi-label classification,'' in \emph{ICML}, 2016, pp.
  1614--1623.

\bibitem{correia2019adaptively}
G.~M. Correia, V.~Niculae, and A.~F. Martins, ``Adaptively sparse
  transformers,'' in \emph{EMNLP}, 2019, pp. 2174--2184.

\bibitem{BahdanauCB14}
D.~Bahdanau, K.~Cho, and Y.~Bengio, ``Neural machine translation by jointly
  learning to align and translate,'' in \emph{ICLR}, 2015.

\bibitem{vaswani2017attention}
A.~Vaswani, ``Attention is all you need,'' \emph{NeurIPS}, 2017.

\bibitem{wiegreffe2019attention}
S.~Wiegreffe and Y.~Pinter, ``Attention is not not explanation,'' in
  \emph{EMNLP}, 2019, pp. 11--20.

\bibitem{serrano2019attention}
S.~Serrano and N.~A. Smith, ``Is attention interpretable?'' in \emph{ACL},
  2019.

\bibitem{meister2021sparse}
C.~Meister, S.~Lazov, I.~Augenstein, and R.~Cotterell, ``Is sparse attention
  more interpretable?'' in \emph{ACL}, 2021, pp. 122--129.

\bibitem{bibal2022attention}
A.~Bibal, R.~Cardon, D.~Alfter, R.~Wilkens, X.~Wang, T.~Fran{\c{c}}ois, and
  P.~Watrin, ``Is attention explanation? an introduction to the debate,'' in
  \emph{ACL}, 2022, pp. 3889--3900.

\bibitem{bai2021attentions}
B.~Bai, J.~Liang, G.~Zhang, H.~Li, K.~Bai, and F.~Wang, ``Why attentions may
  not be interpretable?'' in \emph{SIGKDD}, 2021, pp. 25--34.

\bibitem{brunner2019identifiability}
G.~Brunner, Y.~Liu, D.~Pascual, O.~Richter, M.~Ciaramita, and R.~Wattenhofer,
  ``On identifiability in transformers,'' in \emph{ICLR}, 2020.

\bibitem{sun2021effective}
K.~Sun and A.~Marasovi{'c}, ``Effective attention sheds light on
  interpretability,'' in \emph{ACL}, 2021, pp. 4126--4135.

\bibitem{kobayashi2020attention}
G.~Kobayashi, T.~Kuribayashi, S.~Yokoi, and K.~Inui, ``Attention is not only a
  weight: Analyzing transformers with vector norms,'' in \emph{EMNLP}, 2020.

\bibitem{hu2023seat}
L.~Hu, Y.~Liu, N.~Liu, M.~Huai, L.~Sun, and D.~Wang, ``Seat: stable and
  explainable attention,'' in \emph{AAAI}, 2023, pp. 12\,907--12\,915.

\bibitem{mohankumar2020towards}
A.~K. Mohankumar, P.~Nema, S.~Narasimhan, M.~M. Khapra, B.~V. Srinivasan, and
  B.~Ravindran, ``Towards transparent and explainable attention models,'' in
  \emph{ACL}, 2020, pp. 4206--4216.

\bibitem{nguyen2023learning}
T.~H. Nguyen and K.~Rudra, ``Learning faithful attention for interpretable
  classification of crisis-related microblogs under constrained human budget,''
  in \emph{WWW}, 2023, pp. 3959--3967.

\bibitem{pruthi2020learning}
D.~Pruthi, M.~Gupta, B.~Dhingra, G.~Neubig, and Z.~C. Lipton, ``Learning to
  deceive with attention-based explanations,'' in \emph{ACL}, 2020.

\bibitem{stacey2022supervising}
J.~Stacey, Y.~Belinkov, and M.~Rei, ``Supervising model attention with human
  explanations for robust natural language inference,'' in \emph{AAAI}, 2022,
  pp. 11\,349--11\,357.

\bibitem{rigotti2021attention}
M.~Rigotti, C.~Miksovic, I.~Giurgiu, T.~Gschwind, and P.~Scotton,
  ``Attention-based interpretability with concept transformers,'' in
  \emph{ICLR}, 2021.

\bibitem{li2019exploiting}
Y.~Li, S.~Lin, B.~Zhang, J.~Liu, D.~Doermann, Y.~Wu, F.~Huang, and R.~Ji,
  ``Exploiting kernel sparsity and entropy for interpretable cnn compression,''
  in \emph{CVPR}, 2019, pp. 2800--2809.

\bibitem{lei2016rationalizing}
T.~Lei, R.~Barzilay, and T.~Jaakkola, ``Rationalizing neural predictions,'' in
  \emph{EMNLP}, 2016, pp. 107--117.

\bibitem{jang2017categorical}
E.~Jang, S.~Gu, and B.~Poole, ``Categorical reparameterization with
  {Gumbel-Softmax},'' in \emph{ICLR}, 2017.

\bibitem{xie2019reparameterizable}
S.~M. Xie and S.~Ermon, ``Reparameterizable subset sampling via continuous
  relaxations,'' in \emph{IJCAI}, 2019, pp. 3919--3925.

\bibitem{alemi2022deep}
A.~A. Alemi, I.~Fischer, J.~V. Dillon, and K.~Murphy, ``Deep variational
  information bottleneck,'' in \emph{ICLR}, 2022.

\bibitem{agarwal2021neural}
R.~Agarwal, L.~Melnick, N.~Frosst, X.~Zhang, B.~Lengerich, R.~Caruana, and
  G.~E. Hinton, ``Neural additive models: Interpretable machine learning with
  neural nets,'' \emph{NeurIPS}, vol.~34, pp. 4699--4711, 2021.

\bibitem{hastie2017generalized}
T.~J. Hastie, ``Generalized additive models,'' in \emph{Statistical models in
  S}.\hskip 1em plus 0.5em minus 0.4em\relax Routledge, 2017, pp. 249--307.

\bibitem{sun2022puregam}
X.~Sun, Z.~Wang, R.~Ding, S.~Han, and D.~Zhang, ``Puregam: Learning an
  inherently pure additive model,'' in \emph{KDD}, 2022, pp. 1728--1738.

\bibitem{duong2024cat}
V.~Duong, Q.~Wu, Z.~Zhou, H.~Zhao, C.~Luo, E.~Zavesky, H.~Yao, and H.~Shao,
  ``Cat: Interpretable concept-based taylor additive models,'' in \emph{KDD},
  2024, pp. 723--734.

\bibitem{shapley1953value}
L.~S. Shapley, ``A value for n-person games,'' \emph{Contrib. Theory Games},
  vol.~2, 1953.

\bibitem{bento2021timeshap}
J.~Bento, P.~Saleiro, A.~F. Cruz, M.~A. Figueiredo, and P.~Bizarro, ``Timeshap:
  Explaining recurrent models through sequence perturbations,'' in
  \emph{SIGKDD}, 2021, pp. 2565--2573.

\bibitem{jiang2024seqshap}
G.~Jiang, F.~Zhuang, B.~Song, Y.~Zhu, Y.~Sun, W.~Wang, and D.~Wang, ``Seqshap:
  Subsequence level shapley value explanations for sequential predictions,'' in
  \emph{DASFAA}, 2024.

\bibitem{swamy2025intrinsic}
V.~Swamy, S.~Montariol, J.~Blackwell, J.~Frej, M.~Jaggi, and T.~K{\"a}ser,
  ``Intrinsic user-centric interpretability through global mixture of
  experts,'' in \emph{ICLR}, 2025.

\bibitem{sun2021discerning}
Y.~Sun, H.~Zhu, C.~Qin, F.~Zhuang, Q.~He, and H.~Xiong, ``Discerning
  decision-making process of deep neural networks with hierarchical voting
  transformation,'' \emph{NeurIPS}, vol.~34, pp. 17\,221--17\,234, 2021.

\bibitem{liu2024kan}
Z.~Liu, Y.~Wang, S.~Vaidya, F.~Ruehle, J.~Halverson, M.~Solja{\v{c}}i{\'c},
  T.~Y. Hou, and M.~Tegmark, ``Kan: Kolmogorov-arnold networks,'' \emph{arXiv
  preprint arXiv:2404.19756}, 2024.

\bibitem{kim2020integration}
S.~Kim, P.~Y. Lu, S.~Mukherjee, M.~Gilbert, L.~Jing, V.~Čeperić, and
  M.~Soljačić, ``Integration of neural network-based symbolic regression in
  deep learning for scientific discovery,'' \emph{TNNLS}, pp. 4166--4177, 2020.

\bibitem{ranasinghe2024ginn}
N.~Ranasinghe, D.~Senanayake, S.~Seneviratne, M.~Premaratne, and S.~Halgamuge,
  ``Ginn-lp: A growing interpretable neural network for discovering
  multivariate laurent polynomial equations,'' in \emph{AAAI}, 2024, pp.
  14\,776--14\,784.

\bibitem{kamienny2022end}
P.-A. Kamienny, S.~d'Ascoli, G.~Lample, and F.~Charton, ``End-to-end symbolic
  regression with transformers,'' \emph{NeurIPS}, pp. 10\,269--10\,281, 2022.

\bibitem{bendinelli2023controllable}
T.~Bendinelli, L.~Biggio, and P.-A. Kamienny, ``Controllable neural symbolic
  regression,'' in \emph{ICML}, 2023, pp. 2063--2077.

\bibitem{biggio2021neural}
L.~Biggio, T.~Bendinelli, A.~Neitz, A.~Lucchi, and G.~Parascandolo, ``Neural
  symbolic regression that scales,'' in \emph{ICML}, 2021, pp. 936--945.

\bibitem{valipour2021symbolicgpt}
M.~Valipour, B.~You, M.~Panju, and A.~Ghodsi, ``Symbolicgpt: A generative
  transformer model for symbolic regression,'' \emph{arXiv preprint
  arXiv:2106.14131}, 2021.

\bibitem{landajuela2022unified}
M.~Landajuela, C.~S. Lee, J.~Yang, R.~Glatt, C.~P. Santiago, I.~Aravena,
  T.~Mundhenk, G.~Mulcahy, and B.~K. Petersen, ``A unified framework for deep
  symbolic regression,'' \emph{NeurIPS}, pp. 33\,985--33\,998, 2022.

\bibitem{holtdeep}
S.~Holt, Z.~Qian, and M.~van~der Schaar, ``Deep generative symbolic
  regression,'' in \emph{ICLR}, 2023.

\bibitem{sahoo2018learning}
S.~Sahoo, C.~Lampert, and G.~Martius, ``Learning equations for extrapolation
  and control,'' in \emph{ICML}, 2018, pp. 4442--4450.

\bibitem{koh2020concept}
P.~W. Koh, T.~Nguyen, Y.~S. Tang, S.~Mussmann, E.~Pierson, B.~Kim, and
  P.~Liang, ``Concept bottleneck models,'' in \emph{ICML}, 2020, pp.
  5338--5348.

\bibitem{chen2020concept}
Z.~Chen, Y.~Bei, and C.~Rudin, ``Concept whitening for interpretable image
  recognition,'' \emph{Nat. Mach. Intell.}, vol.~2, no.~12, pp. 772--782, 2020.

\bibitem{blazek2021explainable}
P.~J. Blazek and M.~M. Lin, ``Explainable neural networks that simulate
  reasoning,'' \emph{Nat. Comput. Sci.}, pp. 607--618, 2021.

\bibitem{espinosa2022concept}
M.~Espinosa~Zarlenga, P.~Barbiero, G.~Ciravegna, G.~Marra, F.~Giannini,
  M.~Diligenti, Z.~Shams, F.~Precioso, S.~Melacci, A.~Weller \emph{et~al.},
  ``Concept embedding models: Beyond the accuracy-explainability trade-off,''
  \emph{NeurIPS}, pp. 21\,400--21\,413, 2022.

\bibitem{yuksekgonul2022post}
M.~Yuksekgonul, M.~Wang, and J.~Zou, ``Post-hoc concept bottleneck models,'' in
  \emph{ICLR}, 2022.

\bibitem{marconato2022glancenets}
E.~Marconato, A.~Passerini, and S.~Teso, ``Glancenets: Interpretable,
  leak-proof concept-based models,'' \emph{NeurIPS}, pp. 21\,212--21\,227,
  2022.

\bibitem{kim2023probabilistic}
E.~Kim, D.~Jung, S.~Park, S.~Kim, and S.~Yoon, ``Probabilistic concept
  bottleneck models,'' in \emph{ICML}, 2023, pp. 16\,521--16\,540.

\bibitem{havasi2022addressing}
M.~Havasi, S.~Parbhoo, and F.~Doshi-Velez, ``Addressing leakage in concept
  bottleneck models,'' \emph{NeurIPS}, pp. 23\,386--23\,397, 2022.

\bibitem{steinmann2023learning}
D.~Steinmann, W.~Stammer, F.~Friedrich, and K.~Kersting, ``Learning to
  intervene on concept bottlenecks,'' in \emph{ICML}, 2023.

\bibitem{chauhan2023interactive}
K.~Chauhan, R.~Tiwari, J.~Freyberg, P.~Shenoy, and K.~Dvijotham, ``Interactive
  concept bottleneck models,'' in \emph{AAAI}, 2023, pp. 5948--5955.

\bibitem{oikarinen2023label}
T.~Oikarinen, S.~Das, L.~Nguyen, and L.~Weng, ``Label-free concept bottleneck
  models,'' in \emph{ICLR}, 2023.

\bibitem{yang2023language}
Y.~Yang, A.~Panagopoulou, S.~Zhou, D.~Jin, C.~Callison-Burch, and M.~Yatskar,
  ``Language in a bottle: Language model guided concept bottlenecks for
  interpretable image classification,'' in \emph{CVPR}, 2023, pp.
  19\,187--19\,197.

\bibitem{panousis2024coarse}
K.~Panousis, D.~Ienco, and D.~Marcos, ``Coarse-to-fine concept bottleneck
  models,'' in \emph{NeurIPS}, 2024.

\bibitem{shang2024incremental}
C.~Shang, S.~Zhou, H.~Zhang, X.~Ni, Y.~Yang, and Y.~Wang, ``Incremental
  residual concept bottleneck models,'' in \emph{CVPR}, 2024, pp.
  11\,030--11\,040.

\bibitem{espinosa2024learning}
M.~Espinosa~Zarlenga, K.~Collins, K.~Dvijotham, A.~Weller, Z.~Shams, and
  M.~Jamnik, ``Learning to receive help: Intervention-aware concept embedding
  models,'' \emph{NeurIPS}, 2024.

\bibitem{xu2024energy}
X.~Xu, Y.~Qin, L.~Mi, H.~Wang, and X.~Li, ``Energy-based concept bottleneck
  models: unifying prediction, concept intervention, and conditional
  interpretations,'' in \emph{ICLR}, 2024.

\bibitem{vandenhirtz2024stochastic}
M.~Vandenhirtz, S.~Laguna, R.~Marcinkevičs, and J.~E. Vogt, ``Stochastic
  concept bottleneck models,'' \emph{NeurIPS}, 2024.

\bibitem{barbierorelational}
P.~Barbiero, F.~Giannini, G.~Ciravegna, M.~Diligenti, and G.~Marra,
  ``Relational concept bottleneck models,'' in \emph{NeurIPS}, 2024.

\bibitem{tan2024sparsity}
Z.~Tan, T.~Chen, Z.~Zhang, and H.~Liu, ``Sparsity-guided holistic explanation
  for llms with interpretable inference-time intervention,'' in \emph{AAAI},
  2024, pp. 21\,619--21\,627.

\bibitem{bie2024mica}
Y.~Bie, L.~Luo, and H.~Chen, ``Mica: Towards explainable skin lesion diagnosis
  via multi-level image-concept alignment,'' in \emph{AAAI}, 2024, pp.
  837--845.

\bibitem{dominici2024counterfactual}
G.~Dominici, P.~Barbiero, F.~Giannini, M.~Gjoreski, G.~Marra, and
  M.~Langheinrich, ``Counterfactual concept bottleneck models,'' in
  \emph{ICLR}, 2025.

\bibitem{dominici2024causal}
G.~Dominici, P.~Barbiero, M.~E. Zarlenga, A.~Termine, M.~Gjoreski, G.~Marra,
  and M.~Langheinrich, ``Causal concept graph models: Beyond causal opacity in
  deep learning,'' in \emph{ICLR}, 2025.

\bibitem{sun2024concept}
C.-E. Sun, T.~Oikarinen, B.~Ustun, and T.-W. Weng, ``Concept bottleneck large
  language models,'' in \emph{ICLR}, 2025.

\bibitem{ismail2024concept}
A.~A. Ismail, T.~Oikarinen, A.~Wang, J.~Adebayo, S.~Stanton, T.~Joren,
  J.~Kleinhenz, A.~Goodman, H.~C. Bravo, K.~Cho \emph{et~al.}, ``Concept
  bottleneck language models for protein design,'' in \emph{ICLR}, 2025.

\bibitem{parekh2024restyling}
J.~Parekh, Q.~Bouniot, P.~Mozharovskyi, A.~Newson, and F.~d'Alch{\'e} Buc,
  ``Restyling unsupervised concept based interpretable networks with generative
  models,'' in \emph{ICLR}, 2025.

\bibitem{luyten2024theoretical}
M.~R. Luyten and M.~van~der Schaar, ``A theoretical design of concept sets:
  improving the predictability of concept bottleneck models,'' in
  \emph{NeurIPS}, 2024.

\bibitem{raman2024understanding}
N.~J. Raman, M.~E. Zarlenga, and M.~Jamnik, ``Understanding inter-concept
  relationships in concept-based models,'' in \emph{ICML}, 2024.

\bibitem{yan2023towards}
S.~Yan, Z.~Yu, X.~Zhang, D.~Mahapatra, S.~S. Chandra, M.~Janda, P.~Soyer, and
  Z.~Ge, ``Towards trustable skin cancer diagnosis via rewriting model's
  decision,'' in \emph{CVPR}, 2023, pp. 11\,568--11\,577.

\bibitem{srivastava2024vlg}
D.~Srivastava, G.~Yan, and T.-W. Weng, ``Vlg-cbm: Training concept bottleneck
  models with vision-language guidance,'' in \emph{NeurIPS}, 2024.

\bibitem{van2010example}
T.~Van~Gog and N.~Rummel, ``Example-based learning: Integrating cognitive and
  social-cognitive research perspectives,'' \emph{Educ. Psychol. Rev.},
  vol.~22, pp. 155--174, 2010.

\bibitem{tucker2022prototype}
M.~Tucker and J.~A. Shah, ``Prototype based classification from hierarchy to
  fairness,'' in \emph{ICML}, 2022, pp. 21\,884--21\,900.

\bibitem{rymarczyk2023icicle}
D.~Rymarczyk, J.~van~de Weijer, B.~Zieli{\'n}ski, and B.~Twardowski, ``Icicle:
  Interpretable class incremental continual learning,'' in \emph{ICCV}, 2023,
  pp. 1887--1898.

\bibitem{jiang2023fedskill}
Y.~Jiang, W.~Yu, D.~Song, L.~Wang, W.~Cheng, and H.~Chen, ``Fedskill: Privacy
  preserved interpretable skill learning via imitation,'' in \emph{SIGKDD},
  2023, pp. 1010--1019.

\bibitem{chen2019looks}
C.~Chen, O.~Li, D.~Tao, A.~Barnett, C.~Rudin, and J.~K. Su, ``This looks like
  that: deep learning for interpretable image recognition,'' \emph{NeurIPS},
  vol.~32, 2019.

\bibitem{kim2021xprotonet}
E.~Kim, S.~Kim, M.~Seo, and S.~Yoon, ``Xprotonet: diagnosis in chest
  radiography with global and local explanations,'' in \emph{CVPR}, 2021, pp.
  15\,719--15\,728.

\bibitem{keswani2022proto2proto}
M.~Keswani, S.~Ramakrishnan, N.~Reddy, and V.~N. Balasubramanian,
  ``Proto2proto: Can you recognize the car, the way i do?'' in \emph{CVPR},
  2022, pp. 10\,233--10\,243.

\bibitem{nauta2023pip}
M.~Nauta, J.~Schl{\"o}tterer, M.~van Keulen, and C.~Seifert, ``Pip-net:
  Patch-based intuitive prototypes for interpretable image classification,'' in
  \emph{CVPR}, 2023, pp. 2744--2753.

\bibitem{sacha2023protoseg}
M.~Sacha, D.~Rymarczyk, {\L}.~Struski, J.~Tabor, and B.~Zieli{\'n}ski,
  ``Protoseg: Interpretable semantic segmentation with prototypical parts,'' in
  \emph{WACV}, 2023, pp. 1481--1492.

\bibitem{carmichael2024pixel}
Z.~Carmichael, S.~Lohit, A.~Cherian, M.~J. Jones, and W.~J. Scheirer,
  ``Pixel-grounded prototypical part networks,'' in \emph{WACV}, 2024, pp.
  4768--4779.

\bibitem{hong2023protorynet}
D.~Hong, T.~Wang, and S.~Baek, ``Protorynet-interpretable text classification
  via prototype trajectories,'' \emph{JMLR}, vol.~24, no. 264, pp. 1--39, 2023.

\bibitem{xue2022protopformer}
M.~Xue, Q.~Huang, H.~Zhang, J.~Hu, J.~Song, M.~Song, and C.~Jin,
  ``Protopformer: Concentrating on prototypical parts in vision transformers
  for interpretable image recognition,'' in \emph{IJCAI}, 2024, pp. 1516--1524.

\bibitem{pach2025lucidppn}
M.~Pach, K.~Lewandowska, J.~Tabor, B.~M. Zieli{\'n}ski, and D.~D. Rymarczyk,
  ``Lucid{PPN}: Unambiguous prototypical parts network for user-centric
  interpretable computer vision,'' in \emph{ICLR}, 2025.

\bibitem{wang2021interpretable}
J.~Wang, H.~Liu, X.~Wang, and L.~Jing, ``Interpretable image recognition by
  constructing transparent embedding space,'' in \emph{ICCV}, 2021, pp.
  895--904.

\bibitem{kjaersgaard2024pantypes}
R.~Kj{\ae}rsgaard, A.~Boubekki, and L.~Clemmensen, ``Pantypes: Diverse
  representatives for self-explainable models,'' in \emph{AAAI}, 2024, pp.
  13\,230--13\,237.

\bibitem{gautam2022protovae}
S.~Gautam, A.~Boubekki, S.~Hansen, S.~Salahuddin, R.~Jenssen, M.~H{\"o}hne, and
  M.~Kampffmeyer, ``Protovae: A trustworthy self-explainable prototypical
  variational model,'' \emph{NeurIPS}, vol.~35, pp. 17\,940--17\,952, 2022.

\bibitem{haselhoff2025gaussian}
A.~Haselhoff, K.~Trelenberg, F.~K{\"u}ppers, and J.~Schneider, ``The gaussian
  discriminant variational autoencoder (gdvae): A self-explainable model with
  counterfactual explanations,'' in \emph{ECCV}, 2025, pp. 305--322.

\bibitem{ma2024looks}
C.~Ma, B.~Zhao, C.~Chen, and C.~Rudin, ``This looks like those: Illuminating
  prototypical concepts using multiple visualizations,'' \emph{NeurIPS},
  vol.~36, 2024.

\bibitem{nauta2021neural}
M.~Nauta, R.~Van~Bree, and C.~Seifert, ``Neural prototype trees for
  interpretable fine-grained image recognition,'' in \emph{CVPR}, 2021, pp.
  14\,933--14\,943.

\bibitem{hase2019interpretable}
P.~Hase, C.~Chen, O.~Li, and C.~Rudin, ``Interpretable image recognition with
  hierarchical prototypes,'' in \emph{AAAI on Human Computation and
  Crowdsourcing}, vol.~7, 2019, pp. 32--40.

\bibitem{wang2023learning}
C.~Wang, Y.~Liu, Y.~Chen, F.~Liu, Y.~Tian, D.~McCarthy, H.~Frazer, and
  G.~Carneiro, ``Learning support and trivial prototypes for interpretable
  image classification,'' in \emph{ICCV}, 2023, pp. 2062--2072.

\bibitem{rymarczyk2022interpretable}
D.~Rymarczyk, {\L}.~Struski, M.~G{\'o}rszczak, K.~Lewandowska, J.~Tabor, and
  B.~Zieli{\'n}ski, ``Interpretable image classification with differentiable
  prototypes assignment,'' in \emph{ECCV}, 2022, pp. 351--368.

\bibitem{ragodos2022protox}
R.~Ragodos, T.~Wang, Q.~Lin, and X.~Zhou, ``Protox: Explaining a reinforcement
  learning agent via prototyping,'' \emph{NeurIPS}, vol.~35, pp.
  27\,239--27\,252, 2022.

\bibitem{zhang2022protgnn}
Z.~Zhang, Q.~Liu, H.~Wang, C.~Lu, and C.~Lee, ``Protgnn: Towards
  self-explaining graph neural networks,'' in \emph{AAAI}, 2022, pp.
  9127--9135.

\bibitem{ming2019interpretable}
Y.~Ming, P.~Xu, H.~Qu, and L.~Ren, ``Interpretable and steerable sequence
  learning via prototypes,'' in \emph{SIGKDD}, 2019, pp. 903--913.

\bibitem{kenny2023towards}
E.~M. Kenny, M.~Tucker, and J.~Shah, ``Towards interpretable deep reinforcement
  learning with human-friendly prototypes,'' in \emph{ICLR}, 2023.

\bibitem{rymarczyk2021protopshare}
D.~Rymarczyk, {\L}.~Struski, J.~Tabor, and B.~Zieli{\'n}ski, ``Protopshare:
  Prototypical parts sharing for similarity discovery in interpretable image
  classification,'' in \emph{SIGKDD}, 2021, pp. 1420--1430.

\bibitem{van2022analyzing}
E.~van Krieken, E.~Acar, and F.~van Harmelen, ``Analyzing differentiable fuzzy
  logic operators,'' \emph{Artif. Intell.}, 2022.

\bibitem{petersen2022deep}
F.~Petersen, C.~Borgelt, H.~Kuehne, and O.~Deussen, ``Deep differentiable logic
  gate networks,'' \emph{NeurIPS}, pp. 2006--2018, 2022.

\bibitem{marra2020relational}
G.~Marra, M.~Diligenti, F.~Giannini, M.~Gori, and M.~Maggini, ``Relational
  neural machines,'' in \emph{ECAI}, 2020, pp. 1340--1347.

\bibitem{donadello2017logic}
I.~Donadello, L.~Serafini, and d.~G. Artur, ``Logic tensor networks for
  semantic image interpretation,'' in \emph{IJCAI}, 2017, pp. 1596--1602.

\bibitem{wang2020transparent}
Z.~Wang, W.~Zhang, L.~Ning, and J.~Wang, ``Transparent classification with
  multilayer logical perceptrons and random binarization,'' in \emph{AAAI},
  2020, pp. 6331--6339.

\bibitem{qiao2021learning}
L.~Qiao, W.~Wang, and B.~Lin, ``Learning accurate and interpretable decision
  rule sets from neural networks,'' in \emph{AAAI}, 2021, pp. 4303--4311.

\bibitem{wang2021scalable}
Z.~Wang, W.~Zhang, N.~Liu, and J.~Wang, ``Scalable rule-based representation
  learning for interpretable classification,'' \emph{NeurIPS}, pp.
  30\,479--30\,491, 2021.

\bibitem{wangrule2023learning}
------, ``Learning interpretable rules for scalable data representation and
  classification,'' \emph{TPAMI}, 2023.

\bibitem{yanghyperlogic}
Y.~Yang, W.~Ren, and S.~Li, ``Hyperlogic: Enhancing diversity and accuracy in
  rule learning with hypernets,'' in \emph{NeurIPS}, 2024.

\bibitem{petersenconvolutional}
F.~Petersen, H.~Kuehne, C.~Borgelt, J.~Welzel, and S.~Ermon, ``Convolutional
  differentiable logic gate networks,'' in \emph{NeurIPS}, 2024.

\bibitem{zhang2023learning_neuralrule}
W.~Zhang, Y.~Liu, Z.~Wang, and J.~Wang, ``Learning to binarize continuous
  features for neuro-rule networks.'' in \emph{IJCAI}, 2023, pp. 4584--4592.

\bibitem{ciravegna2020human}
G.~Ciravegna, F.~Giannini, M.~Gori, M.~Maggini, S.~Melacci \emph{et~al.},
  ``Human-driven fol explanations of deep learning,'' in \emph{IJCAI}, 2020,
  pp. 2234--2240.

\bibitem{ciravegna2023logic}
G.~Ciravegna, P.~Barbiero, F.~Giannini, M.~Gori, P.~Lió, M.~Maggini, and
  S.~Melacci, ``Logic explained networks,'' \emph{Artif. Intell.}, 2023.

\bibitem{jain2022extending}
R.~Jain, G.~Ciravegna, P.~Barbiero, F.~Giannini, D.~Buffelli, P.~Lio
  \emph{et~al.}, ``Extending logic explained networks to text classification,''
  in \emph{EMNLP}, 2022, pp. 8838--8857.

\bibitem{ciravegna2020constraint}
G.~Ciravegna, F.~Giannini, S.~Melacci, M.~Maggini, and M.~Gori, ``A
  constraint-based approach to learning and explanation,'' in \emph{AAAI},
  2020, pp. 3658--3665.

\bibitem{barbiero2022entropy}
P.~Barbiero, G.~Ciravegna, F.~Giannini, P.~Lió, M.~Gori, and S.~Melacci,
  ``Entropy-based logic explanations of neural networks,'' in \emph{AAAI},
  2022, pp. 6046--6054.

\bibitem{barbiero2023interpretable}
P.~Barbiero, G.~Ciravegna, F.~Giannini, M.~E. Zarlenga, L.~C. Magister,
  A.~Tonda, P.~Lió, F.~Precioso, M.~Jamnik, and G.~Marra, ``Interpretable
  neural-symbolic concept reasoning,'' in \emph{ICML}, 2023, pp. 1801--1825.

\bibitem{okajima2019deep}
Y.~Okajima and K.~Sadamasa, ``Deep neural networks constrained by decision
  rules,'' in \emph{AAAI}, 2019, pp. 2496--2505.

\bibitem{shi2022explainable}
S.~Shi, Y.~Xie, Z.~Wang, B.~Ding, Y.~Li, and M.~Zhang, ``Explainable neural
  rule learning,'' in \emph{WWW}, 2022, pp. 3031--3041.

\bibitem{yu2023finrule}
L.~Yu, M.~Li, Y.-L. Zhang, L.~Li, and J.~Zhou, ``Finrule: Feature interactive
  neural rule learning,'' in \emph{CIKM}, 2023, pp. 3020--3029.

\bibitem{wuweakly}
Z.~Wu, Z.~X. Zhang, A.~Naik, Z.~Mei, M.~Firdaus, and L.~Mou, ``Weakly
  supervised explainable phrasal reasoning with neural fuzzy logic,'' in
  \emph{ICLR}, 2021.

\bibitem{kontschieder2015deep}
P.~Kontschieder, M.~Fiterau, A.~Criminisi, and S.~R. Bulo, ``Deep neural
  decision forests,'' in \emph{CVPR}, 2015, pp. 1467--1475.

\bibitem{kim2022vit}
S.~Kim, J.~Nam, and B.~C. Ko, ``Vit-net: Interpretable vision transformers with
  neural tree decoder,'' in \emph{ICML}, 2022, pp. 11\,162--11\,172.

\bibitem{lao2024vitree}
D.~Lao, Q.~Liu, J.~Bu, J.~Yan, and W.~Shen, ``Vitree: Single-path neural tree
  for step-wise interpretable fine-grained visual categorization,'' in
  \emph{AAAI}, vol.~38, no.~3, 2024, pp. 2866--2873.

\bibitem{tanno2019adaptive}
R.~Tanno, K.~Arulkumaran, D.~Alexander, A.~Criminisi, and A.~Nori, ``Adaptive
  neural trees,'' in \emph{ICML}, 2019, pp. 6166--6175.

\bibitem{ji2020attention}
R.~Ji, L.~Wen, L.~Zhang, D.~Du, Y.~Wu, C.~Zhao, X.~Liu, and F.~Huang,
  ``Attention convolutional binary neural tree for fine-grained visual
  categorization,'' in \emph{CVPR}, 2020, pp. 10\,468--10\,477.

\bibitem{wangvisual}
W.~Wang, C.~Han, T.~Zhou, and D.~Liu, ``Visual recognition with deep nearest
  centroids,'' in \emph{ICLR}, 2024.

\bibitem{chen2024neural}
G.~Chen, X.~Li, Y.~Yang, and W.~Wang, ``Neural clustering based visual
  representation learning,'' in \emph{CVPR}, 2024, pp. 5714--5725.

\bibitem{zhang2019pathologist}
Z.~Zhang, P.~Chen, M.~McGough, F.~Xing, C.~Wang, M.~Bui, Y.~Xie, M.~Sapkota,
  L.~Cui, J.~Dhillon \emph{et~al.}, ``Pathologist-level interpretable
  whole-slide cancer diagnosis with deep learning,'' \emph{Nat. Mach. Intell.},
  vol.~1, no.~5, pp. 236--245, 2019.

\bibitem{xu2020explainable}
Y.~Xu, X.~Yang, L.~Gong, H.-C. Lin, T.-Y. Wu, Y.~Li, and N.~Vasconcelos,
  ``Explainable object-induced action decision for autonomous vehicles,'' in
  \emph{CVPR}, 2020, pp. 9523--9532.

\bibitem{yi2018neural}
K.~Yi, J.~Wu, C.~Gan, A.~Torralba, P.~Kohli, and J.~Tenenbaum,
  ``Neural-symbolic vqa: Disentangling reasoning from vision and language
  understanding,'' \emph{NeurIPS}, vol.~31, 2018.

\bibitem{zhang2020interpretable}
Q.~Zhang, X.~Wang, Y.~N. Wu, H.~Zhou, and S.-C. Zhu, ``Interpretable cnns for
  object classification,'' \emph{TPAMI}, pp. 3416--3431, 2020.

\bibitem{shen2021interpretable}
X.~Shen, Y.~Luo, Z.~Liu, and L.~Song, ``Interpretable compositional
  convolutional neural networks,'' in \emph{IJCAI}, 2021, pp. 2971--2978.

\bibitem{liang2020training}
H.~Liang, Z.~Ouyang, Y.~Zeng, H.~Su, Z.~He, S.-T. Xia, J.~Zhu, and B.~Zhang,
  ``Training interpretable convolutional neural networks by differentiating
  class-specific filters,'' in \emph{ECCV}, 2020, pp. 622--638.

\bibitem{guo2024picnn}
W.~Guo, J.~Yang, H.~Yin, Q.~Chen, and W.~Ye, ``Picnn: A pathway towards
  interpretable convolutional neural networks,'' in \emph{AAAI}, 2024, pp.
  2003--2012.

\bibitem{rajagopal2021selfexplain}
D.~Rajagopal, V.~Balachandran, E.~H. Hovy, and Y.~Tsvetkov, ``Selfexplain: A
  self-explaining architecture for neural text classifiers,'' in \emph{EMNLP},
  2021, pp. 836--850.

\bibitem{greff2019interpretable}
K.~Greff, S.~Van~Steenkiste, and J.~Schmidhuber, ``Interpretable neural
  predictions with differentiable binary variables,'' \emph{Artif. Intell.},
  vol. 289, p. 103396, 2019.

\bibitem{yang2016hierarchical}
Z.~Yang, D.~Yang, C.~Dyer, X.~He, A.~Smola, and E.~Hovy, ``Hierarchical
  attention networks for document classification,'' in \emph{NAACL}, 2016, pp.
  1480--1489.

\bibitem{ghaeini2018interpreting}
R.~Ghaeini, X.~Z. Fern, and P.~Tadepalli, ``Interpreting recurrent and
  attention-based neural models: a case study on natural language inference,''
  \emph{ACL}, 2018.

\bibitem{subramanian2018spine}
A.~Subramanian, D.~Pruthi, H.~Jhamtani, T.~Berg-Kirkpatrick, and E.~Hovy,
  ``Spine: Sparse interpretable neural embeddings,'' in \emph{AAAI}, 2018.

\bibitem{du2019learning}
M.~Du, N.~Liu, F.~Yang, and X.~Hu, ``Learning credible deep neural networks
  with rationale regularization,'' in \emph{ICDM}, 2019, pp. 150--159.

\bibitem{arik2020protoattend}
S.~O. Arik and T.~Pfister, ``Protoattend: Attention-based prototypical
  learning,'' \emph{JMLR}, vol.~21, no. 210, pp. 1--35, 2020.

\bibitem{chan2022models}
A.~Chan, J.~Qin, Y.~Chen, V.~Srivastava, and P.~Agrawal, ``When can models
  learn from explanations? a formal framework for understanding the roles of
  explanation data,'' in \emph{NeurIPS}, 2022.

\bibitem{zhang2023summarize}
N.~Zhang, Y.~Xiao, and B.~Yu, ``Summarize-then-answer: Generating concise
  explanations for multi-hop reading comprehension,'' in \emph{EMNLP}, 2023.

\bibitem{jacovi2021aligning}
A.~Jacovi and Y.~Goldberg, ``Aligning faithful interpretations with their
  social attribution,'' \emph{TACL}, vol.~9, pp. 294--310, 2021.

\bibitem{kumar2022nile}
S.~Kumar and P.~Talukdar, ``{NILE}: Natural language inference with faithful
  natural language explanations,'' in \emph{ACL}, 2022, pp. 8664--8684.

\bibitem{rajani2019explain}
N.~F. Rajani, B.~McCann, C.~Xiong, and R.~Socher, ``Explain {Y}ourself!
  {L}everaging language models for commonsense reasoning,'' in \emph{ACL},
  2019, pp. 4932--4942.

\bibitem{sun2023lirex}
Y.~Sun, Y.~Ji, Y.~Zhang, and Z.~Wang, ``{LIRE}x: Augmenting language inference
  with relevant explanation,'' in \emph{AAAI}, 2023, pp. 13\,944--13\,952.

\bibitem{camburu2018towards}
O.~Camburu, T.~Rockt{\"a}schel, T.~Lukasiewicz, and P.~Blunsom, ``Towards
  interpretable natural language understanding with explanations as latent
  variables,'' in \emph{NeurIPS}, 2018, pp. 4864--4873.

\bibitem{wiegreffe2021measuring}
S.~Wiegreffe, A.~Marasovi{\'c}, and N.~A. Smith, ``Measuring association
  between labels and free-text rationales,'' in \emph{EMNLP}, 2021, pp.
  10\,266--10\,284.

\bibitem{marasovic2022few}
A.~Marasović, I.~Beltagy, D.~Downey, and M.~E. Peters, ``Few-shot
  self-rationalization with natural language prompts,'' in \emph{NAACL}, 2022,
  pp. 410--424.

\bibitem{narang2020wt5}
S.~Narang, C.~Raffel, K.~Lee, A.~Roberts, N.~Fiedel, and K.~Malkan, ``Wt5?!
  training text-to-text models to explain their predictions,'' \emph{arXiv
  preprint arXiv:2004.14546}, 2020.

\bibitem{liu2019towards}
H.~Liu, Q.~Yin, and W.~Y. Wang, ``Towards explainable nlp: A generative
  explanation framework for text classification,'' in \emph{ACL}, 2019, pp.
  5570--5581.

\bibitem{zhang2023triple}
Y.~Zhang, Y.~Sun, F.~Zhuang, Y.~Zhu, Z.~An, and Y.~Xu, ``Triple dual learning
  for opinion-based explainable recommendation,'' \emph{TOIS}, 2023.

\bibitem{camburu2018snli}
O.-M. Camburu, T.~Rockt{\"a}schel, T.~Lukasiewicz, and P.~Blunsom, ``e-snli:
  Natural language inference with natural language explanations,''
  \emph{Advances in Neural Information Processing Systems}, vol.~31, 2018.

\bibitem{wei2022chain}
J.~Wei, X.~Wang, D.~Schuurmans, M.~Bosma, F.~Xia, E.~Chi, Q.~V. Le, D.~Zhou
  \emph{et~al.}, ``Chain-of-thought prompting elicits reasoning in large
  language models,'' \emph{NeurIPS}, vol.~35, pp. 24\,824--24\,837, 2022.

\bibitem{yao2024tree}
S.~Yao, D.~Yu, J.~Zhao, I.~Shafran, T.~Griffiths, Y.~Cao, and K.~Narasimhan,
  ``Tree of thoughts: Deliberate problem solving with large language models,''
  \emph{NeurIPS}, vol.~36, 2024.

\bibitem{madsen2024self}
A.~Madsen, S.~Chandar, and S.~Reddy, ``Are self-explanations from large
  language models faithful?'' in \emph{ACL}, 2024, pp. 295--337.

\bibitem{yeo2024interpretable}
W.~J. Yeo, R.~Satapathy, R.~S.~M. Goh, and E.~Cambria, ``How interpretable are
  reasoning explanations from prompting large language models?'' in \emph{ACL},
  2024, pp. 1800--1815.

\bibitem{jin2020multi}
W.~Jin, R.~Barzilay, and T.~Jaakkola, ``Multi-objective molecule generation
  using interpretable substructures,'' in \emph{ICML}, 2020, pp. 4849--4859.

\bibitem{MMGNN}
W.~Du, S.~Zhang, D.~Wu, J.~Xia, Z.~Zhao, J.~Fang, and Y.~Wang, ``Mmgnn: A
  molecular merged graph neural network for explainable solvation free energy
  prediction,'' in \emph{IJCAI}, 2024, pp. 5808--5816.

\bibitem{lanciano2020explainable}
T.~Lanciano, F.~Bonchi, and A.~Gionis, ``Explainable classification of brain
  networks via contrast subgraphs,'' in \emph{KDD}, 2020, pp. 3308--3318.

\bibitem{huang2025sehg}
Z.~Huang, W.~Zhou, Y.~Li, X.~Wu, C.~Xu, J.~Fang, Z.~Jia, L.~L{\"u}, and F.~Xia,
  ``Sehg: Bridging interpretability and prediction in self-explainable
  heterogeneous graph neural networks,'' in \emph{WWW}, 2025.

\bibitem{yu2022improving}
J.~Yu, J.~Cao, and R.~He, ``Improving subgraph recognition with variational
  graph information bottleneck,'' in \emph{CVPR}, 2022, pp. 19\,396--19\,405.

\bibitem{yugraph2021}
J.~Yu, T.~Xu, Y.~Rong, Y.~Bian, J.~Huang, and R.~He, ``Graph information
  bottleneck for subgraph recognition,'' in \emph{ICLR}, 2021.

\bibitem{chen2024interpretable}
Y.~Chen, Y.~Bian, B.~Han, and J.~Cheng, ``How interpretable are interpretable
  graph neural networks?'' in \emph{ICML}, 2024.

\bibitem{dai2021towards}
E.~Dai and S.~Wang, ``Towards self-explainable graph neural network,'' in
  \emph{CIKM}, 2021, pp. 302--311.

\bibitem{feng2022kergnns}
A.~Feng, C.~You, S.~Wang, and L.~Tassiulas, ``Kergnns: Interpretable graph
  neural networks with graph kernels,'' in \emph{AAAI}, 2022, pp. 6614--6622.

\bibitem{giannini2024interpretable}
F.~Giannini, S.~Fioravanti, O.~Keskin, A.~Lupidi, L.~C. Magister, P.~Lió, and
  P.~Barbiero, ``Interpretable graph networks formulate universal algebra
  conjectures,'' \emph{NeurIPS}, 2024.

\bibitem{wudiscovering}
Y.~Wu, X.~Wang, A.~Zhang, X.~He, and T.-S. Chua, ``Discovering invariant
  rationales for graph neural networks,'' in \emph{ICLR}, 2022.

\bibitem{wang2024unveiling}
Y.~Wang, S.~Liu, T.~Zheng, K.~Chen, and M.~Song, ``Unveiling global interactive
  patterns across graphs: Towards interpretable graph neural networks,'' in
  \emph{KDD}, 2024, pp. 3277--3288.

\bibitem{yang2025from}
J.~Yang, Y.~Wang, K.~Chen, T.~Zheng, Y.~Zhou, Z.~Xiao, J.~Cao, M.~Song, and
  S.~Liu, ``From {GNN}s to trees: Multi-granular interpretability for graph
  neural networks,'' in \emph{ICLR}, 2025.

\bibitem{annasamy2019towards}
R.~M. Annasamy and K.~Sycara, ``Towards better interpretability in deep
  q-networks,'' in \emph{AAAI}, 2019, pp. 4561--4569.

\bibitem{mott2019towards}
A.~Mott, D.~Zoran, M.~Chrzanowski, D.~Wierstra, and D.~Jimenez~Rezende,
  ``Towards interpretable reinforcement learning using attention augmented
  agents,'' \emph{NeurIPS}, vol.~32, 2019.

\bibitem{shi2020self}
W.~Shi, G.~Huang, S.~Song, Z.~Wang, T.~Lin, and C.~Wu, ``Self-supervised
  discovering of interpretable features for reinforcement learning,''
  \emph{TPAMI}, vol.~44, no.~5, pp. 2712--2724, 2020.

\bibitem{liu2021learning}
G.~Liu, X.~Sun, O.~Schulte, and P.~Poupart, ``Learning tree interpretation from
  object representation for deep reinforcement learning,'' \emph{NeurIPS}, pp.
  19\,622--19\,636, 2021.

\bibitem{wang2019alphastock}
J.~Wang, Y.~Zhang, K.~Tang, J.~Wu, and Z.~Xiong, ``Alphastock: A
  buying-winners-and-selling-losers investment strategy using interpretable
  deep reinforcement attention networks,'' in \emph{SIGKDD}, 2019, pp.
  1900--1908.

\bibitem{yu2023explainable}
Z.~Yu, J.~Ruan, and D.~Xing, ``Explainable reinforcement learning via a causal
  world model,'' in \emph{IJCAI}, 2023, pp. 4540--4548.

\bibitem{zhang2024interpretable}
Y.~Zhang, Y.~Du, B.~Huang, Z.~Wang, J.~Wang, M.~Fang, and M.~Pechenizkiy,
  ``Interpretable reward redistribution in reinforcement learning: a causal
  approach,'' \emph{NeurIPS}, 2024.

\bibitem{yau2020did}
H.~Yau, C.~Russell, and S.~Hadfield, ``What did you think would happen?
  explaining agent behaviour through intended outcomes,'' \emph{NeurIPS}, pp.
  18\,375--18\,386, 2020.

\bibitem{verma2023interpretable}
A.~Verma, H.~M. Le, Y.~Yue, and S.~Chaudhuri, ``Interpretable and explainable
  logical policies via neurally guided symbolic abstraction,'' in \emph{ICLR},
  2023.

\bibitem{guo2023efficient}
J.~Guo, R.~Zhang, S.~Peng, Q.~Yi, X.~Hu, R.~Chen, Z.~Du, X.~Zhang, L.~Li,
  Q.~Guo, and Y.~Chen, ``Efficient symbolic policy learning with differentiable
  symbolic expression,'' in \emph{NeurIPS}, 2023.

\bibitem{li2023differentiable}
X.~Li, H.~Lei, L.~Zhang, and M.~Wang, ``Differentiable logic policy for
  interpretable deep reinforcement learning: A study from an optimization
  perspective,'' \emph{TPAMI}, vol.~45, no.~10, pp. 11\,654--11\,667, 2023.

\bibitem{verma2019lirme}
M.~Verma and D.~Ganguly, ``Lirme: locally interpretable ranking model
  explanation,'' in \emph{SIGIR}, 2019, pp. 1281--1284.

\bibitem{chu2018exact}
L.~Chu, X.~Hu, J.~Hu, L.~Wang, and J.~Pei, ``Exact and consistent
  interpretation for piecewise linear neural networks: A closed form
  solution,'' in \emph{SIGKDD}, 2018, pp. 1244--1253.

\bibitem{huang2023evaluation}
Q.~Huang, M.~Xue, W.~Huang, H.~Zhang, J.~Song, Y.~Jing, and M.~Song,
  ``Evaluation and improvement of interpretability for self-explainable
  part-prototype networks,'' in \emph{ICCV}, 2023, pp. 2011--2020.

\bibitem{sacha2024interpretability}
M.~Sacha, B.~Jura, D.~Rymarczyk, {\L}.~Struski, J.~Tabor, and B.~Zieli{\'n}ski,
  ``Interpretability benchmark for evaluating spatial misalignment of
  prototypical parts explanations,'' in \emph{AAAI}, 2024, pp.
  21\,563--21\,573.

\bibitem{zhou2023solvability}
Y.~Zhou and J.~Shah, ``The solvability of interpretability evaluation
  metrics,'' in \emph{ACL}, 2023, pp. 2399--2415.

\bibitem{pope2019explainability}
P.~E. Pope, S.~Kolouri, M.~Rostami, C.~E. Martin, and H.~Hoffmann,
  ``Explainability methods for graph convolutional neural networks,'' in
  \emph{CVPR}, 2019, pp. 10\,772--10\,781.

\bibitem{yeh2020completeness}
C.-K. Yeh, B.~Kim, S.~Arik, C.-L. Li, T.~Pfister, and P.~Ravikumar, ``On
  completeness-aware concept-based explanations in deep neural networks,''
  \emph{NeurIPS}, pp. 20\,554--20\,565, 2020.

\bibitem{qian2024towards}
W.~Qian, C.~Zhao, Y.~Li, F.~Ma, C.~Zhang, and M.~Huai, ``Towards modeling
  uncertainties of self-explaining neural networks via conformal prediction,''
  in \emph{AAAI}, 2024, pp. 14\,651--14\,659.

\bibitem{zarlenga2023towards}
M.~E. Zarlenga, P.~Barbiero, Z.~Shams, D.~Kazhdan, U.~Bhatt, A.~Weller, and
  M.~Jamnik, ``Towards robust metrics for concept representation evaluation,''
  in \emph{AAAI}, vol.~37, no.~10, 2023, pp. 11\,791--11\,799.

\bibitem{rong2023towards}
Y.~Rong, T.~Leemann, T.-T. Nguyen, L.~Fiedler, P.~Qian, V.~Unhelkar, T.~Seidel,
  G.~Kasneci, and E.~Kasneci, ``Towards human-centered explainable ai: A survey
  of user studies for model explanations,'' \emph{TPAMI}, 2023.

\bibitem{sundararajan2017axiomatic}
M.~Sundararajan, A.~Taly, and Q.~Yan, ``Axiomatic attribution for deep
  networks,'' in \emph{ICML}, 2017, pp. 3319--3328.

\bibitem{setiono1995understanding}
R.~Setiono and H.~Liu, ``Understanding neural networks via rule extraction,''
  in \emph{IJCAI}, vol.~1, 1995, pp. 480--485.

\bibitem{ghosh2023dividing}
S.~Ghosh, K.~Yu, F.~Arabshahi, and K.~Batmanghelich, ``Dividing and conquering
  a blackbox to a mixture of interpretable models: route, interpret, repeat,''
  in \emph{ICML}, 2023.

\bibitem{zhang2019interpreting}
Q.~Zhang, Y.~Yang, H.~Ma, and Y.~N. Wu, ``Interpreting cnns via decision
  trees,'' in \emph{CVPR}, 2019, pp. 6261--6270.

\bibitem{song2021tree}
J.~Song, H.~Zhang, X.~Wang, M.~Xue, Y.~Chen, L.~Sun, D.~Tao, and M.~Song,
  ``Tree-like decision distillation,'' in \emph{CVPR}, 2021, pp.
  13\,488--13\,497.

\bibitem{choi2024adaptive}
J.~Choi, J.~Raghuram, Y.~Li, S.~Banerjee, and S.~Jha, ``Adaptive concept
  bottleneck for foundation models,'' in \emph{ICLR}, 2025.

\bibitem{enoueninstashap}
J.~Enouen and Y.~Liu, ``Instashap: Interpretable additive models explain
  shapley values instantly,'' in \emph{ICLR}, 2025.

\bibitem{krishna2024post}
S.~Krishna, J.~Ma, D.~Slack, A.~Ghandeharioun, S.~Singh, and H.~Lakkaraju,
  ``Post hoc explanations of language models can improve language models,''
  \emph{NeurIPS}, vol.~36, 2024.

\bibitem{liu2025utilizing}
S.~Liu and M.~Zhu, ``Utilizing explainable reinforcement learning to improve
  reinforcement learning: A theoretical and systematic framework,'' in
  \emph{ICLR}, 2025.

\bibitem{quan2024verification}
X.~Quan, M.~Valentino, L.~A. Dennis, and A.~Freitas, ``Verification and
  refinement of natural language explanations through llm-symbolic theorem
  proving,'' in \emph{EMNLP}, 2024.

\bibitem{singh2023augmenting}
C.~Singh, A.~Askari, R.~Caruana, and J.~Gao, ``Augmenting interpretable models
  with large language models during training,'' \emph{Nature Communications},
  vol.~14, no.~1, p. 7913, 2023.

\bibitem{bostrom2024protolm}
K.~Bostrom, N.~Durrani, and P.~Bielik, ``Proto-lm: A prototypical network-based
  framework for built-in interpretability in large language models,'' in
  \emph{EMNLP}, 2023.

\bibitem{gandhi2024large}
S.~Gandhi, A.~Koller, A.~Guha, C.~Klein, Z.~Chen, and J.~Steinhardt, ``Large
  language models are interpretable learners,'' in \emph{ICLR}, 2025.

\bibitem{huben2023sparse}
R.~Huben, H.~Cunningham, L.~R. Smith, A.~Ewart, and L.~Sharkey, ``Sparse
  autoencoders find highly interpretable features in language models,'' in
  \emph{ICLR}, 2023.

\bibitem{dunefskytranscoders}
J.~Dunefsky, P.~Chlenski, and N.~Nanda, ``Transcoders find interpretable llm
  feature circuits,'' in \emph{NeurIPS}, 2023.

\end{thebibliography}

\end{document}